\pdfoutput=1
\documentclass{article}
\usepackage[preprint,main]{neurips_2026}
\usepackage[T1]{fontenc}
\usepackage[utf8]{inputenc}
\usepackage{amsmath,amssymb,amsthm,mathtools}
\usepackage{booktabs,tabularx,longtable,array}
\usepackage{graphicx,microtype,xcolor,hyperref,url}
\usepackage{placeins,flafter}
\usepackage{needspace}
\usepackage{tikz}
\usetikzlibrary{arrows.meta,calc}
\usepackage{alphalph}
\usepackage{fancyhdr}
\hypersetup{colorlinks=true,linkcolor=blue!45!black,citecolor=blue!45!black,urlcolor=blue!45!black,
 pdftitle={Depth or Description? When a One-Step Latent Score Replaces Lookahead in Proactive Information Acquisition},pdfauthor={Biswa Sengupta}}
\newcommand{\E}{\mathbb{E}}

\newcommand{\code}[1]{{\small\nolinkurl{#1}}}

\theoremstyle{definition}
\newtheorem{theorem}{Theorem}[section]
\title{Depth or Description?\\
When a One-Step Latent Score Replaces Lookahead\\
in Proactive Information Acquisition}
\author{Biswa Sengupta\\
LLM Suite Team, JPMorgan Chase \& Co.\\
\texttt{biswa.sengupta@jpmorgan.com}}
\begin{document}
\maketitle
\thispagestyle{fancy}
\raggedbottom
\begin{abstract}
A proactive agent must decide what to ask before it acts. Exact lookahead values combinations of questions but is expensive and leans on a fitted model; a one-step score is cheap but can miss them. Our main finding: the need for depth often lies in how the hidden cause is described. A one-step score on the label can be blind to an interaction among answers. A score on a suitable latent cause is not, and where that latent encodes the decision it replaces most of depth at a fraction of the search. Described differently, the same problem can defeat the score while exact values do not move, so keep the latent small and check it against exact values. Probabilities learned from one language model's answers do not carry to another unless planning allows for both; those read from tokens are worth only a few labels.
 \end{abstract}

\begin{quote}
\footnotesize
\textbf{Disclaimer:} This paper was prepared for informational purposes by the LLM Suite group of JP Morgan
Chase and its affiliates (`JPMC') and is not a product of the Research Department of JP Morgan. JP
Morgan makes no representation, warranty or undertaking whatsoever and disclaims all liability for
the completeness, accuracy or reliability of the information contained herein. This document is not
intended as investment research or investment advice, or a recommendation, offer or solicitation for the
purchase or sale of any security, financial instrument, financial product or service, or to be used in any
way for evaluating the merits of participating in any transaction, and shall not constitute a solicitation
under any jurisdiction or to any person, if such solicitation under such jurisdiction or to such person
would be unlawful.
\end{quote}

\section{Introduction}
A proactive agent must often decide whether to act now or to investigate first. A market-surveillance agent can pull order-flow data before raising an alert; a diagnostic assistant can ask one more question before answering. Investigations cost time and money, so the agent needs a policy that chooses the next question and decides when to stop. We ask when it must look several questions ahead.

The difficulty is that a question's value may appear only in combination with later ones. Consider a case whose class is encoded in one of several blocks of features, with a context feature saying which block. The context reveals nothing about the class on its own, yet it is the right first question. A one-step score such as conditional mutual information (CMI) between the label and the next answer is cheap, but it sees only what a single answer reveals. Exact $k$-step lookahead (D$k$) sees combinations of up to $k$ answers, but its cost grows exponentially in $k$. It also depends on a fitted model, and a search cut short by a compute budget can do worse than a shallower one.

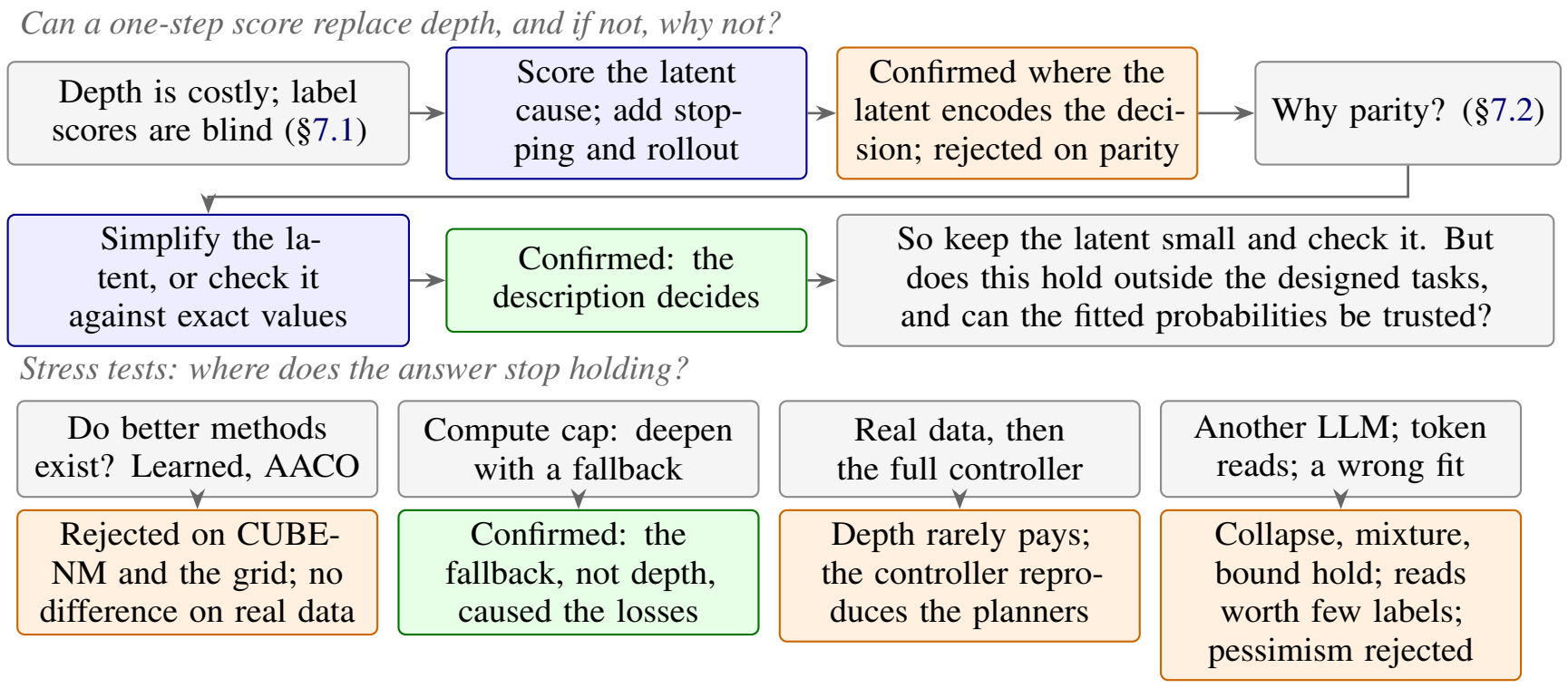
\begin{figure}[t]
\centering
\begin{tikzpicture}[font=\scriptsize, >={Stealth[length=1.5mm]},
  q/.style={draw=black!45, rounded corners=1.5pt, fill=gray!8, align=center, inner sep=2.5pt,
            minimum height=7.5mm},
  s/.style={draw=blue!55!black, rounded corners=1.5pt, fill=blue!7, align=center, inner sep=2.5pt,
            minimum height=7.5mm},
  yes/.style={draw=green!45!black, rounded corners=1.5pt, fill=green!10, align=center, inner sep=2.5pt,
            minimum height=7.5mm},
  mixed/.style={draw=orange!80!black, rounded corners=1.5pt, fill=orange!12, align=center, inner sep=2.5pt,
            minimum height=7.5mm},
  no/.style={draw=red!55!black, rounded corners=1.5pt, fill=red!8, align=center, inner sep=2.5pt,
            minimum height=7.5mm},
  head/.style={font=\scriptsize\itshape, text=black!60, anchor=west},
  flow/.style={->, draw=black!60, line width=.5pt}]
\node[head] at (-0.1,3.62) {Can a one-step score replace depth, and if not, why not?};
\node[q, text width=2.75cm] (p1) at (1.4,2.98) {Depth is costly; label scores are blind (\S\ref{sec:main-decision-aware})};
\node[s, text width=2.45cm] (s1) at (4.45,2.98) {Score the latent cause; add stopping and rollout};
\node[mixed, text width=2.45cm] (v1) at (7.3,2.98) {Confirmed where the latent encodes the decision; rejected on parity};
\node[q, text width=2.05cm] (p2) at (10.15,2.98) {Why parity? (\S\ref{sec:main-representation})};
\node[s, text width=2.75cm] (s2) at (1.4,1.76) {Simplify the latent, or check it against exact values};
\node[yes, text width=2.45cm] (v2) at (4.45,1.76) {Confirmed: the description decides};
\node[q, text width=5.05cm] (t) at (8.6,1.76) {So keep the latent small and check it. But does this hold outside the designed tasks, and can the fitted probabilities be trusted?};
\draw[flow] (p1)--(s1); \draw[flow] (s1)--(v1); \draw[flow] (v1)--(p2);
\draw[flow] (p2.south) |- (1.4,2.37) -- (s2.north);
\draw[flow] (s2)--(v2); \draw[flow] (v2)--(t);
\node[head] at (-0.1,1.1) {Stress tests: where does the answer stop holding?};
\foreach \x/\style/\ptext/\vtext in {
  1.32/mixed/{Do better methods exist? Learned, AACO}/{Rejected on CUBE-NM and the grid; no difference on real data},
  4.1/yes/{Compute cap: deepen with a fallback}/{Confirmed: the fallback, not depth, caused the losses},
  6.88/mixed/{Real data, then the full controller}/{Depth rarely pays; the controller reproduces the planners},
  9.66/mixed/{Another LLM; token reads; a wrong fit}/{Collapse, mixture, bound hold; reads worth few labels; pessimism rejected}}{
  \node[q, text width=2.45cm, minimum height=7mm] at (\x,.53) {\ptext};
  \node[\style, text width=2.45cm, minimum height=7.5mm, anchor=north] at (\x,.09) {\vtext};
  \draw[flow] (\x,.17) -- (\x,.09);}
\end{tikzpicture}
\caption{The argument. Grey: a problem. Blue: the solution tested. Green and orange: a confirmed or a mixed verdict. The first results find what decides whether a one-step score replaces depth; the rest probe the limits of that answer.}\label{fig:arc}
\end{figure}

Our answer: a label score is blind to an interaction hidden from every shorter set of answers, and a score on a latent cause is not. We score questions by latent information gain (LIG), the information an answer carries about the fitted cause, follow its sequence with an exact stopping rule (LIG-stop) and add one step of rollout (Rollout-LIG). Where the latent encodes the decision, this recovered most of depth's value on our tasks; elsewhere depth is still needed, or the latent should change. Figure~\ref{fig:arc} gives the argument: each step is a problem, a solution, and a verdict from proofs and simulations.

\section{Related work}
\paragraph{Planning to acquire information.} Choosing questions is planning in a partially observable decision process \citep{kaelbling1998}; rollout improves a base policy by trying each action followed by it \citep{bertsekas2022rollout}. Adaptive submodularity makes greedy acquisition near-optimal \citep{golovin2011}, but it belongs to an objective, not to a task. On parity, a bit's information about the label is zero until the rest of its block is known, so that objective's gains grow with the history.

\paragraph{Scores on hypotheses rather than labels.} BALD scores a query by its information about parameters \citep{houlsby2011}; equivalence-class edge cutting (EC$^2$) by the hypothesis pairs with different decisions it separates, an objective that stays adaptive submodular on noiseless parity \citep{golovin2010}, although its greedy bound needs about 28 tests on our smallest parity task, whose budget is three (Appendix~\ref{app:tier1}). ECED extends it to noisy answers \citep{chen2017eced}. LIG is of this family; we add a theory of when such a score is blind and how it depends on its description.

\paragraph{Active feature acquisition.} Learned methods include DIME \citep{gadgil2024dime}, GDFS \citep{covert2023gdfs} and SEFA \citep{norcliffe2025}; the Acquisition Conditioned Oracle (AACO) approximates a non-greedy oracle with nearest neighbours \citep{valancius2024}. We use AFABench's CUBE-NM, a context-gated task built to expose myopic acquisition \citep{afabench2026}, and the items of Uncertainty of Thoughts (UoT) \citep{hu2024uot}, with fixed question banks, not dialogue.
\section{The acquisition problem and the controller}\label{sec:main-controller}
\paragraph{The problem.} A hidden cause $z$ from a finite set $\mathcal Z$ generates the answers $O_j$ to questions $j\in\{1,\dots,n\}$ and determines the label $y$ the agent must predict. A history $h$ records the answers received so far, and a belief $q(z)$ summarizes it. At each step the agent either asks an unasked question, at cost $c$, or stops and acts. With signed utility a correct prediction earns $+1$, a wrong one $-1$ and abstaining $0$, minus the costs paid; hard-budget tasks instead require exactly $B$ questions and score accuracy. The agent replans after every answer. For a question $j$ with answer model $L_j(o\mid z,h)$, Bayes' rule gives the answer distribution and the updated belief:
\begin{equation}
 p(o\mid h,j)=\sum_z q(z)L_j(o\mid z,h),\qquad
 q'(z)=\frac{q(z)L_j(o\mid z,h)}{p(o\mid h,j)}.
\end{equation}

\paragraph{The controller.} Our implemented controller follows this loop. It keeps the belief $q$, and a gate decides between investigating and acting: it computes the lookahead value of Equation~\eqref{eq:depth} and investigates when that beats acting now. Guards and a verifier can still change the action. The controller can also weigh tail risk, replacing an action's expected return $R$ by $(1-\mu)\E[R]-\mu\operatorname{CVaR}_{\alpha}(-R)$ with CVaR the conditional value at risk. Every result in Section~\ref{sec:main-results} sets $\mu=0$, which switches that term off. Those tasks are scored on the mean, so any $\mu>0$ would optimize something other than the score. Appendix~\ref{app:risk-planner} tests $\mu>0$ on models with a rare bad state.

\paragraph{Where the probabilities come from.} A language model (LLM) does not produce calibrated probabilities, and the controller never asks one to state a probability. In a new situation with no fitted model, the controller reads the likelihoods $L_g(o\mid z,h)$ from the token log probabilities of the allowed outcomes, under a prompt that assumes the cause is $z$ and asks which outcome $g$ will reveal, from an ensemble of four local models. A temperature fitted on labeled outcomes rescales them, and their pooled row becomes a Dirichlet prior that labeled counts then update. The lookahead makes no LLM call. None of this makes the numbers true; Section~\ref{sec:main-reads} measures them. Every other result in Section~\ref{sec:main-results} fits probabilities to rows of data or counts an LLM's repeated answers. A preference can also be certified from validation data, whatever the probabilities' source (Theorem~\ref{thm:robust}). See Appendix~\ref{sec:controller}.
\section{Planners: depth search versus one-step scores}\label{sec:main-methods}
\begin{figure}[t]
\centering
\begin{tikzpicture}[
  font=\scriptsize, >={Stealth[length=1.5mm]},
  hist/.style={draw, rounded corners=1pt, minimum width=3.4mm, minimum height=2.8mm,
               inner sep=0.5pt, fill=gray!10},
  ans/.style={draw, circle, minimum size=2mm, inner sep=0pt, fill=white},
  term/.style={draw=orange!80!black, fill=orange!35, minimum size=1.7mm, inner sep=0pt},
  pick/.style={draw=blue!65!black, line width=0.9pt},
  chain/.style={draw=blue!65!black, line width=0.9pt, dash pattern=on 2pt off 1pt},
  title/.style={font=\footnotesize\bfseries, anchor=west},
  note/.style={align=center, font=\tiny, text=black!75},
  rowlabel/.style={font=\footnotesize\itshape, anchor=west, text=black!70}]
\node[rowlabel] at (-0.1,2.05) {Depth search and rollout on the same tree: where each search stops};
\foreach \k/\name/\lab in {0/D1/a,1/D2/b,2/D3/c,3/Rollout/d}{
  \begin{scope}[xshift=\k*3.45cm]
    \node[title] at (0,1.6) {(\lab) \name};
    \node[hist] (r-\k) at (1.55,1.15) {$h$};
    \foreach \x/\j in {0.45/1,1.55/2,2.65/3}{
      \node[ans] (c-\k-\j) at (\x,0.6) {};
      \draw[->] (r-\k) -- node[fill=white,inner sep=0.4pt,font=\tiny]{$j_{\j}$} (c-\k-\j);
      \node[hist,minimum width=2.6mm] (h-\k-\j-a) at ($(c-\k-\j)+(-0.22,-0.48)$) {};
      \node[hist,minimum width=2.6mm] (h-\k-\j-b) at ($(c-\k-\j)+(0.22,-0.48)$) {};
      \draw (c-\k-\j) -- (h-\k-\j-a); \draw (c-\k-\j) -- (h-\k-\j-b);}
  \end{scope}}
\foreach \j in {1,2,3}{\foreach \s in {a,b}{\draw (h-0-\j-\s) -- ++(0,-0.3) node[term,below]{};}}
\foreach \dx in {-0.32,0,0.32}{\draw (h-1-1-a) -- ++(\dx,-0.32) node[ans,below,minimum size=1.5mm](g){};
  \draw (g) -- ++(0,-0.2) node[term,below]{};}
\node at ($(h-1-2-a)+(0.2,-0.45)$) {$\cdots$};
\foreach \dx in {-0.32,0,0.32}{\draw (h-2-1-a) -- ++(\dx,-0.32) node[ans,below,minimum size=1.5mm](g){};}
\draw ($(h-2-1-a)+(-0.32,-0.47)$) -- ++(-0.05,-0.22) node[hist,below,minimum width=1.8mm,minimum height=1.8mm](g2){};
\foreach \dx in {-0.15,0.15}{\draw (g2) -- ++(\dx,-0.22) node[ans,below,minimum size=1.2mm](g3){};
  \draw (g3) -- ++(0,-0.14) node[term,below,minimum size=1.3mm]{};}
\node at ($(h-2-2-a)+(0.2,-0.45)$) {$\cdots$};
\foreach \j in {1,2,3}{\foreach \s in {a,b}{\draw[->,chain] (h-3-\j-\s) -- ++(0,-0.42) node[term,below]{};}}
\foreach \k/\txt in {0/{looks 1 question ahead,\\then predicts},1/{2 questions ahead,\\then predicts},2/{3 questions ahead,\\then predicts},3/{1 question, then the D1\\policy until the budget ends}}{
  \node[note] at (\k*3.45+1.55,-1.15) {\txt};}
\node[note,anchor=north west,text width=13.6cm,align=center] at (-0.1,-1.55) {%
\tikz\node[term]{}; value of predicting or abstaining now \qquad
\tikz\draw[chain,->](0,0)--(0.4,0); exact value of continuing a fixed policy\\[1pt]
Lookahead depth $k$ limits how far one search looks. After every answer the agent searches again, so D2 with a budget of four can still ask four questions.};
\begin{scope}[yshift=-4.42cm]
\node[rowlabel] at (-0.1,2.05) {One-step latent scores: a single chain instead of a full tree};
\begin{scope}[xshift=0cm]
  \node[title] at (0,1.6) {(e) LIG};
  \node[hist] (r) at (0.45,1.15) {$h$};
  \foreach \y/\w/\j in {1.35/0.30/1,1.17/0.80/2,0.99/0.50/3}{
    \fill[gray!40] (1.3,\y-0.05) rectangle ++(\w,0.11);
    \node[anchor=east,font=\tiny,inner sep=1pt] at (1.28,\y) {$j_{\j}$};}
  \fill[blue!55] (1.3,1.12) rectangle ++(0.80,0.11);
  \node[font=\tiny,anchor=west] at (1.05,0.76) {score $I(Z;O_j\mid h)$};
  \node[ans] (c) at (0.45,0.55) {};
  \draw[->,pick] (r) -- node[left,inner sep=1pt]{$j_2$} (c);
  \node[hist] (h1) at (0.15,0.0) {}; \node[hist] (h2) at (0.75,0.0) {};
  \draw (c) -- (h1); \draw (c) -- (h2);
  \draw[->,pick] (h1) -- ++(0,-0.35) node[ans,below]{};
  \draw[->,pick] (h2) -- ++(0,-0.35) node[ans,below]{};
  \node[note] at (1.55,-1.0) {ask the most informative\\question until the budget ends};
\end{scope}
\begin{scope}[xshift=3.45cm]
  \node[title] at (0,1.6) {(f) LIG-stop};
  \node[hist] (r) at (0.7,1.15) {$h$};
  \node[font=\tiny,anchor=west,align=left] at (1.05,1.15) {continue only if\\$W(h)>T(h)$};
  \node[ans] (c) at (0.7,0.55) {};
  \draw[->,pick] (r) -- node[left,inner sep=1pt]{$j_2$} (c);
  \node[hist] (h1) at (0.35,0.0) {}; \node[hist] (h2) at (1.05,0.0) {};
  \draw (c) -- (h1); \draw (c) -- (h2);
  \draw[->,pick] (h1) -- ++(0,-0.35) node[ans,below]{};
  \draw (h2) -- ++(0.35,-0.3) node[term,below right=-1pt](t){};
  \node[font=\tiny,anchor=west] at ($(t)+(0.08,0)$) {stop, predict};
  \node[note] at (1.55,-1.0) {the same sequence, stopping\\when continuing is not worth it};
\end{scope}
\begin{scope}[xshift=6.9cm]
  \node[title] at (0,1.6) {(g) Rollout-LIG};
  \node[hist] (r) at (1.55,1.15) {$h$};
  \foreach \x/\j in {0.45/1,1.55/2,2.65/3}{
    \node[ans] (c\j) at (\x,0.6) {};
    \draw[->] (r) -- node[fill=white,inner sep=0.4pt,font=\tiny]{$j_{\j}$} (c\j);
    \node[hist,minimum width=2.6mm] (h\j a) at ($(c\j)+(-0.22,-0.48)$) {};
    \node[hist,minimum width=2.6mm] (h\j b) at ($(c\j)+(0.22,-0.48)$) {};
    \draw (c\j) -- (h\j a); \draw (c\j) -- (h\j b);
    \draw[->,chain] (h\j a) -- ++(0,-0.42) node[term,below]{};
    \draw[->,chain] (h\j b) -- ++(0,-0.42) node[term,below]{};}
  \node[note] at (1.55,-1.0) {score every first question by the\\LIG-stop value $W$ after each answer};
\end{scope}
\node[note,anchor=west,text width=3.1cm,align=left] at (10.45,0.55) {Like rollout (d), but the continuation follows LIG and may stop. Its chains have one question per node, so its cost is about $n|O|^r$, against $(n|O|)^r$ for full depth.};
\end{scope}
\begin{scope}[yshift=-7.05cm]
\node[rowlabel] at (-0.1,1.05) {(h) The same compute limit, two ways to search for D3};
\draw[densely dashed, red!70!black, line width=0.8pt] (9.3,0.8) node[above,font=\tiny,text=red!70!black]{compute limit} -- (9.3,-0.75);
\node[anchor=east,font=\tiny,align=right] at (1.75,0.35) {original\\bounded D3};
\fill[green!25] (1.85,0.2) rectangle (3.05,0.5); \node[font=\tiny] at (2.45,0.35) {D1 fallback \checkmark};
\fill[gray!20] (3.05,0.2) rectangle (9.3,0.5);
\node[font=\tiny] at (6.15,0.35) {depth-3 candidates in greedy order; none completes};
\draw[->] (9.4,0.35) -- (10.0,0.35) node[right,font=\tiny,align=left]{returns the D1 action};
\node[anchor=east,font=\tiny,align=right] at (1.75,-0.45) {iterative\\deepening};
\fill[green!25] (1.85,-0.6) rectangle (2.65,-0.3); \node[font=\tiny] at (2.25,-0.45) {D1 \checkmark};
\fill[green!40] (2.65,-0.6) rectangle (4.75,-0.3); \node[font=\tiny] at (3.7,-0.45) {D2 \checkmark\ (incumbent)};
\fill[blue!12] (4.75,-0.6) rectangle (9.3,-0.3);
\foreach \x/\lab in {4.75/inc.,5.55/ctx,6.35/$j_4$,7.15/$j_1$,7.95/$\cdots$}{
  \draw[blue!50] (\x,-0.6) rectangle ++(0.8,0.3); \node[font=\tiny] at (\x+0.4,-0.45) {\lab};}
\draw[->] (9.4,-0.45) -- (10.0,-0.45) node[right,font=\tiny,align=left]{keeps the completed D2\\decision unless a completed\\D3 candidate beats it};
\node[font=\tiny,anchor=west,align=left,text=black!75] at (1.85,-1.05) {Depth-3 pass: the incumbent first, then candidates by LIG, so the context (ctx) comes early;\\ordered by depth-2 value, the context would come last, because it is worth little two steps ahead.};
\end{scope}
\end{tikzpicture}
\caption{How the planners search, at a history $h$ with three unasked questions. Squares are histories, circles a question's uncertain answer. (a)--(c) D$k$ branches on every unasked question for $k$ levels, then values predicting now, $T(h)$. (d) Rollout looks one question ahead, then follows D1 to the end of the budget. (e) LIG asks the question most informative about the latent cause $Z$. (f) LIG-stop also stops when $W(h)$, the exact value of continuing, does not exceed $T(h)$. (g) Rollout-LIG scores each first question by the LIG-stop value after its answers. (h) Under a compute limit, the original bounded D3 fell back to D1 whenever no depth-three candidate finished; iterative deepening keeps the completed D2 decision.}\label{fig:methods}
\end{figure}
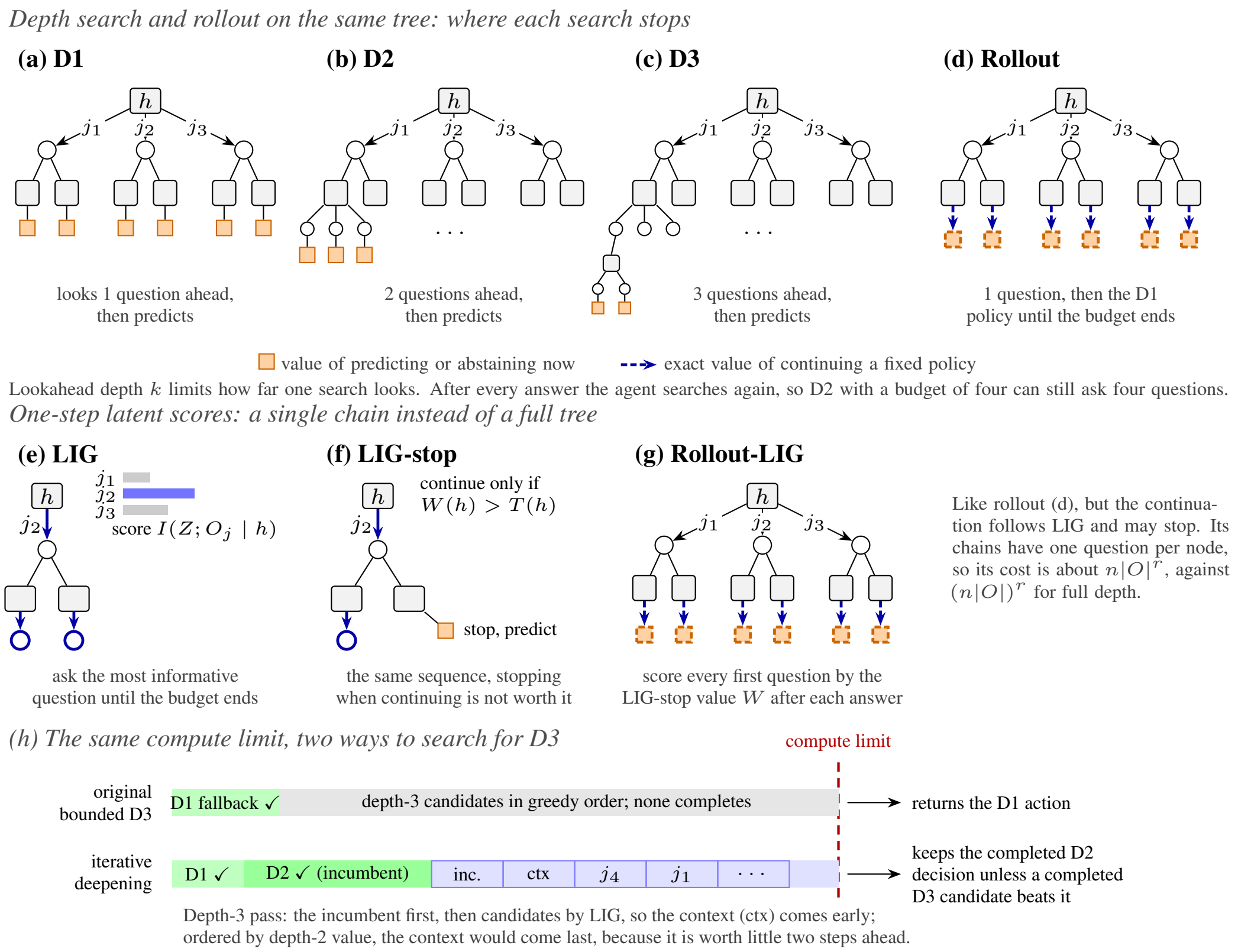

Every planner answers the same question at a history $h$: which question to ask next, or whether to stop. They differ in what they compute for a candidate $j$ (Figure~\ref{fig:methods}).

\paragraph{Depth search values combinations.} Exact $k$-step lookahead, D$k$, uses the Bellman value
\begin{equation}
 V_0(h)=T(h),\qquad V_d(h)=\max\Bigl\{T(h),\ \max_j\Bigl[-c+\sum_o p(o\mid h,j)\,V_{d-1}\bigl(h\oplus(j,o)\bigr)\Bigr]\Bigr\},
 \label{eq:depth}
\end{equation}
asks the maximizing question at depth $k$, observes the answer and searches again. It sees the value of any combination of up to $k$ answers, but branches on every unasked question at every level. Its cost therefore grows roughly as $(n\,|O|)^k$ for $n$ questions with $|O|$ answers each, and it relies on the fitted model at every node. D1 is the one-step value of information (Figure~\ref{fig:methods}a--c).

\paragraph{One-step scores value a single answer.} Conditional mutual information, $\mathrm{CMI}(j)=I(Y;O_j\mid h)$, measures what the next answer reveals about the label. Latent information gain, $\mathrm{LIG}(j)=I(Z;O_j\mid h)$, measures what it reveals about the fitted latent cause $Z$, a mixture component such as a class--context pair. Both cost one pass over the candidates. They differ only in the target. A context feature says nothing about the label on its own, so its CMI is zero. It does say which component generated the case, so its LIG is positive. On its own, LIG asks its best question until the budget ends.

\paragraph{Stopping and rollout make a one-step score a planner.} LIG-stop follows LIG's sequence, $j^*(h)=\arg\max_j \mathrm{LIG}(j)$, and stops when continuing is not worth it. Its value with $r$ questions left is exact:
\begin{equation}
 W_0(h)=T(h),\qquad W_r(h)=\max\Bigl\{T(h),\ -c+\sum_o p(o\mid h,j^*)\,W_{r-1}\bigl(h\oplus(j^*,o)\bigr)\Bigr\}.
\end{equation}
Rollout-LIG scores every first question $j$ by $-c+\sum_o p(o\mid h,j)\,W_{r-1}(h\oplus(j,o))$, asks the best if it beats $T(h)$, and replans. It looks ahead to the end of the budget, but along one question per node, so its cost is about $n\,|O|^{r}$ rather than $(n\,|O|)^{r}$. The same construction with a D1 continuation is ordinary one-step rollout, which we call rollout (Figure~\ref{fig:methods}d--g).

\paragraph{Baselines.} On the same fitted model, UCT (Upper Confidence bounds applied to Trees) samples answer paths instead of enumerating them, and ECED scores a question by the weight it removes from pairs of hypotheses that predict different labels. DIME, GDFS and SEFA train on data and choose one feature at a time. AACO scores whole subsets of unasked features on the nearest training cases, then acquires the subset's single most useful member. Every comparison uses the same data, query budget and scoring rule. Search budgets count candidate--answer transitions, which are logical units rather than floating-point operations. Appendix~\ref{sec:confirmation-methods} gives all settings.

\section{What depth can and cannot buy}\label{sec:main-theory}
Each result is a Lean~4.19 theorem (Appendix~\ref{app:lean-proofs}); finite tests, not proofs, link them to the planners.

\paragraph{A label score can be blind to an interaction.} \emph{Problem:} a one-step score sees only what one answer reveals. \emph{Theorem~\ref{thm:interaction-order}:} in any finite joint with signed utility and abstention, if the label's posterior equals its prior after every history of fewer than $k$ answers, then at a history of size $s$ every lookahead of $n$ questions with $s+n<k$ is worth exactly stopping, and CMI is zero everywhere. Conversely, if some fixed set of $k$ questions beats stopping, every lookahead of depth at least $k$ strictly beats it (Theorem~\ref{thm:interaction-sufficiency}). For the parity of three noisy bits, D2 is worth zero and D3 is not (Theorem~\ref{thm:parity}).

\paragraph{A latent score is never blind.} \emph{Problem:} the planners of Section~\ref{sec:main-methods} score the latent cause, not the label. \emph{Theorem~\ref{thm:visibility}:} if answers are independent given a finite latent cause that determines the label, a fresh question leaves the posterior over the cause unchanged exactly when its answer laws agree on the posterior's support; if no single question changes it, no sequence does, and every lookahead is worth exactly stopping. So if anything is worth asking, some single question has positive LIG. The guarantee is visibility, not ranking: with the whole feature vector as the latent, a distractor moves it.

\paragraph{A latent score depends on its description; exact values do not.} \emph{Problem:} one observable problem admits many latent causes. \emph{Theorem~\ref{thm:refinement}:} moving a question's answer noise into the cause leaves the joint over answers and label, so exact lookahead and any stop-or-query rule's exact value, unchanged. \emph{Theorem~\ref{thm:gap}:} for two fair bits with an XOR label, a four-valued nuisance question, budget two and any cost $0\leq c\leq1/2$, LIG-stop and Rollout-LIG earn $1-2c$ when the cause is the pair of bits and $0$ when it also records the nuisance, however ties break; D2 earns $1-2c$ under both. So visibility plus rollout gives no uniform guarantee, and a check built from exact values cannot be defeated this way.

\paragraph{Model error along a policy's paths.} \emph{Problem:} a bound on the joint's error is loose. \emph{Theorem~\ref{thm:on-policy}:} a policy's true and fitted values differ by at most the reach-weighted $L_1$ error of the answer laws along its histories, plus terminal errors at its stops. A fitted margin above two policies' errors certifies the true preference; with error radii from validation data, the bound needs only the fitted model (Theorem~\ref{thm:robust}). Exact value cannot decrease with depth, but executed replanning can (Theorem~\ref{thm:replanning}).
\section{Tasks}\label{sec:main-datasets}
\paragraph{Context parity: an interaction no one-step score can see.} A fair context bit $C$ selects one of two blocks of $k$ fair bits, and the label is that block's parity, flipped with probability $\epsilon$. CMI is zero for every first question. The right plan asks $C$, then the $k$ bits of its block, worth $1-2\epsilon-(k+1)c$; the best fixed set is worth only $\max\{0,1/2-\epsilon-kc\}$ (Theorem~\ref{thm:adaptivity}). We fit a saturated joint on 120 or 600 samples, with $k\in\{2,3\}$, $\epsilon\in\{0,.1\}$ and $c=.03$; the latent cause is then the feature vector.

\paragraph{CUBE-NM: the context says where to look.} In AFABench's CUBE-NM \citep{afabench2026}, a class $y\in\{0,\dots,7\}$ and a context $z$ are drawn. The context selects a block of ten coordinates, in which three binary digits of $y$ sit at label-dependent positions. The \emph{noisy} version has 16 settings of contexts, sample size, cost, noise and budget, with signed utility. The \emph{noiseless} version has five contexts, 51 feature groups and a budget of 3, 5 or 7 questions (B3, B5, B7), scored by a classifier trained by the benchmark's recipe. Planners predict the fitted model's most probable class or, in the \emph{aligned} variant, value each leaf by that classifier's expected correctness. The fitted latent is the class--context pair.

\paragraph{Common 20 Questions: answers from language models.} The agent identifies one of 111 items, such as Otter or Calendar \citep{hu2024uot}. Qwen and GLM each write a bank of 32 yes/no questions and answer each bank for each item. Six requests per item fit a Beta--Bernoulli answer model, and three held-out ones test the frozen policies. The latent cause is the item, so LIG equals CMI.

\paragraph{Nine real datasets.} Iris, wine, breast cancer, digits, two diabetes sets, PhysioNet, MiniBooNE and zoo give 27 tasks of at most ten training-median binary features, each with five 70/30 splits. A latent-class model is fitted per class, with its size chosen by the Bayesian information criterion (BIC). Planners are scored on held-out rows, with budget three and cost $.01$.

\paragraph{Protocol.} Every study froze its methods, contrasts and expected outcome with source hashes before any result, and intervals are Bonferroni-adjusted within each prespecified family.
\section{Results}\label{sec:main-results}
\Needspace{4\baselineskip}
\subsection{Can a one-step score replace depth?}\label{sec:main-decision-aware}
\begin{figure}[t]
\centering
\includegraphics[width=\linewidth]{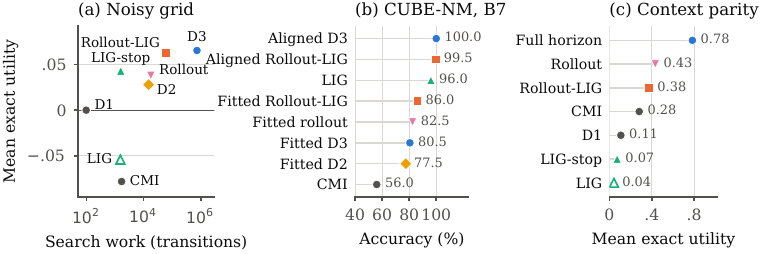}
\caption{One-step latent scores against depth. (a) Utility against compilation work, 160 fits; hollow: LIG. (b) Accuracy at seven questions, five fits. \emph{Aligned} planners target the scored classifier, \emph{fitted} ones the fitted model's most probable class, LIG neither. (c) Context parity, 80 fits, saturated latent.}\label{fig:lig-depth}
\end{figure}

\emph{Problem.} Depth pays in designed tasks: in context parity the full horizon gains $.442$ over one step less ($[.435,.449]$), and across 16 noisy CUBE-NM settings D3 beats D2 by $.0373$ ($[.0323,.0423]$). But it is expensive, and a label score such as CMI is blind to these interactions. \emph{Solution.} Score the latent cause instead, and add exact stopping and rollout (LIG-stop and Rollout-LIG). \emph{Lean.} Theorems~\ref{thm:interaction-order} and~\ref{thm:visibility} show why label scores fail and LIG does not, and Theorem~\ref{thm:rollout} that Rollout-LIG is at least LIG-stop under the fit. None promises a good ranking. \emph{Simulation} (Figure~\ref{fig:lig-depth}, Appendix~\ref{app:decision-aware}). On the noisy grid Rollout-LIG trails exact D3 by $.0031$ ($[-.0080,.0018]$) with about a twelfth of its search. LIG alone, which spends its budget as CMI does, is below zero. On noiseless CUBE-NM at B7, LIG asks for the context first in every fit, at about 1/6,600 of the work of aligned D3. Scored by the other four classifiers, aligned D3 reaches 88.5\% and LIG 95.9\% ($+7.4$, $[4.3,10.4]$, post hoc). D3 planned for its fitted terminal is 15.5 points below LIG ($[10.6,20.4]$): that lead reflects the terminal, not the search. A generic latent-class fit recovers the class--context structure when its size is chosen by cross-validated likelihood (Rollout-LIG $-$ D3: $-.0035$, $[-.0125,.0055]$); with BIC it collapses to one component per class and every planner falls to 41--47\% (Appendix~\ref{app:generic-latent}). \emph{Verdict.} Confirmed where the fitted latent encodes the decision structure. Rejected on context parity, where the fitted latent is the whole feature vector. The next result asks why.

\Needspace{4\baselineskip}
\subsection{Why does the latent score fail on parity?}\label{sec:main-representation}
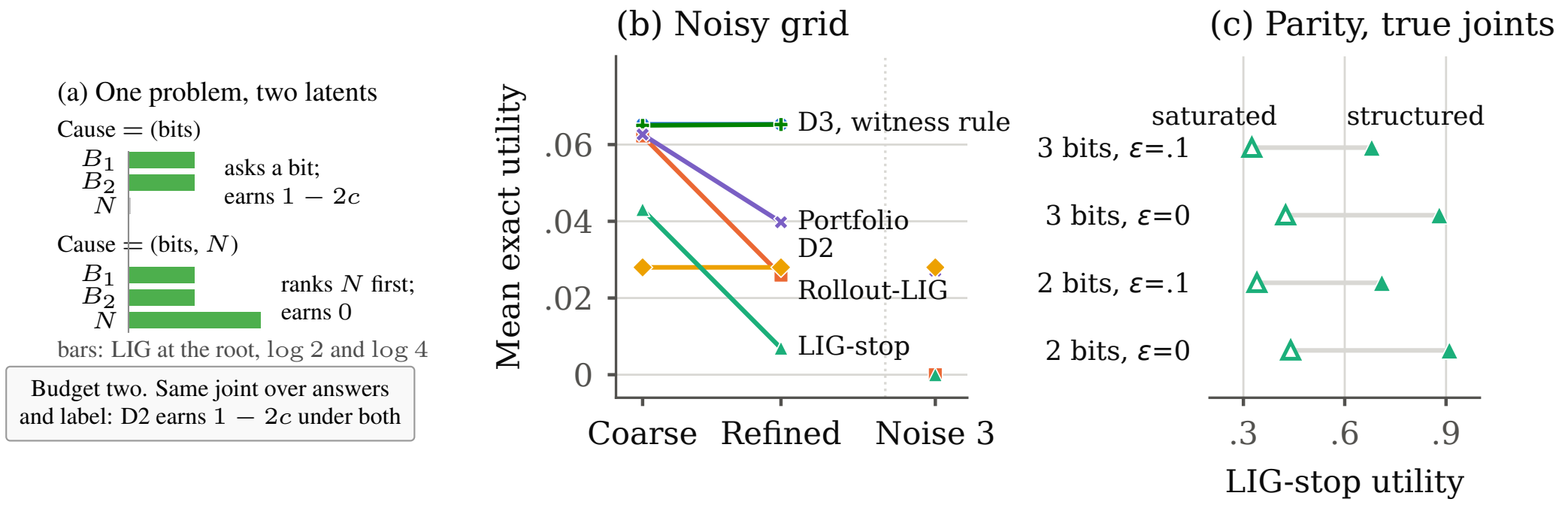
\begin{figure}[t]
\centering
\begin{minipage}[c]{0.315\linewidth}
\centering
\begin{tikzpicture}[font=\tiny, >={Stealth[length=1.2mm]},
  bar/.style={fill=green!55!black!70, draw=none},
  low/.style={fill=gray!45, draw=none}]
\node[font=\scriptsize, anchor=west] at (-.75,2.46) {(a) One problem, two latents};
\node[anchor=west] at (-.75,2.14) {Cause $=$ (bits)};
\foreach \y/\lab/\w/\st in {1.88/$B_1$/.58/bar, 1.68/$B_2$/.58/bar, 1.48/$N$/.02/low}{
  \node[anchor=east, inner sep=1pt] at (-.05,\y) {\lab};
  \fill[\st] (0,\y-.07) rectangle ++(\w,.14);}
\node[anchor=west, align=left] at (.72,1.68) {asks a bit;\\earns $1-2c$};
\node[anchor=west] at (-.75,1.13) {Cause $=$ (bits, $N$)};
\foreach \y/\lab/\w/\st in {.87/$B_1$/.58/bar, .67/$B_2$/.58/bar, .47/$N$/1.16/bar}{
  \node[anchor=east, inner sep=1pt] at (-.05,\y) {\lab};
  \fill[\st] (0,\y-.07) rectangle ++(\w,.14);}
\node[anchor=west, align=left] at (1.22,.67) {ranks $N$ first;\\earns $0$};
\draw[black!45] (0,.36) -- (0,1.96);
\node[anchor=west, text=black!70, align=left] at (-.75,.2) {bars: LIG at the root, $\log2$ and $\log4$};
\node[draw=black!40, rounded corners=1pt, fill=gray!5, align=center, anchor=north] at (.72,.07)
  {Budget two. Same joint over answers\\and label: D2 earns $1-2c$ under both};
\end{tikzpicture}
\end{minipage}\hfill
\begin{minipage}[c]{0.675\linewidth}
\centering
\includegraphics[width=\linewidth]{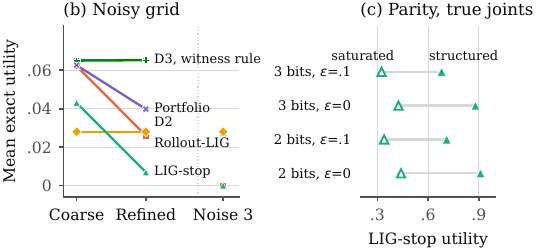}
\end{minipage}
\caption{The description decides. (a) Theorem~\ref{thm:gap}: two fair bits with an XOR label and a four-valued nuisance $N$. (b) Noisy grid, 160 fits. \emph{Coarse} adds an eight-answer question the latent ignores and \emph{refined} records it; \emph{noise 3}, a separate condition, records the noise of three existing questions (D3 and the witness rule not run). (c) LIG-stop on the true parity joints; the \emph{structured} latent keeps the context, selected block and label, and equals full-horizon search.}\label{fig:representation}
\end{figure}

\emph{Problem.} Visibility is not ranking. One observable problem has many latent descriptions, and parity's fitted one, the whole feature vector, scores every question about alike. Is the score at fault, or the description? \emph{Solution.} Keep only the decision structure in the latent where it is known. Otherwise, check LIG against exact values, which no description can change: the witness rule (Appendix~\ref{app:order-rule}), or a cheap portfolio that takes D2 when its root value beats Rollout-LIG's. \emph{Lean.} Moving a question's noise into the cause leaves every exact value unchanged (Theorem~\ref{thm:refinement}), yet in a two-bit family a LIG rule falls from $1-2c$ to $0$ at every cost below $1/2$ (Theorem~\ref{thm:gap}, Figure~\ref{fig:representation}a). \emph{Simulation} (Figure~\ref{fig:representation}b,c; Appendix~\ref{app:representation}). Having the latent record a new eight-answer nuisance question changes root values and witnesses by at most $2\times10^{-15}$ on all 295 fits, yet costs noisy-grid Rollout-LIG $.0363$ ($[.0297,.0429]$); on real data the loss is $.0146$, an interval that includes zero. Recording the noise of the three most uncertain existing questions takes both LIG planners to exactly 0 on all 160 noisy fits. On the true parity joints, the latent (context, selected block, label) describes the same problem, and with it LIG-stop equals full-horizon search at a fifth to a nineteenth of its work. Part of Figure~\ref{fig:lig-depth}c's failure is the fit, not the description: on true joints the saturated latent gives LIG-stop .38 on average, against .07 fitted, and Rollout-LIG .38 either way. The witness rule is unaffected by refinement, at 1.58 times D3's work on the grid. The portfolio costs 14\% of D3's work and does not help on parity, where D2 is too shallow. Neither check differs from Rollout-LIG on real data. Against our expectation, a cross-validated fit given a nuisance block sharing a hidden state never asked the block first in 160 fits. \emph{Verdict.} Confirmed: the description, not only the score, decides LIG's outcome. The witness rule recovers the loss at a cost, the portfolio only part of it. The risk lies in hand-built or oversized latents, not the cross-validated fits we tested.

\Needspace{4\baselineskip}
\subsection{Do learned and non-myopic baselines do better?}
\emph{Problem.} The planners might win only because their fitted model mirrors the generator, or because they plan further ahead. \emph{Solution.} Compare with DIME, GDFS and SEFA, which fit no such model, from pinned author code, and with the non-myopic AACO, adapted to reproduce every action of the authors' code. \emph{Lean.} No proof covers these policies. \emph{Simulation} (Figure~\ref{fig:baselines-budget}a; Appendices~\ref{app:learned-baselines} and~\ref{app:aaco}). On CUBE-NM all baselines fall far below LIG. AACO never asks for the context first: it takes the member of its best subset that helps most alone. On the noisy grid AACO trails D3 by $.0180$ ($[.0113,.0247]$); on the real datasets its interval with D3 and Rollout-LIG includes zero. \emph{Verdict.} Rejected on CUBE-NM, and for AACO on the noisy grid; no difference detected on real data.

\Needspace{4\baselineskip}
\subsection{Does depth survive a limited search budget?}
\begin{figure}[t]
\centering
\includegraphics[width=\linewidth]{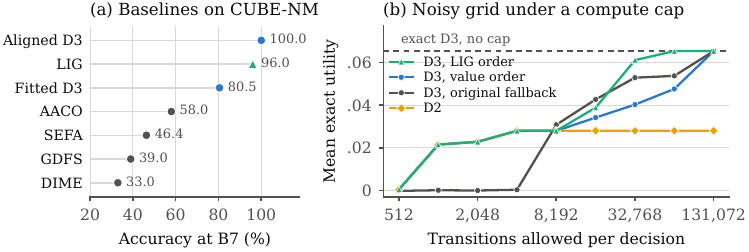}
\caption{(a) Noiseless CUBE-NM accuracy at seven questions, on the same five fits; AACO predicts with the fitted model (56\% with the scored classifier). (b) Noisy-grid utility, 160 fits. Up to 8,192 transitions both deepening orders equal D2.}\label{fig:baselines-budget}
\end{figure}

\emph{Problem.} A search capped at a fixed number of transitions per decision can lose to a shallower one. Our original bounded D3 returned the D1 action whenever no depth-three candidate completed; at 4,096 transitions it lost to D2 by $.0276$ ($[.0244,.0308]$) on the noisy grid and by 38 points on aligned CUBE-NM. \emph{Solution.} Iterative deepening completes depth two first and keeps that decision as its fallback (Figure~\ref{fig:methods}h); depth-three candidates are ordered by depth-two value or, as specified in advance, by LIG. \emph{Lean.} An interrupted pass never discards a completed shallower decision, for any values and interruption pattern (Theorem~\ref{thm:anytime}). \emph{Simulation} (Figure~\ref{fig:baselines-budget}b; Appendix~\ref{app:anytime}). Deepened D3 never falls below D2. LIG ordering beats D2 by $.0331$ at 32,768 transitions ($[.0275,.0387]$) and reaches exact D3 at half the work value ordering needs. The original planner leads at 8,192--16,384 only because the catalogue lists the context first. At 251 CUBE-NM features, LIG reaches the exact optimum at every budget with at most 4{,}100 transitions per episode, where UCT falls to 38--76\% and DIME to 15--22\% (Appendix~\ref{app:scaling}). \emph{Verdict.} Confirmed: the losses came from the fallback.

\Needspace{4\baselineskip}
\subsection{Does it hold on real data and in the implemented controller?}
\begin{figure}[t]
\centering
\includegraphics[width=\linewidth]{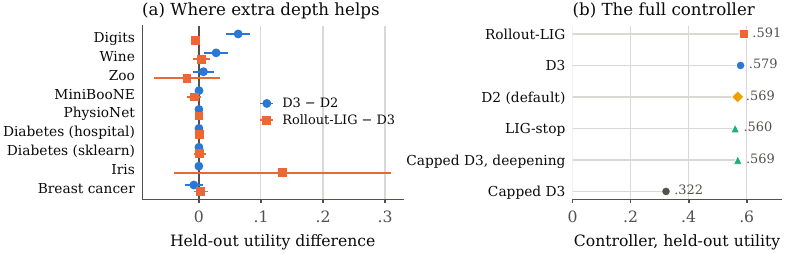}
\caption{(a) Held-out differences, with descriptive 95\% intervals over task--split pairs, which share data. On iris a poor fit misleads D2 and D3 alike. (b) The full controller, averaged over datasets; capped rows allow 128 units per decision.}\label{fig:real-controller}
\end{figure}

\emph{Problem.} Synthetic tasks are built to need depth, and the planners above are research code. \emph{Solution.} Repeat the comparison on the nine real datasets, then run the full controller of Section~\ref{sec:main-controller} on the same fits. \emph{Lean.} Theorem~\ref{thm:exec} proves an executable planner optimal; agreement with the Python planner is checked on finite fixtures, not proved. \emph{Simulation} (Figure~\ref{fig:real-controller}). Held-out D3 minus D2 is $+.0100$ ($[-.0125,.0324]$) overall, and the gain comes from digits and wine. The fitted depth gain predicts the held-out gain (Spearman $.61$; Appendix~\ref{app:tier2}). The controller's D2, D3, Rollout-LIG and LIG-stop match the research results on all 135 splits to $10^{-14}$. Rollout-LIG scores highest, with intervals that include zero, at 74 planning units per decision against 197 for D3. Under a tight cap, a depth-three gate without deepening falls on four of the nine datasets and latent-ordered deepening holds D2's level, though the average gain's interval includes zero ($+.2474$, $[-.0786,.5734]$; Appendix~\ref{app:controller-real}). \emph{Verdict.} Depth beyond two questions rarely pays: the overall interval includes zero, and the gain is concentrated in digits and, less, wine. The controller carries the planners' results over.

\Needspace{4\baselineskip}
\subsection{Do gains survive a change of answering model?}\label{sec:main-answering}
\begin{figure}[t]
\centering
\begin{minipage}[c]{0.325\linewidth}
\centering
\begin{tikzpicture}[font=\tiny, >={Stealth[length=1.2mm]},
  box/.style={draw=black!45, rounded corners=1pt, fill=gray!8, align=center, inner sep=2pt},
  fit/.style={draw=blue!55!black, rounded corners=1pt, fill=blue!7, align=center, inner sep=2pt},
  test/.style={draw=green!45!black, rounded corners=1pt, fill=green!10, align=center, inner sep=2pt}]
\node[font=\scriptsize, anchor=west] at (-1.62,1.42) {(a) Count; do not ask};
\node[box, text width=2.9cm] (q) at (0,1.0) {``Is an otter a living organism?''\\repeated across requests};
\node[fit, text width=1.3cm] (f) at (-.8,.3) {six answers:\\Y Y Y Y Y N};
\node[test, text width=1.3cm] (e) at (.8,.3) {three new:\\Y Y Y};
\node[fit, text width=1.3cm] (t) at (-.8,-.4) {$\theta=\dfrac{5+\frac12}{6+1}$};
\node[test, text width=1.3cm] (s) at (.8,-.4) {compare, from\\all answers};
\draw[->] (q.south-|f) -- (f); \draw[->] (q.south-|e) -- (e);
\draw[->] (f) -- (t); \draw[->] (e) -- (s); \draw[->] (t) -- (s);
\node[anchor=north, align=center, text=black!70] at (0,-.76) {No number is ever stated by a model.};
\end{tikzpicture}
\end{minipage}\hfill
\begin{minipage}[c]{0.655\linewidth}
\centering
\includegraphics[width=\linewidth]{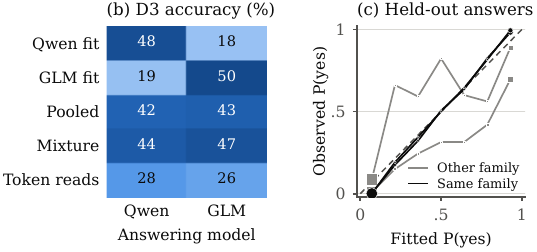}
\end{minipage}
\caption{Counted probabilities, tested on held-out answers. (a) A Beta--Bernoulli fit, with illustrative answers. (b) D3 identification on Common at seven questions (last row: Section~\ref{sec:main-reads}'s four-model reads, no answers counted). (c) Reliability of the fitted yes-probability; dashed: perfect. Marker area follows the answers in a bin: 90--97\% lie in the two extreme bins, where six answers agreed.}\label{fig:answering}
\end{figure}
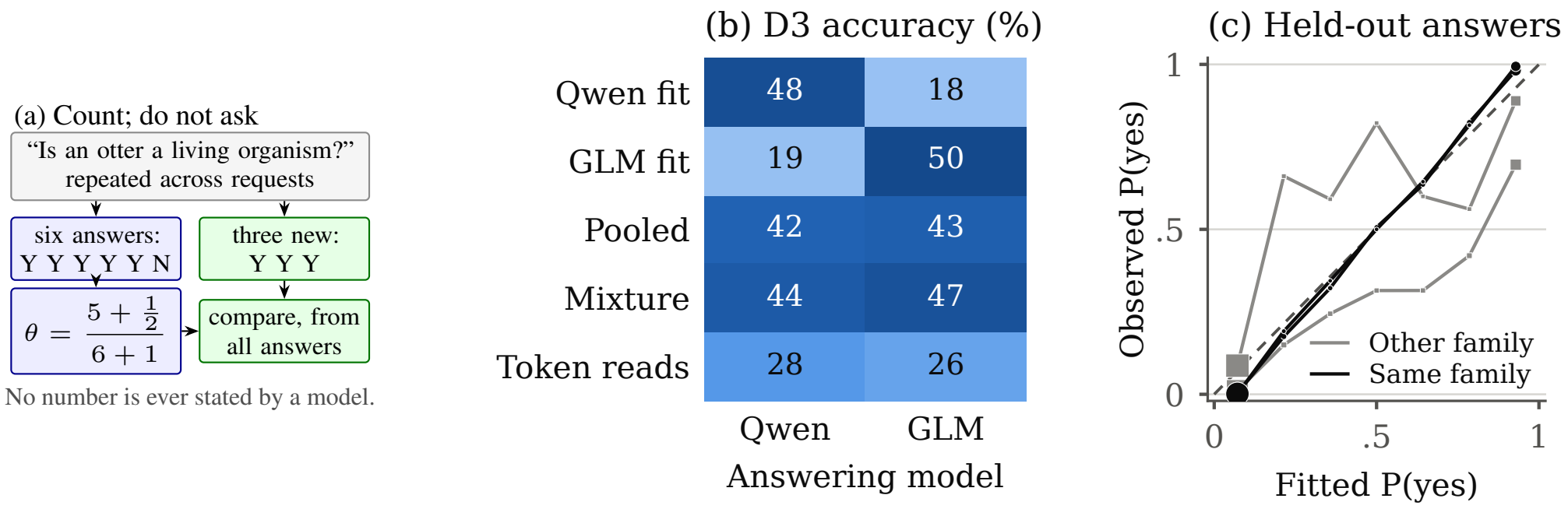

\emph{Problem.} An LLM produces answers, not trustworthy probabilities, and another model may answer differently. \emph{Solution.} Estimate each answer probability by counting repeated answers, test it on held-out answers (Figure~\ref{fig:answering}a), and plan over a mixture of the plausible answering models, with one component per item and family. \emph{Lean.} Theorem~\ref{thm:transfer} limits the regret by the distance between the models; no proof shows that the mixture recovers the loss. \emph{Simulation} (Figure~\ref{fig:answering}b,c; Appendices~\ref{app:families} and~\ref{app:powered-panels}). When all six fitting answers agree, a held-out answer from the same family contradicts them .2--.6\% of the time, and one from the other family 6.6--9.0\%. Cross-family log loss is 2.4--3.1 times higher, and D3 identifies only 18--19\% of items, against 48--50\% within a family. Inferring the family within a mixture recovers most of the loss. A rerun on four new banks and 188 items replicates each finding with a smaller depth gain: within a family D3 beats D2 by 3.7 ($[1.1,6.4]$, Qwen) and 4.5 points ($[0.8,8.3]$, GLM). \emph{Verdict.} The counts predict the model that produced them, not another: collapse and recovery are both confirmed. This does not test whether an answer is true.

\Needspace{4\baselineskip}
\subsection{Are probabilities read from an LLM real?}\label{sec:main-reads}
\emph{Problem.} With no data the controller reads an LLM's token log probabilities, and nothing guarantees them. \emph{Solution.} Use a read as a prior, with conformal sets once labels exist. \emph{Lean.} No proof covers the reads. \emph{Simulation} (Figure~\ref{fig:answering}b, Appendix~\ref{app:elicited}). On known facts the pooled read beats the base rate (log loss .30 against .57) but not class frequencies (.25; the interval includes zero). Stated percentages are worse (.51), at 0 or 100 on 89--99\% of those requests. Temperatures help single models there, not the pool. Plans on reads identify 14--31\% of items within a family, against 48--50\% from counts. Post hoc, the pool, which includes the answering family's reads, reaches 26--28\%, against 18--19\% from other-family counts, and with no labels reads are worth 2--10 rows per class. Mean conformal coverage is .88--.92 at nominal .90; poor reads pay with larger sets. \emph{Verdict.} The numbers can be measured, not guaranteed: a read is a prior worth a few labels, not an estimate. Five of ten frozen expectations held.

\Needspace{4\baselineskip}
\subsection{Can model error be certified?}
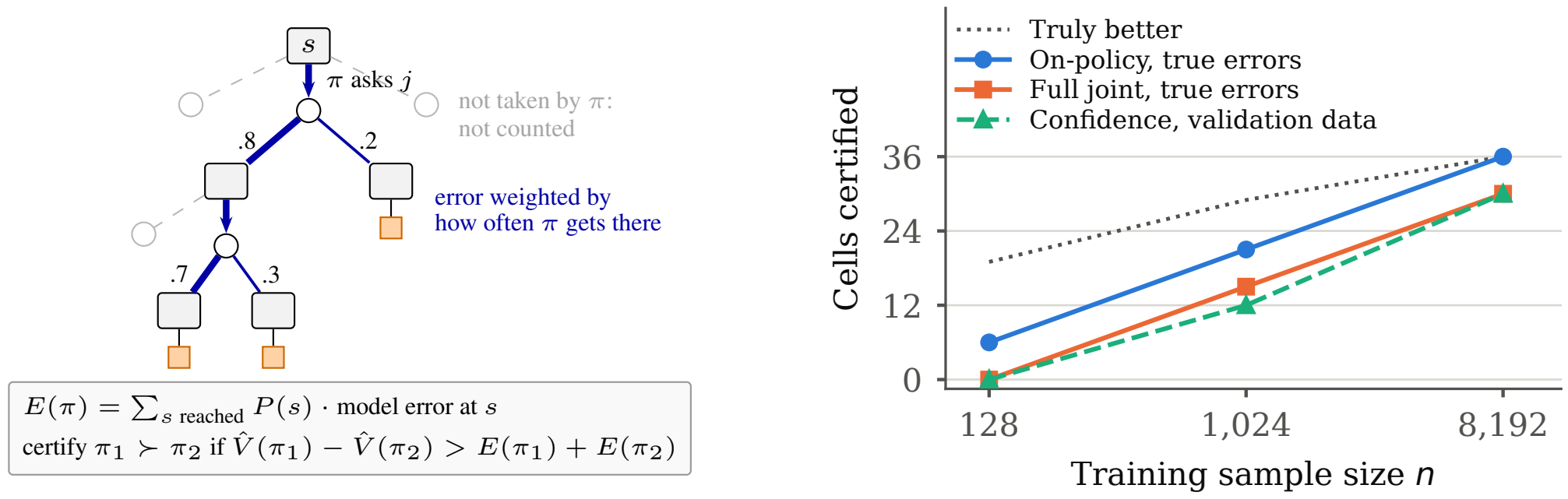
\begin{figure}[htbp]
\centering
\begin{minipage}[c]{0.52\linewidth}
\centering
\begin{tikzpicture}[font=\scriptsize, >={Stealth[length=1.4mm]},
  hist/.style={draw, rounded corners=1pt, minimum width=3.6mm, minimum height=3mm,
               inner sep=0.5pt, fill=gray!10},
  ans/.style={draw, circle, minimum size=2mm, inner sep=0pt, fill=white},
  stop/.style={draw=orange!80!black, fill=orange!35, minimum size=1.8mm, inner sep=0pt},
  hot/.style={draw=blue!65!black, line width=1.6pt},
  warm/.style={draw=blue!65!black, line width=0.7pt},
  off/.style={draw=gray!55, dashed}]
  \node[hist] (r) at (1.6,1.6) {$s$};
  \node[ans] (c) at (1.6,1.05) {};
  \draw[hot,->] (r) -- node[right,font=\tiny]{$\pi$ asks $j$} (c);
  \draw[off] (r) -- ++(-1.0,-0.5) node[ans,draw=gray!55,solid]{};
  \draw[off] (r) -- ++(1.0,-0.5) node[ans,draw=gray!55,solid]{};
  \node[hist] (a) at (0.9,0.45) {}; \node[hist] (b) at (2.3,0.45) {};
  \draw[hot] (c) -- node[left,font=\tiny]{.8} (a); \draw[warm] (c) -- node[right,font=\tiny]{.2} (b);
  \node[ans] (c2) at (0.9,-0.1) {};
  \draw[hot,->] (a) -- (c2);
  \draw[off] (a) -- ++(-0.7,-0.45) node[ans,draw=gray!55,solid]{};
  \node[hist] (a1) at (0.5,-0.65) {}; \node[hist] (a2) at (1.3,-0.65) {};
  \draw[hot] (c2) -- node[left,font=\tiny]{.7} (a1); \draw[warm] (c2) -- node[right,font=\tiny]{.3} (a2);
  \draw (a1) -- ++(0,-0.3) node[stop,below]{}; \draw (a2) -- ++(0,-0.3) node[stop,below]{};
  \draw (b) -- ++(0,-0.3) node[stop,below]{};
  \node[font=\tiny,text=gray!70,anchor=west] at (2.75,1.12) {not taken by $\pi$:};
  \node[font=\tiny,text=gray!70,anchor=west] at (2.75,0.9) {not counted};
  \node[font=\tiny,anchor=west,text=blue!65!black] at (2.55,0.3) {error weighted by};
  \node[font=\tiny,anchor=west,text=blue!65!black] at (2.55,0.08) {how often $\pi$ gets there};
  \node[draw=black!40, rounded corners=1pt, fill=gray!5, align=left, font=\tiny, anchor=north]
    at (1.95,-1.25) {$E(\pi)=\sum_{s\ \text{reached}}P(s)\cdot\text{model error at } s$\\[1pt]
    certify $\pi_1\succ\pi_2$ if $\hat V(\pi_1)-\hat V(\pi_2)>E(\pi_1)+E(\pi_2)$};
\end{tikzpicture}
\end{minipage}\hfill
\begin{minipage}[c]{0.45\linewidth}
\centering
\includegraphics[width=\linewidth]{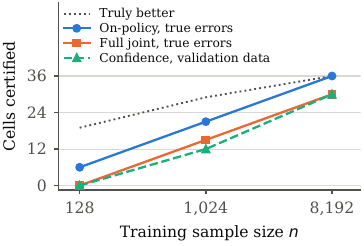}
\end{minipage}
\caption{Certifying a preference despite model error. Left: the on-policy error $E(\pi)$ counts model error only where the policy goes. Right: cells, of 36, where D3 over D2 is certified when training matches the source. Solid: true errors. Dashed: validation radii ($8n$ rows, $\delta=.05$), usable without the true model; $n$ rows certify 0, 0 and 23. Dotted: cells where D3 is truly better.}\label{fig:certificates}
\end{figure}

\emph{Problem.} A plan best under the fit can be worse under the truth. \emph{Solution.} Certify a preference with the on-policy bound, or choose by fitted value minus it. \emph{Lean.} Theorems~\ref{thm:on-policy}--\ref{thm:pessimistic}. \emph{Simulation} (Figure~\ref{fig:certificates}, Appendix~\ref{app:model-error}). With a known model, the bound certifies 452 of 772 fitted preferences when training matches the source, against 275 for a full-joint bound, and 132 of 565 when it does not, against 0, never with the wrong sign; its validation form needs more samples. Pessimistic choice helped under transfer but cost on correct small fits. \emph{Verdict.} The bound is confirmed. Pessimistic choice is rejected where complete validation rows exist: replaying candidates chose better (Appendix~\ref{app:pessimistic}).
\section{Discussion and limitations}\label{sec:discussion}
\paragraph{What we learned.} We began by asking when an agent must look several questions ahead. Depth clearly pays on tasks built to need it, but it costs exponentially and leans on the fit. The theorems located why a cheap score fails: a label score cannot see an interaction, and a latent score can. So we tested a one-step latent score against depth, and where the latent encoded the decision it nearly matched depth, even when a generic fit had to discover that latent. Parity broke it, and asking why changed the question. Described with a smaller latent, the same problem needed no depth; with a larger one, the score collapsed while exact values did not move. The issue was the description as much as the depth. Separate studies then probed its limits. Learned and non-myopic baselines did no better on the designed task. When a compute cap made depth harmful, the fault lay in the fallback. On real data depth rarely paid, and the full controller reproduced the planners. Finally, the probabilities themselves can fail: counts from one language model are accurate for it and confidently wrong for another, a mixture of families recovers most of the loss, token reads are worth only a few labels, and a bound built from one-step errors certifies some preferences despite model error.

\paragraph{When to pay for depth.} Depth pays when a one-step score cannot see the interaction, or ranks a distractor first. So keep the latent small, and check LIG against exact values when its description is in doubt. Under a compute budget, keep the completed shallower decision as the fallback and order candidates by LIG.

\paragraph{What the evidence does not show.} The synthetic tasks establish mechanisms, not how often interactions occur; the LLM results use fixed question banks, not open-ended dialogue. Parity's structured latent was given, not learned. Intervals that include zero are not equivalence tests. The controller's risk and safety layers are not evaluated as a whole. The evidence supports a conditional value-of-planning claim, not general superiority, deployment safety or state-of-the-art performance.

\paragraph{Broader impact.} An agent acting on fitted beliefs can intervene wrongly or withhold a needed action. Nothing here establishes that the controller is safe to deploy in financial or security settings.
 
\FloatBarrier
\clearpage
\begingroup\small
\bibliographystyle{plainnat}
\bibliography{learned_validation/references}
\endgroup
\clearpage
\appendix
\renewcommand{\thesection}{\AlphAlph{\value{section}}}
\renewcommand{\theHsection}{appendix.\AlphAlph{\value{section}}}
\section{Why the context--parity control distinguishes adaptivity}\label{app:context-parity}
Let $C\sim\mathrm{Bernoulli}(1/2)$, let $X_{zj}$ be independent uniform bits for $z\in\{0,1\}$ and $j\in\{1,\ldots,k\}$, and let $E\sim\mathrm{Bernoulli}(\epsilon)$ be independent label noise. The target is
\[
 Y=\left(\bigoplus_{j=1}^{k}X_{Cj}\right)\oplus E.
\]
All quantities are generated before acquisition. An action reveals $C$ or one bit, and the budget is $k+1$. This is a constructed diagnostic distribution, not an empirical model of the LLM panels or CUBE-NM.

An adaptive policy can acquire $C$ and the $k$ bits of the selected block. Its expected signed utility is
\[
 U_{\rm context}=1-2\epsilon-(k+1)c.
\]
For $k\ge2$, a nonadaptive subset of at most $k+1$ features cannot reveal both complete blocks. If it reveals neither complete block, the target remains uniform, even if context is observed. Revealing one complete block without context gives posterior correctness $3/4-\epsilon/2$, and hence signed utility $1/2-\epsilon-kc$. Also revealing context allows correct prediction on the matching half of cases and abstention otherwise, but has the same expected reward and one extra query cost. Partial bits from the other block do not reveal its parity. Therefore the optimal nonadaptive utility is
\[
 U_{\rm subset}^*=\max\{0,\,1/2-\epsilon-kc\}.
\]
At the tested $c=.03$, $k\in\{2,3\}$ and $\epsilon\in\{0,.1\}$, the explicit adaptive policy exceeds this nonadaptive optimum by $1/2-\epsilon-c$, namely $.47$ or $.37$. Exact oracle dynamic programming independently evaluates the optimal adaptive policy in these cells. Learned-policy comparisons then test whether sampled-reference models recover this opportunity. The family establishes a specific mechanism and a controlled distinction between adaptive and fixed-subset acquisition; it does not establish that similar interactions dominate application datasets.

With only $k$ steps of root lookahead, a context-first branch has at most $k-1$ subsequent bits, so it cannot reveal the selected parity. Its signed terminal value remains zero and its first query incurs a cost. A horizon of $k+1$ can value the complete context-dependent sequence. This explains why increasing the required block length shifts the useful planning horizon in this constructed family.
\section{Theorems checked in Lean}\label{app:lean-proofs}
Every theorem below is machine-checked in Lean 4.19 with Mathlib, with only the standard axioms, in exact rational arithmetic apart from the real logarithms of Theorem~\ref{thm:gap}. The theorems apply to the stated mathematical models; they do not verify the Python controller, its risk shaping or LLM likelihoods. Table~\ref{tab:lean-map} says where each is checked.

\emph{Common setting.} Unless stated otherwise, the model is a \emph{finite joint}: finitely many worlds with nonnegative rational masses, features with finite values and a finite label. A query reveals a fresh feature at cost $c\geq0$. A correct prediction scores $+1$, a wrong one $-1$ and abstention $0$. Values are kept in unnormalized mass, so no history needs positive support.

\paragraph{Depth on parity.}
\begin{theorem}[Parity]\label{thm:parity}
Let three independent fair bits have parity as the label, flipped independently with probability $q$; let each relevant answer flip independently with probability $p_i$, and let any finite number of fair distractors be independent, with all rates rational in $[0,1]$. Put $r=(1-2q)\prod_{i=0}^{2}(1-2p_i)$. At every history and finite depth, the Bellman value of the compressed count of unobserved bits equals the explicit-history value and bounds every admissible adaptive policy. From the initial history D2 is worth $0$ and D3 is worth $\max\{0,|r|-3c\}$, so D3 strictly beats D2 exactly when $|r|>3c$. Without noise D3 is worth $1-3c$ for $0\leq c\leq1/3$.
\end{theorem}
The count preserves value, not the label to predict. The theorem assumes exact known channels and no repeated query.

\begin{theorem}[Adaptivity]\label{thm:adaptivity}
For a fair context and two independent fair $k$-bit blocks, $k\geq2$, let the label be the selected block's parity, flipped independently with probability $0\leq\epsilon\leq1/2$. With budget $k+1$, the best fixed subset with a bounded terminal decision is worth $\max\{0,1/2-\epsilon-kc\}$, and finite mixtures of fixed subsets obey the same bound. Asking the context first is worth $1-2\epsilon-(k+1)c$.
\end{theorem}
This compares adaptive with fixed acquisition; it is not a D3-against-D2 or rollout theorem.

\paragraph{When a label score is blind.}
\begin{theorem}[Interaction order]\label{thm:interaction-order}
In a finite joint, suppose that after every history of fewer than $k$ observations the label mass is a fixed law $\pi$ times the history mass. Then the optimal replanning value with $n$ further queries equals stopping at every history of size $s$ with $s+n<k$. So every lookahead shorter than $k$ is worth exactly stopping at the root, and every posterior computable from fewer than $k$ observations equals $\pi$.
\end{theorem}
The statement places no restriction on noise, distractors or dependence among features.

\begin{theorem}[Sufficiency]\label{thm:interaction-sufficiency}
In a finite joint, a policy that asks a fixed list of features and then predicts is never worth more than optimal replanning with at least as many queries. So if some fixed set of $k$ queries beats stopping, every lookahead of depth at least $k$ strictly beats stopping.
\end{theorem}
Under the hypothesis of Theorem~\ref{thm:interaction-order}, depth $k$ is then the shortest depth with value. The converse fails: in context parity an adaptive policy beats stopping when no fixed set does.

\paragraph{Rollout and replanning.}
\begin{theorem}[Rollout]\label{thm:rollout}
In a finite Bellman model with a stop option, let $\ell$-step lookahead score its leaves with a base value that is sequentially improving, meaning attainable by stopping or by one query followed by the base. Then the estimates are nondecreasing in $\ell$; the policy that replans with them after every answer attains at least its own estimate; and it is never worse than the base at any state and budget. The exact value of any stop-or-query policy is sequentially improving, so rollout on any base policy never loses to it under the shared exact model.
\end{theorem}
Hard-budget protocols, approximate values and interrupted search are outside the statement.

\begin{theorem}[Deeper replanning can lose]\label{thm:replanning}
There is a static model with five binary features, 12 positive-weight atoms with deterministic labels and total weight 29, with forced, distinct and costless queries, a MAP prediction and ties kept in catalogue order, in which exact D2 and D3 have root correct masses $24/29$ and $27/29$, while their receding-horizon policies with budget four achieve $29/29$ and $28/29$.
\end{theorem}

\paragraph{Model error.}
\begin{theorem}[On-policy error]\label{thm:on-policy}
Let a true model $p$ and a fitted model $q$ share states, actions, costs and successors, and let a stop-or-query policy have fitted values in $[l,h]$ at every state. Its true and fitted values differ by at most
\[
E=\sum_{\text{query states } s} P_p(s)\,\frac{h-l}{2}\,\bigl\|p(\cdot\mid s,a_s)-q(\cdot\mid s,a_s)\bigr\|_1+\sum_{\text{stop states } s}P_p(s)\,\bigl|T_p(s)-T_q(s)\bigr|,
\]
where $P_p(s)$ is the probability of reaching $s$ under $p$. A fitted margin larger than two policies' errors implies the same true preference.
\end{theorem}

\begin{theorem}[Error radii]\label{thm:robust}
Suppose that at every state the true one-step law lies within $L_1$ radius $r(s,a)$ of the fitted one, and the true terminal value within $\tau(s)$. Then the on-policy error, and so the value gap, is at most $B(s)$, with $B(s)=\tau(s)$ at a stop and
\[
B(s)=\frac{h-l}{2}r(s,a)+\sum_o q(o\mid s,a)\,B(s_o)+\frac{r(s,a)}{2}\Bigl(\max_o B(s_o)-\min_o B(s_o)\Bigr)
\]
at a query with successors $s_o$. A margin above two such bounds gives the same true preference.
\end{theorem}
This uses only the fitted model and the radii. The probability that the radii hold is supplied outside Lean, by a concentration inequality and a union bound (Appendix~\ref{app:model-error}).

\begin{theorem}[Pessimistic choice]\label{thm:pessimistic}
Let candidate $i$ have fitted value $\hat V_i$ within $B_i$ of its true value $V_i$, and choose $k$ maximizing $\hat V_i-B_i$. Then $V_k\geq\hat V_k-B_k$; $V_k\geq V_i-2B_i$ for every $i$; and if $\hat V_k-B_k\geq\hat V_i+B_i$ for every other $i$, then $k$ is truly best. Neither this rule nor choosing by $\hat V$ dominates the other.
\end{theorem}
Choosing by $\hat V$ alone guarantees only $V_k\geq V_i-B_i-B_k$. The rule is not implemented in the controller; Appendix~\ref{app:pessimistic} evaluates it.

\begin{theorem}[Transfer]\label{thm:transfer}
For two fixed adaptive policies on a common finite space of complete worlds, with terminal reward in $[l,h]$, at most $H$ queries and cost at most $c$ per query, if the selected policy is $\epsilon$-optimal against the comparator under a fitted distribution $Q$, its true-value regret under $P$ is at most $\epsilon+2(h-l+Hc)\operatorname{TV}(P,Q)$.
\end{theorem}
Theorem~\ref{thm:on-policy} replaces the full-joint distance by one-step errors along each policy's histories. For the empirical language-model and tabular models neither bound has been evaluated, because their true conditionals are unknown.

\paragraph{Search under a budget.}
\begin{theorem}[Anytime deepening]\label{thm:anytime}
In one decision of iterative deepening, let each horizon search its candidates with the incumbent first, let the completed set hold the stop option and the completed prefix, and let the incumbent be replaced only if it completed, then by a completed candidate of highest value. For any values and any interruption pattern, the incumbent's value at the current horizon never decreases; an incumbent that did not complete is kept, so an interrupted deeper pass never discards a completed shallower decision; the final decision's value at the last horizon is at least the previous decision's; and a complete last pass returns that horizon's maximizer.
\end{theorem}
These are properties of the search's own estimates, not of executed utility.

\paragraph{Latent scores.} Here a latent model is a finite joint with a finite cause $Z$ with a prior, features independent given $Z$, and a label that is a function of $Z$.
\begin{theorem}[Latent visibility]\label{thm:visibility}
In a latent model, a fresh query leaves the posterior over $Z$ unchanged for every answer exactly when its answer laws agree on the posterior's support, that is, when its one-step latent information is zero. If no fresh query moves that posterior, no sequence of further observations does, and every lookahead of any depth is worth exactly stopping.
\end{theorem}
So if any lookahead has value, some single query has positive latent information. For two fair bits with an XOR label, one bit leaves the label posterior at one half but moves the posterior over $Z$. The result concerns visibility, not ranking: when $Z$ is the whole feature vector, every feature, distractors included, moves the posterior. Whether a fitted latent has this structure is an empirical question (Appendix~\ref{app:generic-latent}).

\begin{theorem}[Refinement]\label{thm:refinement}
Let a refinement move the answer noise of a feature $n$ into the latent cause, which then also records that answer. For any latent model and feature, both descriptions give every history the same mass and the same label masses. So exact lookahead, the exact value of any stop-or-query rule and that rule's rollout value agree. If no fresh question moves the refined posterior, none moves the original, so refinement can only add visible questions; it makes the refined feature visible whenever that feature's answer is uncertain.
\end{theorem}
A description can therefore act on a score-driven policy only through the questions the score chooses.

\begin{theorem}[Representation gap]\label{thm:gap}
Take two fair bits, their XOR as the label, a four-valued nuisance question, budget two and any cost $0\leq c\leq1/2$. With the pair of bits as the cause, each bit has latent information $\log2$ and the nuisance $0$. With the nuisance recorded, the nuisance has $\log4$ at the root and after either bit. Any rule that always takes a question of highest score earns $1-2c$ under the first description, with or without one step of rollout, and $0$ under the second, however its ties break. D2 earns $1-2c$ under both. With $c>0$, every root question is strictly worse than stopping under the second description, so executed Rollout-LIG stops.
\end{theorem}
Visibility plus one step of rollout therefore gives no uniform near-optimality guarantee. Appendix~\ref{app:representation} tests this.

\begin{theorem}[Edge cutting]\label{thm:ec2}
Let finitely many hypotheses carry nonnegative masses, a decision class and a deterministic answer to every test. An edge joins two hypotheses of different classes, with weight the product of their masses, and an answer cuts every edge with an endpoint it rules out. All edges are cut exactly when the consistent hypotheses of positive mass share one class. The uncut weight never increases as the history grows (adaptive monotonicity), and a test's expected cut weight never increases as the history grows (adaptive submodularity).
\end{theorem}

\begin{theorem}[Greedy edge cutting]\label{thm:ec2-greedy}
In the setting of Theorem~\ref{thm:ec2}, suppose some adaptive policy identifies the decision on every hypothesis of positive mass within $k$ tests, and any two hypotheses of different classes and positive mass have mass product at least $\eta>0$. At every history the uncut weight is at most $k$ times the largest expected cut weight of a single test. So after $n$ greedy tests the expected uncut weight is at most $(1-1/k)^n$ times its initial value, and with a prior that sums to one, greedy has not identified the decision with probability at most $(1-1/k)^n/\eta$.
\end{theorem}
Hence $k\ln(1/(\delta\eta))$ greedy tests identify the decision with probability at least $1-\delta$. The comparator is measured by its worst-case number of tests. The average-case comparison of \citet{golovin2011} and the noisy ECED score of Appendix~\ref{app:tier1} are not formalized.

\paragraph{Risk.}
\begin{theorem}[Monotone dynamic programming]\label{thm:monotone}
In a finite stop-or-query model, replace the expectation over a query's answers by any aggregator $A_{s,a}$ monotone in the answer values:
\[
V_0(s)=T(s),\qquad V_{n+1}(s)=\max\Bigl\{T(s),\ \max_a\bigl[x(s,a)+A_{s,a}\bigl(o\mapsto V_n(s_{a,o})\bigr)\bigr]\Bigr\}.
\]
At every state and depth, $V_n$ is nondecreasing in $n$, no stop-or-query policy evaluated with the same aggregator exceeds it, and the greedy policy attains it.
\end{theorem}

\begin{theorem}[CVaR]\label{thm:cvar}
For a finite distribution, a loss $L$ and $0\leq\alpha<1$, let $\rho(t)=t+\mathbb E[(L-t)_+]/(1-\alpha)$. An upper $\alpha$-quantile exists and $\rho$ is minimized at every upper quantile; some loss atom is one, so the minimum of $\rho$ over the loss atoms is CVaR. CVaR is at least the mean loss and monotone in the loss. For $0\leq\mu\leq1$, the mix $M_\mu(X)=(1-\mu)\mathbb E[X]-\mu\operatorname{CVaR}_\alpha(-X)$ is monotone in $X$ and never exceeds $\mathbb E[X]$. The value of Theorem~\ref{thm:monotone} with $A=M_\mu$ therefore never exceeds the risk-neutral value of the same model.
\end{theorem}

\paragraph{An executable planner.}
\begin{theorem}[Verified planner]\label{thm:exec}
For any finite joint, cost and depth, the computable planner that scans explicit lists of features and labels returns a decision whose value equals the Bellman value, never re-queries an observed feature, and, run on every world with replanning after each answer, gives a mass-weighted realized utility net of query costs equal to that value.
\end{theorem}

\paragraph{Small constructed examples.} Lean also checks a few finite instances by direct computation. They show why a tempting general claim fails; they are not general theorems.
\begin{itemize}
\item \textbf{Shortlist and interruption.} A deeper search restricted to a shortlist that omits the useful action, or interrupted before that action completes, returns less value than a complete shallower search.
\item \textbf{Low-order agreement does not ensure transfer.} A three-bit joint has the same label-plus-at-most-two-feature marginals under $P$ and $Q$ but full-joint total variation $1/2$. At cost $.03$ the fitted parity policy is worth $.91$ under $Q$ and $-.09$ under $P$, where stopping is worth zero.
\item \textbf{Conditional risk.} At confidence $.9$, a loss of $10$ with probability $.1$ and $0$ otherwise has unconditional upper-tail CVaR $10$, but the probability-weighted mean of the conditional CVaRs after the loss is revealed is $1$. The controller's default one-step CVaR is therefore not an episode-level risk objective.
\end{itemize}

\paragraph{How the theorems reach the code.} Finite tests, not proofs, link the theorems to the Python planners. All are differential checks against values or decisions computed in Lean:
\begin{itemize}
\item the noisy-parity fixtures of Theorem~\ref{thm:parity}: 243 histories in eight noise and cost cases, covering 1,944 posteriors, 7,776 history--depth values and 9,720 candidate values, within $2\times10^{-12}$;
\item four static models with forced accuracy scoring: 3,988 decisions, 1,408 trajectories and 1,236 branch probabilities, with largest difference $6.67\times10^{-16}$;
\item the planner of Theorem~\ref{thm:exec} run on a 32-world noisy-parity model: 376 decisions, which the Python research planner matches exactly, ties included;
\item the search rules of Theorem~\ref{thm:anytime}: 180 recorded decisions and 228 search passes, 96 of them interrupted;
\item histories, budgets and ties: 259 ordered history inputs, 180 budget combinations and 150 depth, query and tick combinations.
\end{itemize}
Numerical tolerances can change a decision at a near tie: at a constructed root the planner's $10^{-12}$ selection tolerance chooses abstention although the exact D3 gain is $3\times10^{-13}$. None of these checks is a refinement proof of the Python source or a floating-point error bound, and none validates the LLM channels, interrupted search or the full controller.

\begin{table}[htbp]
\centering\footnotesize
\setlength{\tabcolsep}{4pt}
\caption{The Lean file that checks each theorem.}\label{tab:lean-map}
\begin{tabular}{@{}ll@{\qquad}ll@{}}
\toprule
Theorem & Lean file & Theorem & Lean file \\
\midrule
\ref{thm:parity} & \code{Parity}, \code{ParityGeneralNoise} & \ref{thm:anytime} & \code{AnytimeDeepening} \\
\ref{thm:adaptivity} & \code{ContextParityUpper}, \code{ContextParityGeneral} & \ref{thm:visibility} & \code{LatentVisibility} \\
\ref{thm:interaction-order} & \code{InteractionOrder} & \ref{thm:refinement}, \ref{thm:gap} & \code{RepresentationRefinement} \\
\ref{thm:interaction-sufficiency} & \code{InteractionOrderSufficiency} & \ref{thm:ec2} & \code{EquivalenceClassEdges} \\
\ref{thm:rollout} & \code{LookaheadRollout} & \ref{thm:ec2-greedy} & \code{EquivalenceClassGreedy} \\
\ref{thm:replanning} & \code{StaticAcquisitionCounterexample} & \ref{thm:monotone} & \code{MonotoneBellman} \\
\ref{thm:on-policy} & \code{OnPolicyError} & \ref{thm:cvar} & \code{CVaRSpec}, \code{CVaRCompute} \\
\ref{thm:robust} & \code{RobustOnPolicy} & \ref{thm:exec} & \code{PlannerExec} \\
\ref{thm:pessimistic} & \code{PessimisticPlanning} & Examples & \code{Bounded}, \code{ConditionalRisk}, \code{ModelMismatch} \\
\ref{thm:transfer} & \code{AdaptiveTransfer}, \code{TransferBounds} & & \\
\bottomrule
\end{tabular}
\end{table}
\section{Model-error experiments and certificates}\label{app:model-error}
All studies in this section were frozen locally before their respective test scoring, not externally preregistered. They add evidence about particular models and tasks; they do not establish general superiority, operational safety or a source-level proof of Python.

\paragraph{Controlled model-error experiment.}
We generate five independent fair bits, with a parity target from the first three and independent label-flip probabilities $q\in\{0,.15,.30\}$. The other bits are distractors. Full-joint count models use a symmetric Dirichlet prior of total mass one and fit either samples from the true generator or samples with independent labels. Each of three sample sizes, six fitting seeds and two query costs ($.03$ and $.10$) is evaluated under each source and noise setting: 216 cells. D1--D3, D1 rollout and forced CMI are frozen under the fitted model and evaluated exactly under the known true joint, with budget three and signed utility. The true joint is available only because this is a simulation.

In the 108 matched-source cells, the sufficient condition that the fitted complete-policy D3--D2 margin exceed $2(2+3c)\operatorname{TV}(P,Q)$ holds in 0/36, 15/36 and 30/36 cells at 128, 1,024 and 8,192 samples, respectively. It holds in none of 108 shifted-source cells. In the shifted cells, 19 fitted margins favor D3 and six have negative true margins; none is certified. All finite bound inequalities hold in the recorded cells. These are descriptive results for a known generator, not a confidence bound on model error in an unknown environment; search work is unequal.

\paragraph{On-policy certificates on the same cells.} A separate stage, frozen before any on-policy error was computed, reproduced all 216 cells from the same training draws. Every fitted and true utility matched the recorded values to $10^{-12}$. The stage then evaluated the on-policy error of each frozen policy.

The primary range is $[-3c,1]$, valid for every fitted node value under signed utility with abstention and three queries; a secondary variant uses each policy's exact fitted range. The inequality held for all 1,080 policies. Table~\ref{tab:onpolicy-certificates} compares the certificates. No on-policy certificate has the wrong sign; the median ratio of the actual gap to the bound is $.63$ in matched cells, and the bound is attained in some cells.

Replacing the true conditionals by an independent held-out sample of the same size overestimates the error in the median, by a factor of 1.41 in matched cells. It underestimates it for 98 of 540 matched policies and 208 of 540 independent-label policies, and used as a certificate it would accept four wrong-sign pairs. Estimated conditionals therefore need their own confidence argument before the bound can certify preferences in an unknown environment.
\begin{table}[htbp]
\centering\footnotesize
\setlength{\tabcolsep}{4pt}
\caption{Certified preferences in the controlled model-error study. Upper rows: D3 over D2 in 36 cells per row. Lower rows: all ordered method pairs with a positive fitted margin. Full joint is the margin condition $2(2+3c)\operatorname{TV}(P,Q)$; on-policy uses the range $[-3c,1]$, or each policy's exact fitted range. No on-policy certificate has the wrong sign. The plug-in column replaces the true conditionals by an independent held-out sample; it is an estimate, not a certificate, and certifies four wrong-sign independent-label pairs.}\label{tab:onpolicy-certificates}
\begin{tabular}{@{}lrrrrrrr@{}}
\toprule
Source & $n$ & Fitted $>0$ & True $>0$ & Full joint & On-policy & Exact range & Plug-in \\
\midrule
Matched & 128 & 25 & 19 & 0 & 6 & 10 & 5 \\
Matched & 1,024 & 29 & 29 & 15 & 21 & 24 & 21 \\
Matched & 8,192 & 36 & 36 & 30 & 36 & 36 & 33 \\
Independent label & 128 & 14 & 6 & 0 & 0 & 0 & 0 \\
Independent label & 1,024 & 5 & 4 & 0 & 0 & 0 & 0 \\
Independent label & 8,192 & 0 & 0 & 0 & 0 & 0 & 0 \\
Matched, all pairs & -- & 772 & -- & 275 & 452 & -- & 371 \\
Independent label, all pairs & -- & 565 & -- & 0 & 132 & -- & 100 \\
\bottomrule
\end{tabular}
\end{table}
 
\paragraph{A confidence certificate.} A further stage, frozen before any confidence bound was computed, reproduced the same 1,080 policies. It then certified preferences with the fitted-model bound of Theorem~\ref{thm:robust}. The radii were built as follows.
\begin{itemize}
\item \textbf{Validation sample.} It is drawn from the true distribution in every cell, of size $m=n$ or $m=8n$. In the independent-label cells training used the wrong source, but a certificate about true performance needs data from the true distribution.
\item \textbf{Radius.} For every query pair and label law a three-query policy can touch (296 per cell), the radius is the validation $L_1$ distance to the fitted law plus the binary deviation $\sqrt{2\ln(2/\delta')/n_h}$ of \citet{weissman2003l1}, capped at 2. Here $\delta'=.05/296$ and $n_h$ counts validation rows at the history.
\item \textbf{Leaves.} A leaf's terminal error is at most its label-law radius, or 0 for abstention.
\end{itemize}
With probability at least .95 every radius in a cell holds, so every certificate issued there is correct.

Table~\ref{tab:confidence-certificates} reports the result. Every frozen expectation held.
\begin{itemize}
\item No certificate has the wrong sign, and the radii held in all 108 training draws at both sizes.
\item At $m=n$ the certificate certifies D3 over D2 only at 8,192 samples (23 cells). At $m=8n$ it certifies 12 cells at 1,024 and 30 at 8,192, against the oracle's 21 and 36.
\item In the independent-label cells it certifies 60 ordered pairs at $m=8n$, all correct, where the plug-in estimate certified four wrong-sign pairs.
\item The recursion itself costs little. With the true radii it is a median 1.002 times the oracle error and certifies 451 matched pairs against the oracle's 452. The concentration term costs a factor of 7.8 at $m=n$ and 3.3 at $m=8n$.
\end{itemize}
The cells are synthetic, with five binary features and three queries. The union bound grows quickly with features and budget.
\begin{table}[htbp]
\centering\footnotesize
\setlength{\tabcolsep}{4pt}
\caption{Confidence certificates in the controlled model-error study. Upper rows: D3 over D2 in 36 cells per row; lower rows: all ordered pairs with a positive fitted margin. Wrong-sign certifications are in brackets. Oracle uses the true conditionals and reach probabilities. The plug-in estimate uses a held-out draw from the training source. Exact radii use the fitted-model recursion with the true $L_1$ distances. The confidence columns use radii from a validation sample of size $m$ drawn from the true distribution, with $\delta=.05$ per cell. The radii held in 108/108 draws at $m=n$ and 108/108 at $m=8n$. Median bound over oracle error: 1.002 (exact radii), 7.8 ($m=n$), 3.3 ($m=8n$).}\label{tab:confidence-certificates}
\begin{tabular}{@{}lrrrrrrr@{}}
\toprule
Source & $n$ & Fitted $>0$ & Oracle & Plug-in & Exact radii & $m=n$ & $m=8n$ \\
\midrule
Matched & 128 & 25 & 6 & 5 & 6 & 0 & 0 \\
Matched & 1,024 & 29 & 21 & 21 & 21 & 0 & 12 \\
Matched & 8,192 & 36 & 36 & 33 & 36 & 23 & 30 \\
Independent label & 128 & 14 & 0 & 0 & 0 & 0 & 0 \\
Independent label & 1,024 & 5 & 0 & 0 & 0 & 0 & 0 \\
Independent label & 8,192 & 0 & 0 & 0 & 0 & 0 & 0 \\
Matched, all pairs & -- & 772 & 452 & 371 & 451 & 100 & 244 \\
Independent label, all pairs & -- & 565 & 132 & 100 (4) & 132 & 0 & 60 \\
\bottomrule
\end{tabular}
\end{table}

\section{The implemented proactive controller}\label{sec:controller}
The controller separates judgments from decisions. For history $h$, it maintains a distribution $q(z)$ over finitely many hypotheses, considers immediate actions $a\in\mathcal I$ and investigations $g\in\mathcal G$, and assigns each investigation a finite outcome vocabulary. Language models supply hypothesis-conditioned rows of probabilities; everything after that is arithmetic on finite tables.

\subsection{Where the likelihoods come from}\label{sec:controller-channel}
No model is asked to state a number. An ensemble of four locally served models (Gemma, Qwen, GLM and Nemotron) supplies them. Each member receives a prompt that assumes the cause is $z$ and asks which outcome $g$ will reveal, with its answer constrained to the outcome vocabulary. The controller reads the log probabilities of those option tokens and normalizes them over the vocabulary, treating an option outside the returned top candidates as impossible. Three further steps then act on these rows:
\begin{itemize}
\item \textbf{Temperature.} One temperature per member is fitted on held-out labeled pairs by minimizing negative log likelihood. It is kept only if the member is miscalibrated (expected calibration error above $.05$) and the fit improves that error by at least $.01$.
\item \textbf{Pooling and labeled counts.} Members' rows are pooled into $m_{gz}(o)$, floored and renormalized, and form Dirichlet parameters
\begin{equation}
 \alpha_{gzo}=c_{gz}m_{gz}(o)+N_{gzo},\qquad
 L_g(o\mid z)=\frac{\alpha_{gzo}}{\sum_{o'}\alpha_{gzo'}}.
 \label{eq:controller-channel}
\end{equation}
Here $c_{gz}$ is a bounded concentration moment-matched from member dispersion and $N$ counts realized outcomes with known labels, so data progressively outweigh elicitation.
\item \textbf{A conformal set.} When labeled hypotheses exist, split conformal prediction gives a set containing the true hypothesis with probability at least $1-\delta$ under exchangeability, with no calibration assumption. The risk term treats every hypothesis in the set as live, whatever its point probability.
\end{itemize}
These are mechanisms, not evidence. The temperature corrects only a global bias, the Dirichlet prior is a modeling choice, and the coverage holds only under exchangeability. Appendix~\ref{app:elicited} measures the reads, the temperature and the coverage on fixed questions. Every other result in Section~\ref{sec:main-results} replaces the elicited rows by fitted tables or by counts of repeated answers, and the conformal set acts only through the risk term, which those results switch off.

\subsection{Update belief, score actions and decide}
For a recognized outcome the shared table gives $p_g(o)=\sum_z\bar q(z)L_g(o\mid z)$ and $q^{g,o}(z)=\bar q(z)L_g(o\mid z)/p_g(o)$, where $\bar q$ is the predictive belief; under the default static cause $\bar q=q$. An unrecognized or absent outcome is not evidence. For a return distribution $R_a$ and loss $L_a=-R_a$, the implementation computes the upper-tail loss CVaR \citep{rockafellar2000} exactly over its finite weighted atoms:
\begin{equation}
 C_\alpha(L_a)=\min_{t\in\mathbb R}\left\{t+\frac{\mathbb E[(L_a-t)_+]}{1-\alpha}\right\}.
 \label{eq:controller-cvar}
\end{equation}
A local selector scores each action by $(1-\mu)U(a)-\mu C_\alpha(L_a)$ plus weighted state and parameter information, and reweights the actor's prior by a Kullback--Leibler regularized rule \citep{ortega2013,levine2018}. With $X(a,h,q)=(1-\mu)U(a,h,q)-\mu C_\alpha(-R_a)$ and $T(h,q)=\max_{a\in\mathcal I}X(a,h,q)$, the gate computes
\begin{equation}
\begin{aligned}
 V_0(h,q)&=T(h,q),\\
 V_d(h,q)&=\max\!\left\{T(h,q),\max_{g\in\mathcal G(h)}\left[X(g,h,q)
 +\sum_o p_g(o)V_{d-1}(h\oplus(g,o),q^{g,o})\right]\right\},
\end{aligned}
 \label{eq:controller-value}
\end{equation}
and investigates when the first investigation's expected continuation, net of cost, beats acting now. With a deterministic cost $c_g$, $X(g,h,q)=-c_g$. The gate prices decision value in reward units and does not add the selector's information bonuses again. Guards, approval and an optional verifier can still alter the action. The runtime's defaults are $\mu=.5$, depth two and point-table expectimax. Expectation backups of these scores do not implement nested dynamic CVaR; an opt-in nested backup does (Appendix~\ref{app:risk-planner}). Reservation-value and Monte Carlo gates are alternative configurations with no optimality claim.

\subsection{How the experiments relate to the controller}\label{sec:controller-bridge}
The experiments keep the pattern of Equation~\eqref{eq:controller-value}: condition on a candidate answer, update a belief, evaluate the best continuation and replan after the actual answer. They use a fitted table, $\mu=0$ and a terminal decision specialized to classification or signed accuracy, and they do not run the LLM selector, CVaR score, guards, verifier or cross-run memory. The runtime gate also implements the LIG search and latent-ordered deepening of Section~\ref{sec:main-methods}, and Appendix~\ref{app:controller-real} shows that with a fitted latent-class model it makes exactly the research planners' decisions. That evidence does not cover the language-model channels, selector, verifier, memory or risk weighting. Appendix~\ref{app:elicited} measures those channels on fixed questions. Future work should ablate selector, gate, risk and verifier on independent scenarios.
\section{Datasets and what each evaluation measures}\label{sec:datasets}
Section~\ref{sec:main-datasets} summarizes the tasks. We do not pool accuracy and signed utility, and synthetic controls establish mechanisms, not application-level generalization.

\paragraph{Context--parity controls.}
A uniform binary context selects one of two independent blocks of $k\in\{2,3\}$ uniform bits. The label is the selected block's parity, flipped with probability $\epsilon\in\{0,.1\}$. The budget is $B=k+1$ and each query costs $.03$. Context and any proper subset of the selected block do not determine parity. We enumerate the finite generator and fit a saturated joint from $N\in\{120,600\}$ sampled references, with total Dirichlet concentration one. Eight cells have ten independent fits each. Known-generator policies are diagnostic controls; learned policies receive only their sampled references.

\paragraph{Noisy CUBE-NM.}
We retain the pinned author constructor and use a context-backoff model. A uniform class $y\in\{0,\ldots,7\}$ and independent uniform context $z$ select a ten-coordinate block. Within it, the three binary digits $b_r(y)$ are inserted at positions $(y+r)\bmod 10$, for $r=0,1,2$; other coordinates have background mean $.5$. Thus the locations as well as the values of the informative entries depend on the label. A noisy one-hot context vector precedes the blocks, giving $11C$ raw coordinates and $1+10C$ acquisition groups. The base grid crosses $C\in\{2,5\}$ contexts, $N\in\{120,600\}$, and $c\in\{.01,.05,.10\}$ at $B=3$. Four additional cells vary informative-feature noise to $.05/.2$ or budget to $2/4$, holding $C=2,N=600,c=.01$ and the other default parameters fixed. Context and background standard deviations remain $.1/.3$. Each of these 16 prespecified cells has ten fresh training seeds. Training-only tertiles discretize observations, and three-fold full-vector NLL selects emission strength on the unchanged grid. Gaussian probabilities and context-confusion probabilities give the exact expected return of every frozen policy.

\paragraph{Noiseless CUBE-NM.}
AFABench's pinned noiseless setting has five contexts, eight classes, 55 raw coordinates and 51 unit-cost acquisition groups \citep{afabench2026}. The initial confirmation cohort has five independent generator/split and shared-MLP fits; the added cohort has five further fits under the same 600/200/200 training/validation/reserved-test split. The author classifier recipe uses random masks, 100 epochs and validation checkpoint selection. Earlier cross-classifier analyses score the initial cohort's masks with all five original MLPs. The learned-baseline comparison uses the same own-fit scorer for every method in each of the ten fits and reports the two cohorts separately. The primary outcome averages all 40 uniform class--context atoms. SEFA additionally averages ten acquisition seeds, so its internal-randomness expectation remains estimated. Reserved-test scores are secondary. Splits are disjoint by row index; the same finite patterns can recur across splits. Fresh fits measure fitting variation within this task; they are not new task distributions or a reproduction of published baseline scores.

\paragraph{Common 20 Questions.}
The public UoT vocabulary contains 111 unique candidates: 27 animals, 26 foods, 37 objects and 21 places \citep{hu2024uot}. Qwen and GLM each generate one new bank of 32 semantic yes/no questions before answers are collected. For every bank, model and item, six fresh requests supply fitting vectors and three disjoint requests supply examiner vectors. Each request answers all 32 questions jointly. All four fitting/examiner directions are evaluated: Qwen--Qwen, Qwen--GLM, GLM--GLM and GLM--Qwen. Policies acquire exactly $B\in\{3,5,7\}$ answers, then make one candidate-index guess outside the query budget. This is closed-set identification under fixed prompts, not factual answer validation, unseen-item generalization, or UoT's free-form dialogue benchmark. Appendix~\ref{app:third-examiner} adds a third examining family.
Under the uniform target prior, a binary $B$-query policy has accuracy at most $\min\{1,2^B/111\}$: each of its at most $2^B$ leaves can correctly identify mass no greater than $1/111$. The generic ceilings are therefore 7.2\%, 28.8\% and 100\% at the three budgets. A particular question bank can impose a lower ceiling.

\paragraph{Asset licenses.} License names below were read from the local copies of each asset, or from the source's own page where noted.
\begin{itemize}
\item \textbf{Code.} AFABench, including the CUBE-NM generator and evaluator, and the SEFA author code, also used for DIME, are MIT licensed. The AACO author repository declares no license; we run it only locally, for research comparison.
\item \textbf{Tabular data.} Zoo, Wine, Breast Cancer, Iris, Digits and MiniBooNE are UCI Machine Learning Repository datasets, released under CC BY 4.0; Wine, Breast Cancer, Iris and Digits were loaded through scikit-learn. The scikit-learn Diabetes regression data is distributed with scikit-learn (BSD-3-Clause). The three-class Diabetes CSV is used as distributed with AFABench; the local copy does not state its upstream source or license. The PhysioNet 2012 Challenge data, used through AFABench, are under the Open Data Commons Attribution License v1.0, per the PhysioNet page.
\item \textbf{Question vocabularies.} The Common item list comes from the Uncertainty of Thoughts repository, which declares no license. We use only the item and category names, for research evaluation.
\item \textbf{Language models.} The models were run locally through Ollama, with license names taken from the local model packages: Qwen3.6 27B (Apache 2.0), GLM-4.7-Flash (MIT), Gemma 4 26B (Apache 2.0) and Nemotron 3 33B (NVIDIA Open Model License).
\end{itemize}

\paragraph{Compute.} All planning experiments ran CPU-only on one workstation: an Apple M5 Max with 18 cores and 64\,GB of memory, under macOS~26 and~27 according to the saved environment records. Numerical libraries were limited to one thread. The language-model panels ran on the same machine, serving the models locally through Ollama.

Search work is reported as logical transitions, and wall times come from this non-isolated host. Each rerun in Appendices~\ref{app:anytime}, \ref{app:decision-aware}, \ref{app:tier1}, \ref{app:tier2} and~\ref{app:families} finished in under an hour, and the certificate study in Appendix~\ref{app:model-error} in under a minute. Total compute for the whole project, including development and abandoned runs, was not logged.
\section{Planning, prediction and computational budgets}\label{sec:confirmation-methods}
\subsection{Compare complete policies under a common task}
Let $h=(S,u_S)$ be the acquired history and $b$ the remaining query budget. A fitted joint $\widehat p$ supplies the class posterior $q(y\mid h)$ and future-outcome probabilities $\widehat p(o\mid h,j)$. For forced accuracy tasks, $T(h)=\max_yq(y\mid h)$ and
\begin{equation}
 V_0(h)=T(h),\qquad
 V_d(h)=\max_{j\notin S}\sum_o\widehat p(o\mid h,j)V_{d-1}(h\oplus(j,o)).\label{eq:confirmation-search}
\end{equation}
A depth-$D$ planner chooses its next query with horizon $\min(D,b)$, then replans. A two-step planner can therefore acquire seven groups. Signed-utility tasks instead use $T(h)=\max\{0,2\max_yq(y\mid h)-1\}$, subtract $c$ per query and allow stopping at every node. For a frozen policy $\pi$, evaluation sums all terminal leaf masses under the evaluation distribution. We distinguish this complete-policy value from a truncated search-root value.
Here ``exact'' means finite outcome or support summation rather than Monte Carlo policy evaluation. Noisy-generator probabilities use Gaussian CDFs and numerical quadrature for context confusion, so their values remain subject to integration and floating-point precision.

Rollout evaluates each candidate query followed by the complete D1 continuation through the remaining budget, and repeats that improvement after each observation. CMI maximizes conditional mutual information and spends the available query budget; on signed-utility tasks it can still abstain at the terminal. It does not optimize a cost-sensitive stopping rule. Paid context-first reveals context and then uses D1. The forced-D1 control must spend its budget but retains the same signed terminal and final abstention option. Exact nonadaptive subset selection maximizes fitted expected utility over all subsets of size at most $B$; it cannot choose later queries based on earlier answers. These controls distinguish longer-horizon valuation, willingness to acquire, and adaptive branching.

\subsection{Hold the observation model fixed when changing the policy}
The noisy grid uses the NLL-selected class--context mixture with context-backoff emissions. The constructed parity family instead uses a saturated finite joint, so it can represent the required interaction. Known-generator policies use the same action space and reward as learned policies, but are labeled oracle diagnostics. We report exact generator regret, value error, acquisition count and outcome-distribution error at histories visited by the policy.

The primary noiseless observation model uses distinct-value bins with uniform concentration-one emissions and component masses.

For Common, $s_{yj}$ YES answers among $n$ fitting requests give
\begin{equation}
 \theta_{yj}=\frac{s_{yj}+\alpha/2}{n+\alpha},\qquad
 \widehat p(y,u)=\frac1{111}\prod_{j=1}^{32}\theta_{yj}^{u_j}(1-\theta_{yj})^{1-u_j}.
\end{equation}
These are smoothed answer frequencies, not token log probabilities or model-reported confidence. Primary fits use $n=6,\alpha=1$. Prespecified sensitivities use $\alpha\in\{.2,4\}$. Conditional independence given the item remains an assumption, although answers are elicited jointly. No parameter is selected using examiner scores. Prospective policies are compiled from fitting answers before examination.

\subsection{Align planning with the evaluated classifier}
The ordinary accuracy objective optimizes the fitted MAP predictor, while AFABench scores a separate masked MLP $f_\psi$. The aligned control replaces only the search terminal by
\begin{equation}
 T_\psi(h)=q\!\left(\arg\max_y f_\psi(y\mid x_S,S)\,\middle|\,h\right).\label{eq:aligned-terminal}
\end{equation}
Thus the reward is expected correctness of the frozen classifier under the same fitted posterior, not its softmax confidence. The transition law and query budget remain unchanged. Alignment uses distinct-value encodings, permitting raw acquired values to be reconstructed without unobserved coordinates. The initial cohort also scores each resulting policy's masks with all five classifiers. This is an objective-alignment intervention using an established expected-reward construction, not a new Bellman recursion.

\subsection{Compare search under the same resource allowance}
The matched comparisons assign every acquisition decision the same allowance of $512$, $4{,}096$ or $32{,}768$ candidate--outcome transitions. They compare bounded D1, D2, D3 and rollout. Each computes a complete D1 fallback, then considers deeper candidates in a fixed greedy-first order. Only completed candidate values can replace the fallback; interrupted values are discarded. Search caches are cleared between decisions. These are approximate resource-bounded policies, not exact D2 or D3 results. Appendix~\ref{app:anytime} adds iterative deepening with path reuse under the same allowances.

An additional matched control applies these same allowances to D1--D3 and rollout with classifier-aligned terminals, at $B=3,7$. It clears classifier-prediction caches as well as search caches and records planning-time classifier-history evaluations. This control separates depth from changing the predictor objective.
\section{Results: when additional search changes the outcome}\label{sec:confirmation-results}
\subsection{Longer horizons recover a designed information dependency}
The context--parity family supplies a positive control with an explicit reason to plan. Acquiring context and the selected block gives signed utility $1-2\epsilon-(k+1)c$. The best nonadaptive subset gives $\max\{0,1/2-\epsilon-kc\}$, leaving an adaptive advantage of $.47$ or $.37$ in the tested cells (Appendix~\ref{app:context-parity}). A horizon of $k$ cannot value the complete context-first sequence; $k+1$ can.

Across the eight learned-model cells, full-budget depth improves mean exact utility by $.442$ over one shorter horizon (adjusted interval $[.435,.449]$) and by $.434$ over fitted optimal subset acquisition ($[.425,.443]$). These contrasts average ten independent fitting blocks; they are not uncertainty over the exact generator evaluation. The noisy, low-reference $k=3$ cell remains below the oracle. The result establishes that additional lookahead can recover adaptive value when the observation model represents the required dependency. It does not establish that this dependency is common in applications.

\subsection{The noisy-grid gain depends on both the task and available search}
Across the 16 prespecified noisy settings, unrestricted D3 improves mean exact utility by $.0373$ over D2 (adjusted interval $[.0323,.0423]$) and by $.0274$ over rollout ($[.0182,.0366]$). This is heterogeneous: all six five-context cells abstain under the fitted D1--D3 and rollout policies, although known-generator D3 has positive utility in four lower-cost cells. Zero value error for abstention therefore does not certify accurate observation probabilities. Forced D1 has negative mean utility in 15 of 16 cells; merely forcing acquisitions does not reproduce the gain. The appendix retains every cell and oracle comparison.

\begin{table}[htbp]
\centering\small
\setlength{\tabcolsep}{4pt}
\caption{Noisy-grid mean utility under equal per-decision transition allowances. The final column is D3's actual expected transitions per episode, so it can exceed one decision's allowance. These are bounded approximations with a declared greedy fallback.}\label{tab:matched-noisy}
\begin{tabular}{@{}rrrrrr@{}}
\toprule
Allowance & D1 & D2 & D3 & Rollout & D3 work \\
\midrule
512 & 0.000 & 0.000 & -0.000 & -0.000 & 465 \\
4,096 & 0.000 & 0.028 & 0.000 & 0.039 & 4,033 \\
32,768 & 0.000 & 0.028 & 0.053 & 0.038 & 34,179 \\
\bottomrule
\end{tabular}
\end{table}
 The benefit changes under equal search allowances (Table~\ref{tab:matched-noisy}). At 4,096 transitions per decision, bounded D3 is worse than D2 by $.0276$ (adjusted interval $[-.0308,-.0244]$) and worse than rollout by $.0386$ ($[-.0442,-.0330]$). At 32,768, D3 instead improves by $.0249$ over D2 ($[.0221,.0276]$) and by $.0150$ over rollout ($[.0070,.0229]$). At 512, neither contrast establishes an improvement. The largest-allowance D3 uses about 34,179 expected transitions per episode, versus 12,198 for D2 and 13,734 for rollout. Thus a common ceiling can permit a depth benefit while still assigning different realized work. This result concerns the specified candidate order and complete-greedy fallback; it is not a universal computational threshold. In particular, the 4,096 loss arises because this D3 returns the D1 action when no depth-three root candidate completes. With iterative deepening, D3 equals D2 at 4,096. At 32,768 it exceeds D2 by $.0123$ ($[.0074,.0171]$), or by $.0331$ ($[.0275,.0387]$) when its deeper candidates are ordered by latent information. The original bounded D3's 32,768 advantage depends on the catalogue listing context first: with context listed last, the same planner scores $.0335$ (Appendix~\ref{app:anytime}).

\subsection{Two-step aligned planning nearly saturates noiseless CUBE-NM}
The original five fits distinguish available information from scored accuracy. At seven queries, ordinary D2, D3 and rollout all reach 100\% internal MAP accuracy, but their own-classifier accuracies are 77.5\%, 80.5\% and 82.5\%. D3--D2 is $+3.0$ percentage points (adjusted interval $[-6.6,12.6]$), and D3--rollout is $-2.0$ points ($[-10.5,6.5]$). Neither establishes a depth advantage.

Changing D3's terminal objective to expected correctness of the scored classifier raises its own-classifier accuracy to 100\%, a gain of 19.5 points ($[13.7,25.3]$). Aligned D2 already reaches 99.5\%. Scoring aligned D3's masks with the other four classifiers reduces mean accuracy to 88.5\%, indicating classifier-specific adaptation. Appendix~\ref{app:noiseless-controls} retains the cross-classifier matrix and full original-cohort results.

\begin{table}[htbp]
\centering\small
\caption{Shared-classifier accuracy (percent) on the noiseless CUBE-NM support. Fresh and original cohorts each have five fits. SEFA averages ten acquisition seeds per fit; other policies are deterministic. All methods use the same splits, groups and own-fit scorer. Original denotes reused data/scorer fits; neural baselines are newly trained.}\label{tab:learned-baselines}
\begin{tabular}{@{}lrrrr@{}}
\toprule
Method & Fresh B3 & Fresh B5 & Fresh B7 & Original B7 \\
\midrule
D1 & 29.00 & 42.50 & 51.00 & 50.00 \\
D2 & 56.00 & 77.00 & 79.00 & 77.50 \\
D3 & 56.00 & 76.50 & 77.50 & 80.50 \\
Rollout & 56.00 & 77.00 & 85.00 & 82.50 \\
CMI & 27.00 & 49.00 & 63.50 & 56.00 \\
Aligned D1 & 30.50 & 44.00 & 54.50 & 55.50 \\
Aligned D2 & 71.00 & 99.00 & 99.50 & 99.50 \\
Aligned D3 & 71.00 & 99.00 & 100.00 & 100.00 \\
DIME & 23.50 & 26.50 & 33.50 & 33.00 \\
SEFA & 28.95 & 36.55 & 45.55 & 46.35 \\
\bottomrule
\end{tabular}
\end{table}

\paragraph{Additional fits and learned acquisition.}
The fresh cohort tests the same conclusions against trained DIME and SEFA (Table~\ref{tab:learned-baselines}). At seven queries, ordinary D3 reaches 77.50\%, DIME 33.50\% and SEFA 45.55\%. With the aligned objective, D1 reaches 54.50\%, D2 99.50\% and aligned D3 100.00\%. Appendix~\ref{app:learned-baselines} reports all four adjusted learned-baseline contrasts and the complete cohorts. The finite planner explicitly conditions emissions on class and context, whereas the neural baselines learn their representations. Aligned search also consults the scored MLP at hypothetical histories, whereas the neural acquisition rules use private predictors. Cross-method differences therefore combine acquisition, model class and predictor objective; they do not isolate planning depth or establish general superiority over learned acquisition.

In a separate replication family, fresh-cohort D3--D2 is -1.50 [-8.04, 5.04] points and D3--rollout is -7.50 [-16.24, 1.24]. These depth contrasts must be distinguished from changing the terminal objective.
Neither learned baseline acquires context first in these additional-fit evaluations. At seven queries, the compatible-support posterior diagnostic averages 37.50\% for DIME and 51.32\% for SEFA; their private predictors reach 32.50\% and 47.65\%. The observed gaps therefore involve the acquired information as well as prediction, rather than only disagreement with the shared MLP. These diagnostics do not identify which training or model-class change would close the gaps.
 
\subsection{The original bounded D3 loses through its fallback}
At seven queries and 32,768 transitions per decision, ordinary D2, D3 and rollout reach 77.5\%, 56.5\% and 72.5\% shared-MLP accuracy. With the same classifier-aligned terminal, they reach 99.5\%, 61.5\% and 98.0\%. The latter D3--D2 gap is $-38.0$ points (adjusted interval $[-49.8,-26.2]$); D3--rollout is $-36.5$ ($[-49.6,-23.4]$).

Bounded D3 completes no root candidate and starts with the mandatory greedy fallback in all five fits. At the largest allowance, its first five decisions match that fallback on every fit--support case. Aligned D2 completes all 51 root candidates, chooses context first, and leaves at most 0.5 points of headroom. Thus these losses reflect incomplete search under the specified order and fallback, not an intrinsic disadvantage of exact depth-three planning. Table~\ref{tab:aligned-matched} reports all allowances.

The iterative-deepening rerun confirms this directly. With the same fits, scorers and allowances, aligned D3 recovers 36.5 points ($[24.0,49.0]$) at 4,096 and 38.0 points ($[26.2,49.8]$) at 32,768, matching D2 at both budgets and all allowances (Table~\ref{tab:anytime-aligned}). The ordinary-terminal matched comparison above was not rerun and retains the original fallback.

\subsection{Common's same-model depth gain weakens under transfer and limited search}
The fresh panels distinguish new answers from a change of answering model (Table~\ref{tab:llm-fresh}). With six fitting requests per item and seven queries, Qwen-fitted D3 reaches 47.75\% on independent Qwen answers, compared with 38.59\% for D2. The paired gain is 9.16 percentage points, with an adjusted target-bootstrap interval of $[4.20,14.41]$. This exceeds the protocol's three-point practical reference within the fixed-bank, fitted-panel interpretation. GLM-fitted D3 reaches 50.45\% on GLM answers versus 44.44\% for D2; its six-point gain has interval $[-.30,12.46]$ and remains unresolved after adjustment.

Neither same-model comparison establishes an advantage over rollout. Rollout reaches 53.30\% on Qwen and 52.10\% on GLM; D3--rollout intervals are $[-12.31,1.31]$ and $[-9.16,5.67]$ points. CMI also reaches 52.85\% and 51.50\%, respectively. Bank variation is substantial: Qwen--Qwen D3 scores 56.46\% on the Qwen-generated bank and 39.04\% on the GLM-generated bank; rollout scores 70.27\% and 36.34\%. The equal-bank average therefore does not imply a common ranking in each bank.
\begin{table}[htbp]
\centering\small
\setlength{\tabcolsep}{4pt}
\caption{Independent Common examiner panels: seven-query identification accuracy (\%), averaging the two banks equally. Policies fit six requests per item and use concentration one. Each bank/direction scores 111 targets with three disjoint examiner requests. Q and G denote Qwen and GLM; arrows identify fitting and examination.}\label{tab:llm-fresh}
\begin{tabular}{@{}lrrrr@{}}
\toprule
Policy & $Q\to Q$ & $Q\to G$ & $G\to G$ & $G\to Q$ \\
\midrule
D1 & 32.73 & 16.82 & 37.69 & 14.11 \\
D2 & 38.59 & 16.52 & 44.44 & 16.22 \\
D3 & 47.75 & 18.32 & 50.45 & 18.62 \\
Rollout & 53.30 & 16.67 & 52.10 & 15.92 \\
CMI & 52.85 & 20.42 & 51.50 & 17.42 \\
\bottomrule
\end{tabular}
\end{table}
 
Transfer is much weaker (Appendix~\ref{app:families} plans over both families). Qwen-fitted D3 scores 18.32\% on GLM answers, and GLM-fitted D3 scores 18.62\% on Qwen answers. Their gains over D2 are 1.80 points ($[-2.40,5.86]$) and 2.40 points ($[-2.70,7.21]$). Neither transfer direction establishes a gain over D2 or rollout. These intervals also leave practically relevant effects unresolved; they are not evidence of equivalence. At three queries, same-model accuracy is already close to the generic 7.2\% binary-tree ceiling. At five queries, D3 reaches 26.88\% and 27.03\% against the generic 28.8\% ceiling. A small depth gap at a restricted acquisition budget can therefore reflect little remaining accuracy headroom.

The unrestricted gain also does not survive as an efficiency result. At 32,768 transitions per decision, bounded D3 reaches 38.74\% on Qwen and 41.74\% on GLM, compared with 38.59\% and 44.44\% for D2. Bounded rollout reaches 49.40\% and 49.25\%. D3 spends approximately 166,719 transitions per episode, versus 21,060 for D2 and about 132,000 for rollout. The common ceiling admits different realized work, and this implementation's bounded D3 is not the strongest same-model policy. This Common comparison uses the original D1 fallback and was not rerun with iterative deepening. The appendix retains every allowance and direction.

The fitted seven-query D3 values, 32.69\% for Qwen and 36.55\% for GLM, underpredict same-model accuracy and overpredict cross-model accuracy. This change of sign reinforces the need to evaluate complete policies on the intended answer distribution. Appendix~\ref{app:confirmation-diagnostics} interprets this.

\section{Iterative deepening and path reuse under matched allowances}\label{app:anytime}
\paragraph{Design.} The original bounded planner computes a complete D1 fallback, then evaluates root candidates at the full horizon in greedy-first catalogue order. Only completed candidates can replace the fallback. When no depth-three root candidate completes, bounded D3 therefore returns the D1 action, even if a depth-two search would have completed.

The iterative-deepening variant (ID) keeps the same mandatory fallback, allowance, tie rule and discarding of interrupted values. For D3 it first completes depth two over all root candidates, then searches depth three with candidates ordered by their depth-two values. A partial depth-three pass may replace the depth-two decision only if its first candidate, the depth-two best, completed. Theorem~\ref{thm:anytime} gives this rule's per-decision contract.

Path reuse (R) keeps memo entries computed at a decision, including classifier predictions, visible to its descendants on the executed path and never to its siblings. This is the work an online agent retains between its own decisions. A cache hit charges no transitions, because the ancestor that computed the entry was charged. Deepening changes only D3; D1, D2 and rollout differ only by reuse.

\paragraph{Protocol and reproduction.} Both reruns use the fits, scorers, allowances and evaluation of the original noisy-grid and aligned matched stages. Their specifications and source hashes were frozen before any result with deepening or reuse. With both disabled, the implementation reproduced all 1,920 noisy-grid policies (identical utility and expected work) and all 120 aligned policies (identical episodes, accuracies and online transitions). Unit tests compare unbounded variants with exact D2, D3 and rollout, check that reused entries never cross to siblings, and check that an interrupted depth-three pass keeps the completed depth-two decision.

\paragraph{Results.} Tables~\ref{tab:anytime-noisy}--\ref{tab:anytime-contrasts} report all allowances and the prespecified contrasts.
\begin{itemize}
\item Deepening removes the fallback losses in both studies, and aligned D3+ID equals D2 at every budget and allowance.
\item With reuse, 100 of 775 seven-query decisions at 32,768 complete depth three. Accuracy nevertheless remains 99.5\%, against 100\% for unbounded aligned D3.
\item Reuse saves at most 3.4\% of noisy-grid work. On aligned CUBE-NM it saves 0.4--11\%, except 31\% for the original bounded D3 at $B=3$ and 32,768. The per-decision allowance, not total work, limits the root decision.
\end{itemize}
Transition counts are logical work, not floating-point operations, and seeds ran concurrently, so no timing claim is made.

\paragraph{Why value ordering is not enough.} At 32,768, value-ordered deepening is below the original bounded D3 on the noisy grid ($-.0130$, $[-.0167,-.0092]$). Context has little depth-two value, so value ordering searches it last, and the depth-two pass consumes part of the allowance. The original planner reaches context early only because the catalogue lists it first.

\paragraph{Latent-information ordering (prespecified).} This ordering keeps the depth-two best first and ranks the remaining depth-three candidates by $I(Z;O_j\mid h)$. Here $Z$ is the fitted mixture component and $O_j$ the answer to query $j$; ties fall back to the original greedy-first catalogue order.

Unlike depth-two value, this score does not vanish for queries whose value appears only further ahead. Computing it charges one transition per positive-probability outcome of each candidate, as for CMI, which is negligible: 32,995 against 32,997 expected transitions per episode at 32,768.

The protocol fixed the ordering, the comparators and the contrasts before any of these runs.
\begin{itemize}
\item \textbf{Noisy grid at 32,768.} The latent ordering reaches $.0611$, against $.0653$ for exact unbounded D3 and $.0280$ for D2, recovering 89\% of the exact depth-three gain. It selects context at the root in 50 of 160 fits (Tables~\ref{tab:anytime-latent} and~\ref{tab:anytime-latent-contrasts}).
\item \textbf{Permuted catalogue (context listed last).} The latent ordering is unchanged ($.0610$; difference $.0000$, $[-.0004,.0003]$). The catalogue-order variant loses its advantage ($.0353$), and the original bounded D3 falls to $.0335$.
\item \textbf{Smaller allowances.} At 4,096 and 512, the latent ordering equals value ordering and D2.
\item \textbf{Aligned CUBE-NM.} The latent ordering equals D2 at both budgets and every allowance, as expected given D2's half-point headroom at seven queries.
\end{itemize}
These results cover one synthetic family. The score also requires a fitted model whose latent components separate the relevant causes.

\begin{table}[htbp]
\centering\small
\setlength{\tabcolsep}{4pt}
\caption{Noisy-grid exact utility under the same per-decision allowances, 16 cells and ten fits. D3 (orig.) is the original bounded planner; ID adds iterative deepening and R adds path reuse. D2 and rollout include path reuse, which moves no mean by more than .0004. Work is expected online transitions per episode.}\label{tab:anytime-noisy}
\begin{tabular}{@{}rrrrrrrr@{}}
\toprule
Allowance & D2 & Rollout & D3 (orig.) & D3+ID & D3+ID+R & Work, orig. & Work, ID+R \\
\midrule
512 & .0005 & $-$.0001 & $-$.0001 & .0002 & .0005 & 465 & 481 \\
4,096 & .0280 & .0387 & .0004 & .0280 & .0280 & 4,033 & 4,872 \\
32,768 & .0280 & .0379 & .0529 & .0399 & .0403 & 34,179 & 32,997 \\
\bottomrule
\end{tabular}
\end{table}
\begin{table}[htbp]
\centering\footnotesize
\setlength{\tabcolsep}{4pt}
\caption{Classifier-aligned noiseless CUBE-NM accuracy (\%) under the same allowances, averaging five own-classifier fits. Columns follow Table~\ref{tab:anytime-noisy}. Work is mean online transitions per episode for D3+ID+R; H3 counts its decisions that complete depth three. Path reuse changes accuracy only for rollout at B7 and 32,768 (98.0 to 98.5).}\label{tab:anytime-aligned}
\begin{tabular}{@{}rrrrrrrrr@{}}
\toprule
$B$ & Allowance & D2 & Rollout & D3 (orig.) & D3+ID & D3+ID+R & Work & H3 \\
\midrule
3 & 512 & 31.0 & 31.0 & 31.0 & 31.0 & 31.0 & 1,024 & 0 \\
3 & 4,096 & 65.5 & 47.5 & 43.5 & 65.5 & 65.5 & 8,192 & 0 \\
3 & 32,768 & 72.0 & 72.0 & 43.5 & 72.0 & 72.0 & 45,035 & 0 \\
7 & 512 & 55.5 & 55.5 & 55.5 & 55.5 & 55.5 & 3,074 & 0 \\
7 & 4,096 & 98.0 & 82.5 & 61.5 & 98.0 & 98.0 & 24,576 & 0 \\
7 & 32,768 & 99.5 & 98.5 & 61.5 & 99.5 & 99.5 & 166,900 & 100 \\
\bottomrule
\end{tabular}
\end{table}
\begin{table}[htbp]
\centering\small
\setlength{\tabcolsep}{4pt}
\caption{Prespecified primary contrasts (D3+ID+R against D2+R and rollout+R) and the deepening effect on the original bounded D3. Noisy: ten replicate-block means, Bonferroni-6 (primary) or Bonferroni-12 (factorial) $t_9$ intervals. Aligned: five paired fits, Bonferroni-6 $t_4$ intervals, percentage points. Contrasts at 512 are zero or include zero and are not shown.}\label{tab:anytime-contrasts}
\begin{tabular}{@{}lrll@{}}
\toprule
Study & Allowance & Contrast & Mean [adjusted interval] \\
\midrule
Noisy & 4,096 & D3+ID+R $-$ D2+R & .0000 [.0000, .0000] \\
Noisy & 4,096 & D3+ID+R $-$ rollout+R & $-$.0107 [$-$.0160, $-$.0054] \\
Noisy & 32,768 & D3+ID+R $-$ D2+R & +.0123 [+.0074, +.0171] \\
Noisy & 32,768 & D3+ID+R $-$ rollout+R & +.0024 [$-$.0074, +.0121] \\
Noisy & 4,096 & D3+ID $-$ D3 (orig.) & +.0276 [+.0239, +.0312] \\
Noisy & 32,768 & D3+ID $-$ D3 (orig.) & $-$.0130 [$-$.0167, $-$.0092] \\
Aligned B7 & 4,096 & D3+ID+R $-$ D2+R & 0.0 [0.0, 0.0] \\
Aligned B7 & 4,096 & D3+ID+R $-$ rollout+R & +15.5 [+8.4, +22.6] \\
Aligned B7 & 4,096 & D3+ID $-$ D3 (orig.) & +36.5 [+24.0, +49.0] \\
Aligned B7 & 32,768 & D3+ID+R $-$ D2+R & 0.0 [0.0, 0.0] \\
Aligned B7 & 32,768 & D3+ID+R $-$ rollout+R & +1.0 [$-$2.0, +4.0] \\
Aligned B7 & 32,768 & D3+ID $-$ D3 (orig.) & +38.0 [+26.2, +49.8] \\
\bottomrule
\end{tabular}
\end{table}
\begin{table}[htbp]
\centering\footnotesize
\setlength{\tabcolsep}{4pt}
\caption{Noisy-grid utility at 32,768 transitions per decision, 160 fits, with the original catalogue (context listed first) and a permuted catalogue (context listed last). Context counts the fits whose root decision acquires context. Only the latent-information ordering keeps its utility when context moves.}\label{tab:anytime-latent}
\begin{tabular}{@{}lrrrr@{}}
\toprule
Method & Original & Context & Permuted & Context \\
\midrule
D2 & .0280 & 0 & .0280 & 0 \\
Rollout & .0379 & 1 & .0379 & 1 \\
D3, D1 fallback & .0529 & 53 & .0335 & 0 \\
Deepening, value order & .0403 & 7 & .0389 & 0 \\
Deepening, catalogue order (post hoc) & .0587 & 51 & .0353 & 0 \\
Deepening, latent order & .0611 & 50 & .0610 & 50 \\
\bottomrule
\end{tabular}
\end{table}
\begin{table}[htbp]
\centering\small
\setlength{\tabcolsep}{4pt}
\caption{Latent-ordering contrasts. Noisy and aligned B7 rows are the prespecified primary family (Bonferroni-9 $t_9$ and $t_4$; aligned in percentage points). Permuted and invariance rows are prespecified secondary contrasts (Bonferroni-15 $t_9$). Contrasts at 512 are zero or include zero.}\label{tab:anytime-latent-contrasts}
\begin{tabular}{@{}lrll@{}}
\toprule
Study & Allowance & Contrast & Mean [adjusted interval] \\
\midrule
Noisy & 4,096 & Latent $-$ D2 & .0000 [.0000, .0000] \\
Noisy & 4,096 & Latent $-$ rollout & $-$.0107 [$-$.0164, $-$.0050] \\
Noisy & 4,096 & Latent $-$ value order & .0000 [.0000, .0000] \\
Noisy & 32,768 & Latent $-$ D2 & +.0331 [+.0275, +.0387] \\
Noisy & 32,768 & Latent $-$ rollout & +.0232 [+.0116, +.0348] \\
Noisy & 32,768 & Latent $-$ value order & +.0208 [+.0170, +.0246] \\
Aligned B7 & 4,096 & Latent $-$ D2 & 0.0 [0.0, 0.0] \\
Aligned B7 & 4,096 & Latent $-$ rollout & +15.5 [+7.6, +23.4] \\
Aligned B7 & 4,096 & Latent $-$ value order & 0.0 [0.0, 0.0] \\
Aligned B7 & 32,768 & Latent $-$ D2 & 0.0 [0.0, 0.0] \\
Aligned B7 & 32,768 & Latent $-$ rollout & +1.0 [$-$2.3, +4.3] \\
Aligned B7 & 32,768 & Latent $-$ value order & 0.0 [0.0, 0.0] \\
Permuted & 32,768 & Latent $-$ catalogue order & +.0257 [+.0212, +.0302] \\
Permuted & 32,768 & Latent $-$ value order & +.0222 [+.0184, +.0260] \\
Permuted & 32,768 & Latent $-$ D2 & +.0330 [+.0269, +.0392] \\
Permuted & 32,768 & Catalogue order $-$ D3, D1 fallback & +.0018 [$-$.0020, +.0056] \\
Noisy & 32,768 & Latent, permuted $-$ original & .0000 [$-$.0004, +.0003] \\
\bottomrule
\end{tabular}
\end{table}
 
\paragraph{Compute curves and a tree-search baseline.} A further stage, frozen before its results, reran the bounded planners on all 160 noisy-grid fits at nine allowances from 512 to 131,072. At 512, 4,096 and 32,768 they reproduced the frozen results above to $10^{-12}$. The stage also added UCT, which builds a fresh tree per decision:
\begin{itemize}
\item it descends by UCB1 with exploration constant 1, sampling answers from the fitted model;
\item it scores each new node by its exact stop value;
\item it charges one unit per sampled query and per evaluation, and starts a simulation only if its worst-case cost fits in the allowance.
\end{itemize}
UCT chose to stop at the root in every fit at every allowance (utility $.0000$), $.0280$ below D2 at 4,096 and 32,768 ($[-.0316,-.0244]$). Appendix~\ref{app:tier1} shows that two choices mattered:
\begin{itemize}
\item \textbf{The exploration constant.} With $c=0.1$, the same stop-leaf UCT leaves the root on 33 fits and reaches $.0201$ at 32,768.
\item \textbf{The leaf value.} Exact rollout leaves raise UCT to about $.033$.
\end{itemize}

The frozen expectation was only partly met. Latent ordering is the best bounded method at 32,768 and 65,536, and ties the other deepening variants at 131,072. At 16,384, however, the original D1-fallback D3 is higher ($.0427$ against $.0388$), because the catalogue lists context first. The D1-fallback curve is also not the lowest at 8,192.
\begin{figure}[htbp]
\centering
\includegraphics[width=.82\linewidth]{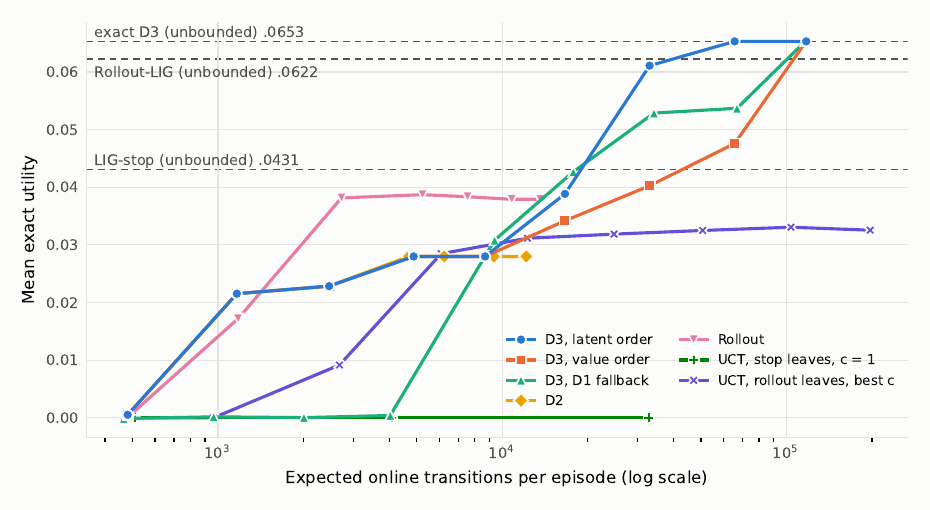}
\caption{Noisy-grid mean exact utility (160 fits) against expected search transitions per episode, at per-decision allowances from 512 to 131,072. Below 8,192 both deepening orders coincide with D2. Dashed lines are unbounded references.}\label{fig:compute-curves}
\end{figure}
\FloatBarrier
\section{One-step scores on the fitted latent component}\label{app:decision-aware}
\paragraph{Methods.} Let $Z$ be the fitted mixture component: a class--context pair on CUBE-NM and the noisy grid, and the whole feature vector in the saturated context-parity joint.
\begin{itemize}
\item \textbf{LIG} acquires $\arg\max_j I(Z;O_j\mid h)$, breaking ties by lowest index. It spends the budget like CMI and may abstain at the end on signed tasks.
\item \textbf{LIG-stop} follows the same query sequence. It stops when the exact value of continuing that sequence, computed by a recursion along it, does not exceed stopping now.
\item \textbf{Rollout-LIG} scores every candidate by one query followed by the LIG-stop continuation. On hard-budget CUBE-NM the continuation is forced LIG, scored either with the fitted MAP terminal or with the classifier-aligned terminal of Section~\ref{sec:confirmation-methods}.
\end{itemize}
Scoring one candidate charges one transition per positive-probability outcome, as for CMI. Work is total compilation transitions over the frozen policy tree, with caches shared across decisions for every method.

\paragraph{Protocol.} The methods, comparators, contrast families and the expected outcome were frozen with source hashes before any result.
\begin{itemize}
\item \textbf{Comparators} are the frozen noisy-grid, context-parity and noiseless CUBE-NM policies on the same fits, splits and scorers.
\item \textbf{Context-last control:} a catalogue with context moved last reruns every noisy-grid and parity comparator with the same exact planners.
\end{itemize}

\paragraph{Results.} Tables~\ref{tab:decision-noisy}--\ref{tab:decision-contrasts} report every method and contrast.
\begin{itemize}
\item \textbf{Noisy grid.} The expected ordering held: LIG-stop exceeds D1 and CMI, Rollout-LIG is at least LIG-stop under the fitted model, and both are unchanged under the context-last catalogue. Under the true generator Rollout-LIG is below LIG-stop in 4 of 480 noisy-grid and parity fits, a model-error effect outside the rollout theorem.
\item \textbf{CUBE-NM.} The fitted MAP terminal again misleads search: Rollout-LIG with that terminal scores below LIG itself (86.0\% against 96.0\% at seven queries), as ordinary D2 and D3 do relative to their aligned versions.
\item \textbf{Context parity (expectation failed).} The protocol predicted that LIG would succeed when context is listed first and fail when it is listed last, because ties would follow catalogue order. Instead it performs poorly under both catalogues: fitted noise breaks the ties, and maximum latent information in a saturated joint does not track decision relevance.
\end{itemize}

\paragraph{Scope.} These are synthetic CUBE families whose fitted mixture shares the generator's class--context structure, and a designed parity family. A model whose latent components do not separate the decision-relevant causes gives the score nothing to rank. On item-identification panels the latent component is the target itself, so the score equals CMI. Appendix~\ref{app:generic-latent} replaces these fits with a generic latent-class fit that is not told the class--context structure.

\begin{table}[htbp]
\centering\small
\setlength{\tabcolsep}{4pt}
\caption{Noisy CUBE-NM grid, 16 cells and ten fits, exact generator utility. LIG and CMI spend the whole budget and may abstain at the end. The last column reports whether the context-last catalogue gives the same mean to four decimals; fixed-subset and context-first controls were not rerun.}\label{tab:decision-noisy}
\begin{tabular}{@{}lrrrc@{}}
\toprule
Method & Utility & Expected queries & Work & Same, context last \\
\midrule
D1 & .0000 & 0.00 & 97 & yes \\
CMI & $-$.0780 & 3.00 & 1,683 & yes \\
LIG & $-$.0541 & 3.00 & 1,533 & yes \\
LIG-stop & .0431 & 0.87 & 1,573 & yes \\
D2 & .0280 & 0.68 & 14,856 & yes \\
Rollout & .0379 & 0.96 & 17,782 & yes \\
Rollout-LIG & .0622 & 0.96 & 61,110 & yes \\
D3 & .0653 & 1.16 & 738,271 & yes \\
Best fixed subset & .0135 & 0.96 & -- & -- \\
Paid context first & $-$.0412 & 1.01 & 360 & -- \\
\bottomrule
\end{tabular}
\end{table}
\begin{table}[htbp]
\centering\small
\setlength{\tabcolsep}{4pt}
\caption{Context-parity controls, eight cells and ten fits, exact generator utility under the original and context-last catalogues. The fitted model is a saturated joint, so the latent component is the whole feature vector and every feature carries the same latent information.}\label{tab:decision-parity}
\begin{tabular}{@{}lrr@{}}
\toprule
Method & Original & Context last \\
\midrule
D1 & .1077 & .1071 \\
LIG & .0428 & .0898 \\
LIG-stop & .0722 & .1132 \\
CMI & .2814 & .2820 \\
Rollout-LIG & .3751 & .3550 \\
Rollout & .4333 & .4310 \\
$D_B$ & .7838 & .7838 \\
\bottomrule
\end{tabular}
\end{table}
\begin{table}[htbp]
\centering\footnotesize
\setlength{\tabcolsep}{4pt}
\caption{Noiseless CUBE-NM own-classifier accuracy (\%), five fits, at each hard budget $B$. Comparators are the frozen primary policies on the same fits and scorers. The last two columns use the classifier-aligned terminal.}\label{tab:decision-cube}
\begin{tabular}{@{}rrrrrrrrrr@{}}
\toprule
$B$ & D1 & CMI & D2 & D3 & Rollout & LIG & RL (MAP) & Al.\ D3 & RL (aligned) \\
\midrule
3 & 28.0 & 31.5 & 53.0 & 53.0 & 53.0 & 60.0 & 54.5 & 72.0 & 68.5 \\
5 & 42.0 & 37.0 & 74.5 & 75.5 & 76.5 & 90.5 & 73.0 & 100.0 & 99.0 \\
7 & 50.0 & 56.0 & 77.5 & 80.5 & 82.5 & 96.0 & 86.0 & 100.0 & 99.5 \\
\bottomrule
\end{tabular}
\end{table}
\begin{table}[htbp]
\centering\footnotesize
\setlength{\tabcolsep}{4pt}
\caption{Prespecified contrasts. Primary: noisy grid Bonferroni-4 $t_9$ over ten replicate blocks; CUBE-NM B7 Bonferroni-3 $t_4$ over five fits, in percentage points. Secondary families use the same adjustments and are descriptive. Context-last noisy contrasts equal the original ones and are omitted.}\label{tab:decision-contrasts}
\begin{tabular}{@{}llll@{}}
\toprule
Family & Study & Contrast & Mean [adjusted interval] \\
\midrule
Primary & Noisy & Rollout-LIG $-$ D3 & $-$.0031 [$-$.0080, +.0018] \\
Primary & Noisy & Rollout-LIG $-$ rollout & +.0243 [+.0136, +.0351] \\
Primary & Noisy & LIG-stop $-$ D1 & +.0431 [+.0368, +.0494] \\
Primary & Noisy & LIG-stop $-$ CMI & +.1211 [+.1137, +.1285] \\
Primary & CUBE-NM B7 & LIG $-$ CMI & +40.0 [+9.2, +70.8] \\
Primary & CUBE-NM B7 & LIG $-$ D3 & +15.5 [+10.6, +20.4] \\
Primary & CUBE-NM B7 & Rollout-LIG (aligned) $-$ aligned D3 & $-$0.5 [$-$2.5, +1.5] \\
Secondary & Parity & Rollout-LIG $-$ $D_B$ & $-$.4088 [$-$.4388, $-$.3788] \\
Secondary & Parity & LIG-stop $-$ $D_B$ & $-$.7117 [$-$.7435, $-$.6798] \\
Secondary & Parity, context last & Rollout-LIG $-$ $D_B$ & $-$.4288 [$-$.4674, $-$.3902] \\
Secondary & Parity, context last & LIG-stop $-$ $D_B$ & $-$.6707 [$-$.7039, $-$.6374] \\
Secondary & CUBE-NM B3 & LIG $-$ CMI & +28.5 [+21.8, +35.2] \\
Secondary & CUBE-NM B3 & LIG $-$ D3 & +7.0 [$-$3.1, +17.1] \\
Secondary & CUBE-NM B3 & Rollout-LIG (aligned) $-$ aligned D3 & $-$3.5 [$-$8.5, +1.5] \\
Secondary & CUBE-NM B5 & LIG $-$ CMI & +53.5 [+31.2, +75.8] \\
Secondary & CUBE-NM B5 & LIG $-$ D3 & +15.0 [+4.6, +25.4] \\
Secondary & CUBE-NM B5 & Rollout-LIG (aligned) $-$ aligned D3 & $-$1.0 [$-$3.4, +1.4] \\
\bottomrule
\end{tabular}
\end{table}
 Table~\ref{tab:traces} shows the CUBE-NM mechanism: both planners ask for the context, but MAP-terminal D3 then spends two questions on a block the context has ruled out.

\begin{table}[htbp]
\centering\footnotesize
\setlength{\tabcolsep}{3pt}
\caption{Decision traces replayed from saved runs. CUBE-NM: fit 52142, B7; $\star$ marks a position that carries the class in that context. Common: Qwen bank, fit and examiner, B7. On this examiner replicate CMI alone is right on 21 items and D3 alone on 7; we show one of each.}\label{tab:traces}
\begin{tabularx}{\linewidth}{@{}l>{\raggedright\arraybackslash}Xl@{}}
\toprule
Planner & Questions asked, in order (answer) & Outcome \\
\midrule
\multicolumn{3}{@{}l}{\emph{CUBE-NM: class 6, context 1}} \\
LIG & context; block 1 positions 6$\star$, 5, 8$\star$, 9, 7$\star$, 2 & correct \\
D3 & context; block 1 positions 6$\star$, 5, 3, 1; block 0 positions 0, 1 & wrong \\
\midrule
\multicolumn{3}{@{}l}{\emph{Common: the item is Otter}} \\
CMI & found in the wild? yes; animal? yes; aquatic? yes; durable? no; savory? no; flies? no; terrestrial? no & Blobfish \\
D3 & living organism? yes; aquatic? yes; cold climates? yes; eaten for sustenance? no; terrestrial? no; North America? yes; smaller than a laptop? yes & Otter \\
\midrule
\multicolumn{3}{@{}l}{\emph{Common: the item is Sloth}} \\
CMI & found in the wild? yes; animal? yes; aquatic? no; hot climates? no; flies? no; smaller than a laptop? yes; North America? no & Sloth \\
D3 & living organism? yes; aquatic? no; hot climates? no; outside the Americas? no; animal? yes; flies? no; found in the wild? yes & Raccoon \\
\bottomrule
\end{tabularx}
\end{table}
\FloatBarrier
\section{ECED, catalogue order and a stronger tree search}\label{app:tier1}
This appendix answers three questions about the baselines of Section~\ref{sec:main-decision-aware} and Figure~\ref{fig:compute-curves}.
\begin{itemize}
\item Does a published decision-aware score do what the latent score does?
\item Does LIG's CUBE-NM result depend on the catalogue listing the context first?
\item Does UCT stop at the root only because of its leaves and exploration constant?
\end{itemize}
The methods, comparators, contrast families and expectations were frozen with source hashes before any result.

\paragraph{ECED.} ECED \citep{chen2017eced} defines a score over root causes $\theta$ and their targets $r(\theta)$. Here the root causes are the fitted mixture components and the targets are their labels.
\begin{itemize}
\item \textbf{Edges} join components with different labels and carry weight $P[\theta,h]\,P[\theta',h]$.
\item \textbf{Likelihood ratio:} $\lambda_{\theta,x}=P[x\mid\theta]/\max_{x'}P[x'\mid\theta]$.
\item \textbf{Score of query $j$:} the expected value over answers $x$ of
\[
\sum_{\{\theta,\theta'\}}w_\theta w_{\theta'}\bigl(1-\lambda_{\theta,x}\lambda_{\theta',x}\bigr)\;-\;\sum_{\{\theta,\theta'\}}w_\theta w_{\theta'}\bigl(1-\max_{\theta''}\lambda_{\theta'',x}^2\bigr),
\]
where the sums run over edges and $w$ is the posterior. The first term is the discounted edge weight removed; the second is the offset, the reduction any answer guarantees.
\item \textbf{Equivalence:} with noiseless answers the offset vanishes and the score is the expected weight of edges cut, as in EC$^2$ \citep{golovin2010}. This objective is adaptive monotone and adaptive submodular (Theorem~\ref{thm:ec2}), even on noiseless parity, whereas information about the label is not: on noiseless context parity with two-bit blocks, a bit's label information is zero until the context and the rest of its block are known, and then one bit. Greedy cutting identifies the decision with probability $1-\delta$ within $k\ln(1/(\delta\eta))$ tests when some policy always identifies it within $k$ (Theorem~\ref{thm:ec2-greedy}). On that task $k=3$ and $\eta=1/1024$, so at $\delta=.1$ the bound is about 28 tests, against a budget of three.
\end{itemize}
ECED, ECED-stop and Rollout-ECED replace LIG's score in the three policies of Appendix~\ref{app:decision-aware} and are charged the same work. Tests check the score against an explicit pairwise sum and the noiseless equivalence.

\paragraph{Catalogue control on CUBE-NM.} Features 1--50 are listed first and the context last, so ties favour the lowest permuted index. Planners run on the permuted model, and a wrapper maps actions and histories back to the original indices for execution and for the classifier. Every comparator is rerun, except aligned D3 at $B=7$, whose cost (about 78 CPU-minutes per fit) the protocol excludes.

\paragraph{UCT.} Every variant charges one unit per sampled query and per evaluation, as in Appendix~\ref{app:anytime}.
\begin{itemize}
\item \textbf{Stop leaves} score a new node by its exact stop value. They were rerun with $c\in\{0.03,0.1,0.3,1\}$; with $c=1$ they reproduce the UCT of Appendix~\ref{app:anytime} on all 160 fits to $10^{-12}$.
\item \textbf{Rollout leaves} score a new node by the exact value of the D1-policy continuation from it. That evaluation costs one unit plus every transition the continuation charges, and caches are cleared at each decision. The continuation runs under the unspent allowance; a simulation that would exceed it is abandoned without backup. Rollout leaves ran at all nine allowances with the same four constants.
\end{itemize}

\paragraph{Results.} Table~\ref{tab:tier1-contrasts} gives the primary contrasts.
\begin{itemize}
\item \textbf{Catalogue order does not matter} (Table~\ref{tab:tier1-cube}). With the context last, LIG and ECED still ask for it first in every fit at every budget, and every accuracy is unchanged.
\item \textbf{The comparators do not move either.} CMI and D1 choose the same root features as before.
\item \textbf{ECED on the noisy grid and CUBE-NM} (Table~\ref{tab:tier1-eced}). ECED is close to LIG. With rollout the two are indistinguishable; with exact stopping alone ECED is $.0097$ lower ($[-.0146,-.0047]$). Like LIG, ECED asks for the context first when it spends the budget.
\item \textbf{ECED on context parity.} ECED beats LIG under both catalogues. Its edges join only differently labelled components, so its score carries label information that the saturated latent score lacks. Rollout-ECED is nonetheless $.2905$ below full-horizon search ($[-.3472,-.2338]$, Bonferroni-2 $t_9$).
\item \textbf{UCT with rollout leaves} leaves the root and plateaus near $.033$ (Table~\ref{tab:tier1-uct}). That is above D2 from 8,192 on, but below exact one-step rollout at every allowance, although rollout uses the same leaf values with exact expectations.
  \begin{itemize}
  \item It still stops at the root on 111--126 of 160 fits at 4,096 and above. Of those stops, 110 of 117 at 32,768 are on cells 0--11, where no bounded method averages more than $.018$.
  \item One exact D1 continuation costs about 500 units, so at 512 every search abandons its first leaf.
  \end{itemize}
\item \textbf{UCT with stop leaves} leaves the root once the exploration constant is small: with $c=0.1$ it reaches $.0201$ at 32,768.
\end{itemize}

\paragraph{Expectations.} The frozen expectations held, with one exception.
\begin{itemize}
\item \textbf{Held:} LIG keeps its CUBE-NM accuracy with the context last. ECED is within $.01$ of LIG on the noisy grid in point estimate. Tuned rollout-leaf UCT reaches D2 by 32,768 but stays below latent-ordered D3 from there. Exploration constants below 1 do better.
\item \textbf{Failed:} we expected ECED to follow catalogue order on parity, like LIG. It did not.
\end{itemize}

\paragraph{Scope.} The tuned exploration constant is chosen on the evaluation fits, so it favours UCT. UCT's expected online work per episode can exceed the latent-ordered D3's at the same per-decision allowance; at 32,768 it is 50,735 against 32,995. ECED is evaluated only on the fitted components; with components that do not separate the decision-relevant causes, it too has nothing to rank.

\begin{table}[htbp]
\centering\footnotesize
\setlength{\tabcolsep}{4pt}
\caption{Primary contrasts with family-wise 95\% intervals: three ECED contrasts (ten replicate blocks, Bonferroni-3 $t_9$), three context-last CUBE-NM contrasts in accuracy points (five fits, Bonferroni-3 $t_4$), and three UCT contrasts at the best exploration constant for each allowance (Bonferroni-3 $t_9$).}\label{tab:tier1-contrasts}
\begin{tabular}{@{}llr@{}}
\toprule
Setting & Contrast & Mean [interval] \\
\midrule
Noisy grid & Rollout-ECED $-$ Rollout-LIG & $-$.0035 [$-$.0081, +.0011] \\
Noisy grid & ECED-stop $-$ LIG-stop & $-$.0097 [$-$.0146, $-$.0047] \\
Noisy grid & Rollout-ECED $-$ exact D3 & $-$.0065 [$-$.0120, $-$.0011] \\
CUBE-NM $B=7$ & LIG $-$ CMI (context last) & +40.0 [+9.2, +70.8] \\
CUBE-NM $B=7$ & LIG $-$ D3 (context last) & +15.5 [+10.6, +20.4] \\
CUBE-NM $B=7$ & LIG, context last $-$ original & 0.0 [0.0, 0.0] \\
Noisy grid, 4,096 & UCT, rollout leaves $-$ D3 latent order & +.0005 [$-$.0099, +.0109] \\
Noisy grid, 32,768 & UCT, rollout leaves $-$ D3 latent order & $-$.0286 [$-$.0334, $-$.0238] \\
Noisy grid, 131,072 & UCT, rollout leaves $-$ D3 latent order & $-$.0327 [$-$.0443, $-$.0212] \\
\bottomrule
\end{tabular}
\end{table}
\begin{table}[htbp]
\centering\footnotesize
\setlength{\tabcolsep}{4pt}
\caption{ECED against the latent-information score under the same wrappers. Noisy grid and context parity: mean exact utility (160 and 80 fits). CUBE-NM: own-classifier accuracy (\%, five fits) for spend-budget and, in the rollout rows, the classifier-aligned terminal. Exact depth is D3 on the noisy grid, full-horizon $D_B$ on parity and aligned D3 on CUBE-NM. Noisy-grid values are identical under the context-last catalogue.}\label{tab:tier1-eced}
\begin{tabular}{@{}llrrr@{\hspace{14pt}}rrr@{}}
\toprule
& & Noisy & Parity & Parity, last & B3 & B5 & B7 \\
\midrule
Spend budget & ECED & $-$.0716 & .2511 & .2609 & 57.5 & 93.0 & 95.5 \\
 & LIG & $-$.0541 & .0428 & .0898 & 60.0 & 90.5 & 96.0 \\
Exact stopping & ECED & .0334 & .2708 & .2807 & -- & -- & -- \\
 & LIG & .0431 & .0722 & .1132 & -- & -- & -- \\
One-step rollout & ECED & .0587 & .4933 & .4933 & 68.5 & 98.0 & 100.0 \\
 & LIG & .0622 & .3751 & .3550 & 68.5 & 99.0 & 99.5 \\
Exact depth & D3 / $D_B$ / al.\ D3 & .0653 & .7838 & .7838 & 72.0 & 100.0 & 100.0 \\
\bottomrule
\end{tabular}
\end{table}
\begin{table}[htbp]
\centering\footnotesize
\setlength{\tabcolsep}{4pt}
\caption{Noiseless CUBE-NM own-classifier accuracy (\%, five fits) with the context listed first (orig.) and last. Aligned D3 was not rerun at $B=7$ with the context last (n.r.) because of its cost.}\label{tab:tier1-cube}
\begin{tabular}{@{}lrrrrrr@{}}
\toprule
& \multicolumn{2}{c}{$B=3$} & \multicolumn{2}{c}{$B=5$} & \multicolumn{2}{c}{$B=7$} \\
Method & orig. & last & orig. & last & orig. & last \\
\midrule
LIG & 60.0 & 60.0 & 90.5 & 90.5 & 96.0 & 96.0 \\
Rollout-LIG (MAP) & 54.5 & 54.5 & 73.0 & 73.0 & 86.0 & 86.0 \\
Rollout-LIG (aligned) & 68.5 & 68.5 & 99.0 & 99.0 & 99.5 & 99.5 \\
ECED & 57.5 & 57.5 & 93.0 & 93.0 & 95.5 & 95.5 \\
Rollout-ECED (aligned) & 68.5 & 68.5 & 98.0 & 98.0 & 100.0 & 100.0 \\
D1 & 28.0 & 28.0 & 42.0 & 42.0 & 50.0 & 50.0 \\
CMI & 31.5 & 31.5 & 37.0 & 37.0 & 56.0 & 56.0 \\
D2 & 53.0 & 53.0 & 74.5 & 74.5 & 77.5 & 77.5 \\
D3 & 53.0 & 53.0 & 75.5 & 75.5 & 80.5 & 80.5 \\
Rollout & 53.0 & 53.0 & 76.5 & 76.5 & 82.5 & 82.5 \\
Aligned D2 & 72.0 & 72.0 & 98.0 & 98.0 & 99.5 & 99.5 \\
Aligned D3 & 72.0 & 72.0 & 100.0 & 100.0 & 100.0 & n.r. \\
\bottomrule
\end{tabular}
\end{table}
\begin{table}[htbp]
\centering\footnotesize
\setlength{\tabcolsep}{4pt}
\caption{UCT on the noisy grid: mean exact utility (160 fits) by leaf value and exploration constant $c$. Rollout leaves use the exact D1-policy continuation. The last two columns are for rollout leaves with the best $c$: fits whose root decision is to stop, and decisions whose last simulation was abandoned at the allowance.}\label{tab:tier1-uct}
\begin{tabular}{@{}rrrrr@{\hspace{12pt}}rrrr@{\hspace{12pt}}rr@{}}
\toprule
& \multicolumn{4}{c}{Rollout leaves, $c=$} & \multicolumn{4}{c}{Stop leaves, $c=$} & & \\
Allowance & .03 & .1 & .3 & 1 & .03 & .1 & .3 & 1 & Root stop & Aband. \\
\midrule
512 & .0000 & .0000 & .0000 & .0000 & .0000 & .0016 & .0007 & .0000 & 160 & 160 \\
1,024 & .0000 & .0000 & .0000 & .0000 & -- & -- & -- & -- & 160 & 153 \\
2,048 & .0076 & .0092 & .0028 & .0002 & -- & -- & -- & -- & 129 & 228 \\
4,096 & .0192 & .0254 & .0285 & .0059 & .0000 & .0020 & .0008 & .0000 & 117 & 179 \\
8,192 & .0255 & .0284 & .0311 & .0265 & -- & -- & -- & -- & 120 & 171 \\
16,384 & .0251 & .0301 & .0319 & .0311 & -- & -- & -- & -- & 119 & 159 \\
32,768 & .0252 & .0319 & .0325 & .0314 & .0001 & .0201 & .0076 & .0000 & 117 & 150 \\
65,536 & .0235 & .0305 & .0331 & .0313 & -- & -- & -- & -- & 113 & 149 \\
131,072 & .0240 & .0325 & .0272 & .0237 & -- & -- & -- & -- & 120 & 25 \\
\bottomrule
\end{tabular}
\end{table}
 \FloatBarrier
\section{Depth on local real datasets and a deepen-when-needed rule}\label{app:tier2}
This appendix asks four questions on real tabular data:
\begin{itemize}
\item Does exact three-step search beat two-step search held out?
\item Does a one-step latent score with stopping and rollout recover depth's gain?
\item Does a rule that deepens only when needed keep D3's utility at less work?
\item Does depth's gain under the fitted model predict its held-out gain?
\end{itemize}
Methods, tasks, splits, contrasts and expectations were frozen with source hashes before any policy result.

\paragraph{Data and tasks.} We used only datasets already on disk:
\begin{itemize}
\item scikit-learn Iris, Wine, Breast Cancer and Digits;
\item scikit-learn Diabetes, with the regression target split at its median;
\item the three-class Diabetes, PhysioNet and MiniBooNE CSVs distributed with AFABench;
\item UCI Zoo.
\end{itemize}
A task is a dataset and at most ten features whose median split leaves at least 5\% on each side. Datasets with more such features get four random feature subsets, drawn without labels. That gives 27 tasks, each with five stratified 70/30 splits, for 135 fits.

Features are binarized at the training median; missing values map to 0. Each class is fitted by a Bernoulli mixture with $K\in\{1,2,3,4\}$ components, chosen by BIC. With $K=1$ the fit is naive Bayes and the latent score equals label information. BIC chose $K=1$ in 41 fits, 2 in 9, 3 in 18 and 4 in 67.

Policies use signed utility, a budget of three and query cost .01, so D3 is the exact full-horizon optimum under the fit. They are scored exactly on the held-out rows and under the fit.

\paragraph{Deepen when needed.} At each history this rule runs exact depth-three search only when the best fixed set of queries beats Rollout-LIG's own value by more than $.002$ (the quantity of Theorem~\ref{thm:interaction-sufficiency}). Appendix~\ref{app:order-rule} replaces it with a rule applied once per fit.

\paragraph{Results.} Table~\ref{tab:tier2-datasets} reports held-out utility by dataset; Table~\ref{tab:tier2-contrasts} gives the contrasts.
\begin{itemize}
\item \textbf{Depth beyond two queries rarely pays.} Held-out D3 $-$ D2 is $+.0100$ $[-.0125,+.0324]$. The gain comes from Digits ($+.063$) and, less, Wine ($+.028$). Over tasks rather than task--split pairs, whose splits share data, both intervals include zero ($[-.006,.133]$ and $[-.021,.076]$, four tasks each; post hoc). On Digits, with ten classes, a prediction is worth making only once the posterior exceeds one half, which takes about three pixels. On PhysioNet and the three-class Diabetes every method ties with D1.
\item \textbf{Myopia matters more than depth.} On MiniBooNE, D1 asks 0.44 queries on average and scores $.475$. D2, D3 and rollout reach $.593$, and D3 adds nothing over D2.
\item \textbf{The latent score with rollout recovers depth's gain where there is one.}
  \begin{itemize}
  \item On Digits Rollout-LIG is $.005$ below D3 at half D3's work.
  \item Under the fit, Rollout-LIG $-$ D3 is $-.0067$ $[-.0141,+.0007]$.
  \item Held out the interval is wide because of model error. On Iris (45 test rows per split), the naive-Bayes fit ranks D3 above Rollout-LIG ($.657$ against $.639$), but held out they score $.451$ against $.585$.
  \end{itemize}
\item \textbf{The fitted depth gain predicts the held-out gain.} Across the 27 tasks, Spearman's $\rho$ is $.61$.
\item \textbf{Deepening when needed matches D3 but saves no work here.} It is $.0021$ below D3 held out ($[-.0090,+.0049]$), deepens at 93 of 649 compiled decisions, and uses 1.19 times D3's work, because exact D3 is cheap with ten binary features and a budget of three.
\item \textbf{ECED again behaves like LIG:} Rollout-ECED $-$ Rollout-LIG is $+.0017$ $[-.0065,+.0099]$.
\end{itemize}

\paragraph{Expectations.}
\begin{itemize}
\item \textbf{Met.} Held-out D3 $-$ D2 lies within $\pm.01$ on seven of nine datasets, and the fitted and held-out depth gains are positively rank-correlated.
\item \textbf{Partly met.} Rollout-LIG is within .01 of D3 held out on seven of nine datasets; the exceptions are Iris (model error) and Zoo ($-.019$).
\item \textbf{Failed:}
  \begin{itemize}
  \item The rule deepened at 14\% of decisions rather than under 10\%, and it used more work than D3 rather than under half.
  \item Rollout-LIG recovered less of D3's fitted gain over D1 where $K>1$ (54\%, 9 tasks) than where $K=1$ (78\%, 7 tasks), not more. On one MiniBooNE task it trails D3 by $.04$: the fitted components there separate structure that does not bear on the decision.
  \end{itemize}
\end{itemize}

\paragraph{Scope.}
\begin{itemize}
\item These are nine local datasets, not a population. Tasks within a dataset share data, so inference is at the dataset level with $n=9$, and intervals are wide.
\item Median binarization, ten features and a budget of three keep the interaction structure coarse.
\item BIC chose the largest $K$ on every large dataset.
\end{itemize}

\begin{table}[htbp]
\centering\footnotesize
\setlength{\tabcolsep}{4pt}
\caption{Held-out signed utility on nine local datasets: means over each dataset's tasks and five splits (budget three, query cost .01). The last row is mean compilation transitions per fit. RL is Rollout-LIG and RE Rollout-ECED. Deepen runs exact D3 only where a fixed set of at most three queries beats Rollout-LIG by more than .002.}\label{tab:tier2-datasets}
\begin{tabular}{@{}lrrrrrrrr@{}}
\toprule
Dataset & D1 & CMI & D2 & D3 & Rollout & RL & RE & Deepen \\
\midrule
Iris & .5138 & .6144 & .4507 & .4505 & .4507 & .5855 & .6036 & .4506 \\
Wine & .7985 & .7756 & .7739 & .8016 & .8075 & .8056 & .7987 & .8056 \\
Breast cancer & .7985 & .7829 & .7926 & .7844 & .7820 & .7873 & .7918 & .7858 \\
Digits & .0000 & .0202 & .0024 & .0656 & .0068 & .0604 & .0667 & .0604 \\
Diabetes (sklearn) & .4822 & .4798 & .4865 & .4865 & .4863 & .4875 & .4891 & .4865 \\
Diabetes (3-class) & .6166 & .5924 & .6162 & .6162 & .6162 & .6166 & .6166 & .6166 \\
PhysioNet & .7156 & .6856 & .7156 & .7156 & .7156 & .7156 & .7156 & .7156 \\
MiniBooNE & .4751 & .5809 & .5929 & .5930 & .5930 & .5856 & .5782 & .5930 \\
Zoo & .7030 & .6926 & .6885 & .6958 & .6702 & .6765 & .6758 & .6765 \\
\midrule
\textbf{Mean of datasets} & .5670 & .5805 & .5688 & .5788 & .5698 & .5912 & .5929 & .5767 \\
Work (transitions) & 62 & 111 & 759 & 2,915 & 1,057 & 1,386 & 1,381 & 3,649 \\
\bottomrule
\end{tabular}
\end{table}
\begin{table}[htbp]
\centering\footnotesize
\setlength{\tabcolsep}{4pt}
\caption{Contrasts on the nine local datasets: dataset-level means, $t_8$ intervals with the listed Bonferroni multiplicity. Across the 27 tasks, fitted and held-out D3 $-$ D2 have Spearman correlation .61 ($p=.0007$). The deepen-when-needed rule (Deepen) deepened at 93 of 649 compiled decisions and used 1.19 times D3's work.}\label{tab:tier2-contrasts}
\begin{tabular}{@{}lllrr@{}}
\toprule
Family & Contrast & Scale & Mult. & Mean [interval] \\
\midrule
Primary & D3 $-$ D2 & held-out & 3 & +.0100 [$-$.0125, +.0324] \\
Primary & Rollout-LIG $-$ D3 & held-out & 3 & +.0124 [$-$.0344, +.0591] \\
Primary & Deepen $-$ D3 & held-out & 3 & $-$.0021 [$-$.0090, +.0049] \\
Secondary & D3 $-$ D2 & fitted & 3 & +.0075 [$-$.0103, +.0252] \\
Secondary & Rollout-LIG $-$ D3 & fitted & 3 & $-$.0067 [$-$.0141, +.0007] \\
Secondary & Deepen $-$ D3 & fitted & 3 & $-$.0011 [$-$.0028, +.0006] \\
Secondary & Rollout-LIG $-$ rollout & held-out & 4 & +.0214 [$-$.0280, +.0707] \\
Secondary & LIG-stop $-$ D1 & held-out & 4 & $-$.0067 [$-$.1016, +.0883] \\
Secondary & Rollout-ECED $-$ Rollout-LIG & held-out & 4 & +.0017 [$-$.0065, +.0099] \\
Secondary & D3 $-$ rollout & held-out & 4 & +.0090 [$-$.0131, +.0311] \\
\bottomrule
\end{tabular}
\end{table}
 \FloatBarrier
\section{Planning over several model families}\label{app:families}
Policies fitted to one model's Common answers lose most of their accuracy on another model's answers (Section~\ref{sec:main-results}). This appendix asks two questions:
\begin{itemize}
\item Does planning over the plausible answering models avoid that collapse?
\item How well do the fitted answer channels predict held-out answers?
\end{itemize}
It uses only the saved Common panels: two banks, Qwen and GLM, six fitting and three examiner answers per item. No new model calls were made. The fits, contrasts and expectations were frozen, with the panels' hashes, before any new policy was compiled. Every single-family accuracy reproduces Table~\ref{tab:llm-fresh} to two decimals.

\paragraph{Fits.} All use concentration one, seven forced queries and MAP-item accuracy.
\begin{itemize}
\item \textbf{Single-family:} the paper's Qwen and GLM fits.
\item \textbf{Pooled:} one Beta--Bernoulli fit on both families' twelve answers per item.
\item \textbf{Mixture:} one component per item and family, each with that family's fitted channel and prior one half. The label is the item. The planner marginalizes over which family answers and infers it from the answers.
\end{itemize}

\paragraph{Accuracy.} Table~\ref{tab:family-accuracy} reports every fit, method and examiner.
\begin{itemize}
\item \textbf{The mixture avoids the collapse.} Mixture D3 reaches 44.14\% on Qwen answers and 46.55\% on GLM answers. That is 25.5 and 28.2 points above the other family's fit, and within 4 points of the same-family fit, with intervals that include zero.
\item \textbf{Inferring the family beats pooling answers.} Pooled D3 reaches 42.04\% and 42.94\%.
\item \textbf{Depth adds little inside the mixture.} Mixture D3 is 3.6 and 3.2 points above mixture D2, with intervals that include zero.
\item \textbf{Robustness has a price.} The mixture's rollout and CMI are 6--9 points below their same-family counterparts.
\end{itemize}

\paragraph{Calibration.} Table~\ref{tab:family-calibration} scores each fit's predicted yes-probability against held-out examiner answers.
\begin{itemize}
\item \textbf{Cross-family channels are confidently wrong.} Their log loss is 2.4--3.1 times the same-family value and their Brier score 3--7 times. Yet their expected calibration error barely differs on the GLM examiner (.063 against .062).
\item \textbf{So the two families answer some questions differently with high confidence.} A single-family planner's sharp posterior then rules out the true item.
\item \textbf{The mixture's averaged channel calibrates worse than the pooled fit} in log loss and ECE. It still plans better, because its posterior identifies the family.
\end{itemize}

\paragraph{Expectations.} The frozen expectations held, with one exception.
\begin{itemize}
\item \textbf{Held:}
  \begin{itemize}
  \item mixture D3 within 8 points of same-family D3 and at least 15 points above cross-family D3;
  \item pooled between same-family and cross-family;
  \item cross-family log loss and ECE above same-family.
  \end{itemize}
\item \textbf{Failed:} on the GLM examiner the mixture's ECE exceeds both single fits'.
\end{itemize}

\paragraph{Scope.}
\begin{itemize}
\item Both families were seen in fitting. An unseen family needs new answer panels and model calls.
\item The ensemble is a Bayesian mixture, not max-min robust planning.
\item Calibration concerns channels estimated from sampled answers, not probabilities stated by the models.
\item With two banks and three examiner repeats, intervals are wide.
\end{itemize}

\begin{table}[htbp]
\centering\footnotesize
\setlength{\tabcolsep}{4pt}
\caption{Common identification accuracy (\%, equal-bank mean) at seven queries for four fits, on Qwen (Q) and GLM (G) examiner answers. Pooled fits both families' twelve answers per item; Mixture has one component per item and family and infers the family from the answers. Mixture D3 minus same-family D3: $-$3.60 [$-$12.54, +5.11] on Q and $-$3.90 [$-$10.96, +3.30] on G. Minus cross-family D3: +25.53 [+18.17, +32.43] and +28.23 [+19.67, +37.09]. Paired target bootstrap, Bonferroni-4, percentage points.}\label{tab:family-accuracy}
\begin{tabular}{@{}lrrrrrrrr@{}}
\toprule
 & \multicolumn{2}{c}{Qwen} & \multicolumn{2}{c}{GLM} & \multicolumn{2}{c}{Pooled} & \multicolumn{2}{c}{Mixture} \\
Method & Q & G & Q & G & Q & G & Q & G \\
\midrule
D1 & 32.73 & 16.82 & 14.11 & 37.69 & 32.73 & 37.84 & 33.63 & 39.34 \\
D2 & 38.59 & 16.52 & 16.22 & 44.44 & 38.89 & 41.14 & 40.54 & 43.39 \\
D3 & 47.75 & 18.32 & 18.62 & 50.45 & 42.04 & 42.94 & 44.14 & 46.55 \\
Rollout & 53.30 & 16.67 & 15.92 & 52.10 & 45.65 & 40.69 & 47.15 & 43.69 \\
CMI & 52.85 & 20.42 & 17.42 & 51.50 & 41.59 & 41.29 & 43.99 & 42.19 \\
\bottomrule
\end{tabular}
\end{table}
\begin{table}[htbp]
\centering\footnotesize
\setlength{\tabcolsep}{4pt}
\caption{Calibration of each fit's answer channel on held-out examiner answers (equal-bank means over 111 items, 32 questions and three repeats). ECE uses ten equal-width bins. For the mixture, the channel is the family-averaged prediction before any answer is seen.}\label{tab:family-calibration}
\begin{tabular}{@{}llrrr@{}}
\toprule
Fit & Examiner & Log loss & Brier & ECE \\
\midrule
Qwen & Qwen & .092 & .012 & .069 \\
Qwen & GLM & .320 & .089 & .063 \\
GLM & Qwen & .287 & .080 & .103 \\
GLM & GLM & .136 & .028 & .062 \\
Pooled & Qwen & .114 & .028 & .068 \\
Pooled & GLM & .148 & .041 & .059 \\
Mixture & Qwen & .145 & .031 & .098 \\
Mixture & GLM & .175 & .044 & .085 \\
\bottomrule
\end{tabular}
\end{table}
 \FloatBarrier
\section{Powered language-model panels}\label{app:powered-panels}
\emph{Problem.} The Common panels have 111 items and two banks, so most of their primary intervals span 10--16 points. Their per-item noise is almost all per bank and item, and independent across banks, so more banks and items add power where more examiner repeats would not.

\emph{Approach.} We wrote two new 32-question banks for the 111 Common items. We also added a second vocabulary: the 188 unique items of THING200 from the same UoT file \citep{hu2024uot}, with two banks. In each vocabulary Qwen wrote one bank and GLM the other. Both families answered every bank with six fitting and three examiner requests per item. The prompts, decoding, fits, planners, family mixture and replay are those used on the first panels. That gives 598 new bank--item units, against 222 there.

The protocol, contrasts and six expectations were frozen before any request. All 10{,}764 answer requests were valid at their first attempt. Every replayed accuracy matches the exact empirical-mixture accuracy to $10^{-15}$.

The primary analysis is D3$-$D2 and D3$-$rollout at B7 with single-family fits, in the four fit--examiner directions, over the 598 new units. Its intervals use 10{,}000 bootstrap draws that resample items within strata and keep an item's banks together, with Bonferroni-8.

\emph{Evidence} (Table~\ref{tab:powered-panels}). All six expectations held.
\begin{itemize}
\item \textbf{Depth pays within a family, by less than the first panels suggested.} D3 beats D2 by 3.7 points for Qwen and 4.5 for GLM; the earlier Qwen estimate was 9.2. Across families the difference is under 2 points, with intervals that include zero.
\item \textbf{Rollout's lead over Qwen's D3 replicates.} It is 6.2 points, and all of it comes from Things: 9.5 points there against 0.8 on Common. The other three D3$-$rollout intervals include zero.
\item \textbf{The cross-family collapse replicates.} On Qwen and GLM answers, cross-family D3 identifies 12.9\% and 13.7\% of items, against 33.9\% and 39.7\% within the family. Cross-family channels have 1.9--3.9 times the same-family log loss.
\item \textbf{The family mixture again recovers it.} It is within 1.7 points of the same-family fit (intervals $\pm5$) and 19--27 points above the cross-family fit. On GLM answers, inferring the family beats pooling the answers by 2.8 points ($[0.4,5.3]$, unadjusted).
\item \textbf{Things is harder at every setting}, as expected: its 188 items exceed the 128 leaves of B7.
\end{itemize}

\emph{Why rollout beats D3 for Qwen} (post hoc). Qwen answers almost deterministically: its same-family Brier score is .009--.011. Under the fit, an item with six identical answers to a question gives that answer with probability $13/14$.
\begin{itemize}
\item \textbf{D3's value saturates at the root.} Any three questions whose eight answer cells each hold such an item are worth $8(13/14)^3/N$. Between 8 and 27 of the 32 questions tie at that maximum.
\item \textbf{The tie rule then decides.} The action-order rule picks the bank's first question, which ranks between 1st and 19th by information.
\item \textbf{It is a planning effect, not a transfer effect.} The gap is present under the fit itself: 18.2\% for D3 against 23.7\% for rollout on Things.
\item \textbf{Untested.} Breaking D3's ties by information might remove the gap; we did not test it.
\end{itemize}

\emph{Scope.}
\begin{itemize}
\item Both families were seen in fitting and wrote the banks.
\item Intervals are conditional on these banks and fits, and banks differ widely: Qwen's D3 identifies 61.9\% of Common items with Qwen's bank and 30.6\% with GLM's.
\end{itemize}

\begin{table}[htbp]
\centering\footnotesize
\setlength{\tabcolsep}{4pt}
\caption{Powered panels at B7: held-out contrasts in points, with Bonferroni-8 intervals. \emph{New}: the 598 units of the four new banks. \emph{With earlier}: 820 units, adding the two earlier Common banks. Last column: D3 accuracy (\%) on the new units.}\label{tab:powered-panels}
\begin{tabular}{@{}lrrrrr@{}}
\toprule
Fit $\to$ examiner & D3$-$D2, new & D3$-$rollout, new & D3$-$D2, with earlier & D3$-$rollout, with earlier & D3 (\%) \\
\midrule
Qwen $\to$ Qwen & $+3.7$ $[1.1,6.4]$ & $-6.2$ $[-10.0,-2.5]$ & $+5.2$ $[2.8,7.7]$ & $-6.1$ $[-9.4,-2.8]$ & 33.9 \\
Qwen $\to$ GLM & $+0.4$ $[-1.7,2.6]$ & $+1.4$ $[-1.7,4.5]$ & $+0.8$ $[-1.1,2.7]$ & $+1.5$ $[-1.3,4.3]$ & 13.7 \\
GLM $\to$ Qwen & $+1.7$ $[-1.3,4.6]$ & $+0.2$ $[-3.4,3.7]$ & $+1.9$ $[-0.4,4.3]$ & $+0.9$ $[-2.5,4.1]$ & 12.9 \\
GLM $\to$ GLM & $+4.5$ $[0.8,8.3]$ & $-0.8$ $[-5.8,4.0]$ & $+4.9$ $[1.7,8.3]$ & $-1.0$ $[-5.4,3.0]$ & 39.7 \\
\midrule
Mixture $\to$ Qwen & & & & & 32.2 \\
Mixture $\to$ GLM & & & & & 40.4 \\
\bottomrule
\end{tabular}
\end{table}
\FloatBarrier
\section{Tail risk: testing $\mu>0$}\label{app:risk-planner}
\emph{Problem.} Every result in Section~\ref{sec:main-results} sets $\mu=0$, so the CVaR term has no effect there, although the runtime's default is $\mu=.5$. Those tasks are scored on the mean, so a nonzero weight would optimize something other than the score. Does the risk term do what it claims on a task built for it? \emph{Approach.} Theorems~\ref{thm:monotone} and~\ref{thm:cvar} make dynamic programming valid for the nested mix $M_\mu$, and the recursive objective is then time-consistent: after any answer, the continuation of the root's optimal plan is optimal from that history. The gate's default backup averages continuation values, which is not an episode-level risk objective (Appendix~\ref{app:lean-proofs}, conditional risk). We add an opt-in nested backup and measure realized tails.

\paragraph{Nested risk in the gate.} The default gate, Equation~\eqref{eq:controller-value}, averages continuation values. The conditional-risk example of Appendix~\ref{app:lean-proofs} shows that this is not an episode-level risk objective. The option \code{gate.risk\_backup = "nested"}, available with expectimax only, uses the nested mix instead:
\[
V_d(h,q)=\max\Bigl\{T(h,q),\ \max_g\bigl[X(g,h,q)+M_\mu\bigl(V_{d-1}(h_{g,O})\bigr)\bigr]\Bigr\},\qquad O\sim p_g.
\]
Here $h_{g,o}$ is the history and belief after answer $o$, and the mix $M_\mu$ of Theorem~\ref{thm:cvar} is taken over the answer law $p_g$. This is the model of Theorem~\ref{thm:monotone} with $A=M_\mu$. Tests check the following:
\begin{itemize}
\item at $\mu=0$ the two backups agree exactly at depths one to three;
\item at $\mu\in\{.25,.5,1\}$ the nested value never exceeds the expectation value over 60 belief--depth pairs, and is strictly lower in some;
\item branch-and-bound pruning and the lookahead memo remain exact, because the pruning ceiling still holds when $M_\mu\leq\mathbb E$;
\item invalid backup modes are rejected.
\end{itemize}

\paragraph{A constructed example.} A cause is safe or risky with probabilities $.9$ and $.1$. Acting pays $+1$ if safe and $-1$ if risky, and waiting pays zero. A probe costs $.02$ and reads calm with probability $.9$ if safe and $.5$ if risky. With $\alpha=.8$ and $\mu=.5$, acting now is worth $.4$. After a calm reading, which occurs with probability $.86$, acting is worth $.651$. After an alarm the best action is to wait, worth zero.
\begin{itemize}
\item \textbf{Expectation backup:} the probe is worth $-.02+.86\times.651=.54$, so the gate probes.
\item \textbf{Nested backup:} the mix of those continuation values is $.378$, so the probe is worth $.358$ and the gate acts now.
\end{itemize}
Nested risk penalizes variation in the continuation value itself, so it can value information below the expectation backup. The next paragraph measures it.

\begin{figure}[htbp]
\centering
\includegraphics[width=.8\linewidth]{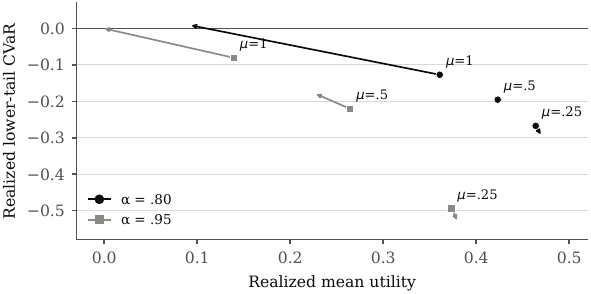}
\caption{Realized utility over 300 models at lookahead depth two. Each arrow runs from the expectation backup to the nested backup at one $(\alpha,\mu)$, with the lower-tail CVaR taken at the planner's $\alpha$; higher is better on both axes. The nested backup raises the tail appreciably only at $\mu=.5$, $\alpha=.95$, and at $\mu=1$, where it gives up the mean.}\label{fig:risk-tails}
\end{figure}

\paragraph{Realized tails.} With a protocol frozen before any result, we ran the runtime planner on 300 random models with a rare bad state (prior .04--.15), a bold act that loses $L\in\{2,4,8\}$ there, a hedge, a wait and two or three costly sensors, one of which sees the bad state. Each episode is fixed by the answer vector, so the realized-utility distribution is exact (Figure~\ref{fig:risk-tails}):
\begin{itemize}
\item At $\mu=.5$ the policies differ in 112--170 of 300 models. At $\alpha=.95$ the nested backup raises the realized lower-tail CVaR by $.024$ ($[.000,.048]$) at depth one and $.040$ ($[.014,.067]$) at depth two, and lowers the mean by $.027$ and $.037$ (Bonferroni-8 over models). At $\alpha=.8$ neither changes detectably.
\item At $\mu=.25$ it makes the tail worse: CVaR at $.9$ falls by about $.10$ and the chance of losing at least $L/2$ rises by a third. The nested mix penalizes the spread that an informative probe creates, so a mildly risk-averse planner probes less and takes the bold act unprotected.
\item At $\mu=1$ it is degenerate: with $\alpha=.95$ it always waits.
\end{itemize}
So the nested backup optimizes its own time-consistent objective, which is not the static CVaR of the episode; it lowered realized tails only at the stricter level, at a cost in mean, and can raise them at low risk weight.

\paragraph{How this reaches the runtime.} The runtime's CVaR routine, its nested aggregator, its action values and the gate's belief-state value under both backups match exact rational values computed in Lean within $10^{-12}$: 305 sets of weighted atoms and 299 root values. The nested value is below the expectation value in 106 of the 299 and never above it. These are finite checks, not a proof of the Python source.
\section{Discovering the latent cause}\label{app:generic-latent}
The fits of Appendix~\ref{app:decision-aware} share the generator's class--context structure. This appendix asks whether the latent-score results survive when that structure must be discovered. The fit, its three selection rules, the contrasts and the expected outcome were frozen with source hashes before any generic fit of the study data.

\paragraph{Fit.} A categorical latent-class model has $K$ components per class and features independent given the component. EM runs from five seeded starts with Dirichlet(2) MAP smoothing. The fit sees the training rows, their labels and the feature cardinalities only; it is not told which feature is the context. $K$ is shared by all classes and chosen in three prespecified ways:
\begin{itemize}
\item \textbf{cv (primary):} $K\in\{1,\dots,6\}$ maximizing 5-fold class-stratified cross-validated log likelihood;
\item \textbf{BIC:} $K\in\{1,\dots,6\}$ minimizing BIC, the rule of Appendix~\ref{app:tier2};
\item \textbf{$K$ = contexts:} the generator's number of contexts, a diagnostic that separates model selection from structure discovery.
\end{itemize}
Every method, including D1--D3, rollout and CMI, plans with the generic fit. All are scored under the same true models and classifiers as Appendix~\ref{app:decision-aware}.

\paragraph{What the fits found.} Recovery is measured by the normalized mutual information (NMI) between each training row's most probable component within its class and its class--context pair.
\begin{itemize}
\item \textbf{cv} chose two components in all 70 noisy-grid fits with 600 rows and two contexts (NMI .999), and six in all five CUBE-NM fits (NMI at least .97).
\item \textbf{BIC} chose one component in every CUBE-NM fit and in 94 of 160 noisy-grid fits, so the latent is the label itself.
\end{itemize}

\paragraph{Results.} Tables~\ref{tab:generic-methods} and~\ref{tab:generic-contrasts} report every method and contrast.
\begin{itemize}
\item \textbf{Noisy grid: nothing is lost.} Rollout-LIG is level with D3 on the generic fit ($-.0035$ [$-.0125$, $+.0055$]) with about a sixteenth of its work, and the generic fit matches the structured one ($+.0013$ [$-.0026$, $+.0052$]). Every value is identical under the context-last catalogue.
\item \textbf{CUBE-NM: LIG still wins, at some cost.} LIG asks for the context first in every episode. It beats CMI by 33.0 points and D3 by 9.0 points, and is 7.0 points below its structured accuracy, an interval that includes zero. The gap is 11.0 points at five queries, an interval that excludes zero.
\item \textbf{BIC: every planner collapses.} With one component per class, LIG equals CMI at every budget. A naive Bayes model cannot represent context-dependent features, so every planner, D3 included, falls to 41.5--47.0\% on CUBE-NM. On the noisy grid, Rollout-LIG and D3 both lose about .008 against the structured fit.
\end{itemize}

\paragraph{Expectations.} Before running we expected the following:
\begin{itemize}
\item BIC would choose one component in many fits;
\item cross-validation would recover CUBE-NM's 40 class--context patterns, and LIG would ask for the context first;
\item generic LIG would match its structured accuracy on CUBE-NM;
\item on the noisy grid, generic Rollout-LIG would fall below the structured fit, most with 120 rows and five contexts.
\end{itemize}
The first two held and the last two did not. On CUBE-NM generic LIG is lower. On the noisy grid there was no shortfall: with 120 rows and two contexts the generic fit did better (.0201 against .0038). The five-context cells are worth zero to every planner under both fits at budget three, so they could show neither a gain nor a loss.

\paragraph{Scope.} Both families are class--context mixtures that a generic categorical mixture can represent; a latent outside the fit's family is not tested. The ten two-context noisy-grid cells carry all the value at these budgets. CUBE-NM has five fits, and one $K$ is shared by all classes. The CUBE-NM gap is consistent with softer context assignments: the generic components' mass-weighted largest context probability is .70, against 1 for the structured fit. The study does not isolate this cause.

\begin{table}[htbp]
\centering\small
\setlength{\tabcolsep}{4pt}
\caption{Planning with a generic latent-class fit that is not told the class--context structure. Noisy grid: mean exact utility over 160 fits, original catalogue. CUBE-NM: own shared-classifier accuracy (\%) at seven queries, five fits; the MAP terminal except the aligned rows. Generic fits choose the number of components per class by 5-fold cross-validated likelihood (cv) or BIC; structured fits are those of Appendix~\ref{app:decision-aware}. CUBE-NM spends a fixed budget, so LIG-stop equals LIG there.}\label{tab:generic-methods}
\begin{tabular}{@{}lrrrrr@{}}
\toprule
 & \multicolumn{2}{c}{Noisy grid} & \multicolumn{3}{c}{CUBE-NM B7} \\
\cmidrule(lr){2-3}\cmidrule(l){4-6}
Method & Generic (cv) & Structured & Generic (cv) & Structured & Generic (BIC) \\
\midrule
CMI & $-$.0743 & $-$.0780 & 56.0 & 56.0 & 43.0 \\
D2 & .0256 & .0280 & 79.5 & 77.5 & 41.5 \\
D3 & .0671 & .0653 & 80.0 & 80.5 & 45.0 \\
Rollout & .0395 & .0379 & 81.5 & 82.5 & 43.0 \\
LIG & $-$.0478 & $-$.0541 & 89.0 & 96.0 & 43.0 \\
LIG-stop & .0452 & .0431 & -- & -- & -- \\
Rollout-LIG & .0635 & .0622 & 85.0 & 86.0 & 42.0 \\
Rollout-LIG (aligned) & -- & -- & 94.0 & 99.5 & 46.0 \\
Aligned D2 & -- & -- & 96.0 & 99.5 & 47.0 \\
\bottomrule
\end{tabular}
\end{table}
\begin{table}[htbp]
\centering\footnotesize
\setlength{\tabcolsep}{4pt}
\caption{Prespecified generic-fit contrasts. Primary: noisy grid Bonferroni-4 $t_9$ over ten replicate blocks; CUBE-NM B7 Bonferroni-3 $t_4$ over five fits, in percentage points. Secondary families use the same adjustments and are descriptive. Context-last noisy contrasts equal the original ones and are omitted.}\label{tab:generic-contrasts}
\begin{tabular}{@{}llll@{}}
\toprule
Family & Study & Contrast & Mean [adjusted interval] \\
\midrule
Primary & Noisy, cv & Rollout-LIG $-$ D3 & $-$.0035 [$-$.0125, +.0055] \\
Primary & Noisy, cv & Rollout-LIG $-$ rollout & +.0240 [+.0151, +.0330] \\
Primary & Noisy, cv & LIG-stop $-$ CMI & +.1195 [+.1117, +.1273] \\
Primary & Noisy, cv & Rollout-LIG, generic $-$ structured & +.0013 [$-$.0026, +.0052] \\
Primary & CUBE-NM B7, cv & LIG $-$ CMI & +33.0 [+25.1, +40.9] \\
Primary & CUBE-NM B7, cv & LIG $-$ D3 & +9.0 [+2.3, +15.7] \\
Primary & CUBE-NM B7, cv & LIG, generic $-$ structured & $-$7.0 [$-$18.0, +4.0] \\
Secondary & Noisy, BIC & Rollout-LIG, generic $-$ structured & $-$.0079 [$-$.0135, $-$.0024] \\
Secondary & Noisy, BIC & D3, generic $-$ structured & $-$.0091 [$-$.0164, $-$.0017] \\
Secondary & Noisy, BIC & Rollout-LIG $-$ D3 & $-$.0020 [$-$.0112, +.0073] \\
Secondary & Noisy, $K$ = contexts & Rollout-LIG, generic $-$ structured & +.0022 [$-$.0015, +.0060] \\
Secondary & CUBE-NM B5, cv & LIG, generic $-$ structured & $-$11.0 [$-$21.7, $-$0.3] \\
Secondary & CUBE-NM B7, cv & Rollout-LIG (aligned), generic $-$ structured aligned D3 & $-$6.0 [$-$10.5, $-$1.5] \\
Secondary & CUBE-NM B7, $K$ = 5 & LIG $-$ D3 & +5.5 [$-$0.3, +11.3] \\
Secondary & CUBE-NM B7, $K$ = 5 & LIG, generic $-$ structured & $-$9.0 [$-$17.0, $-$1.0] \\
Secondary & CUBE-NM B7, BIC & LIG, generic $-$ structured & $-$53.0 [$-$64.0, $-$42.0] \\
\bottomrule
\end{tabular}
\end{table}
 \FloatBarrier
\section{AACO as a non-myopic baseline}\label{app:aaco}
\emph{Problem.} The learned baselines of Appendix~\ref{app:learned-baselines} choose one feature at a time. A fair test of whether depth or a latent score is needed should include a method that plans over subsets. \emph{Approach.} AACO (Acquisition Conditioned Oracle) \citep{valancius2024} makes each decision as follows:
\begin{enumerate}
\item take the $k$ nearest training cases on the observed features;
\item score every candidate subset of unasked features by the mean loss, over those neighbours, of a predictor given the observed features and the subset, plus its cost;
\item stop if the empty subset wins, and otherwise acquire the member of the winning subset with the lowest loss when added alone.
\end{enumerate}
We adapted the authors' code to our tasks, fixing every adaptation before any result:
\begin{itemize}
\item \textbf{Budget.} Candidate subsets respect the query budget, and on hard-budget tasks they fill it.
\item \textbf{Sampling.} Where there are more than 10,000 candidate subsets, 10,000 are sampled, as in the paper's ACO-knn+10k.
\item \textbf{Neighbours.} Rows tied at the $k$-th distance are kept. With nothing observed, every training row is a neighbour, which replaces the authors' hand-picked first feature.
\item \textbf{Loss and predictor.} Signed tasks use the realized utility of the predictor's decision, and accuracy tasks use log loss. The predictor is the planners' fitted model, or the frozen shared classifier on CUBE-NM.
\end{itemize}
A test shows that our decision step reproduces every action of the authors' \code{aaco\_rollout} in their own setting. The protocol was frozen before any AACO result, with $k=5$ primary and $k=20$ secondary.

\emph{Evidence} (Figure~\ref{fig:baselines-budget}a, which plots AACO with the fitted model as predictor; with the shared classifier it reaches 56\%).
\begin{itemize}
\item \textbf{CUBE-NM.} AACO reaches 56--58\% at seven questions, 44 points below aligned D3 ($[37.5,50.5]$) and 38 points below LIG ($[22.5,53.5]$). It never asks for the context first.
\item \textbf{Noisy grid.} AACO trails D3 by $.0180$ ($[.0113,.0247]$) and Rollout-LIG by $.0149$ ($[.0076,.0223]$), with the shortfall in the 600-row two-context cells. It reaches $.0473$ and beats rollout ($+.0094$) and CMI.
\item \textbf{Nine real datasets.} AACO reaches $.5973$, level with D3 ($+.0185$, $[-.0373,.0744]$) and Rollout-LIG ($+.0062$, $[-.0183,.0306]$). It gains most on iris (.623 against .451 for D3), where the latent-class fit is poor, because its neighbour estimates do not rely on the fit.
\end{itemize}

\emph{Why it fails on CUBE-NM.} This diagnostic was not prespecified. Recomputing one root decision shows that the subset search works as intended. The best sampled subset contains the context group, and so do 14\% of the samples. The ordering step then acquires the member with the lowest loss when added alone. The context alone barely changes the class loss (2.04, against 2.07 with nothing observed and 1.95 for the best single digit), so a digit is always asked first. The subset objective is non-myopic, but the rule that picks the next question is myopic on the label.

\emph{Expectations.} Before running, we expected AACO to ask for the context early on CUBE-NM. That failed decisively. We expected it to trail the planners on the noisy grid most at 120 rows; it trailed them, but at 600 rows. We expected it to be within $.02$ of D3 on most real datasets, which held on five of nine.
\section{The full controller on real data}\label{app:controller-real}
\emph{Problem.} The planning results above come from research planners. The controller of Section~\ref{sec:main-controller} adds a belief filter, a selector, a gate and guards, any of which could change decisions. \emph{Approach.} We give the controller the latent-class fits of Appendix~\ref{app:tier2} through a latent-class observation model. Its hypotheses are the fitted components, and each investigation's answer law is that component's emission row. We then run it on every held-out case of the 27 tasks and five splits, with no model calls. The settings are three investigations and a terminal action, payoff $+1/-1/0$, cost $.01$ and $\mu=0$. The controller is deterministic given the answers, so each unique held-out pattern is run once and weighted by its label counts, about 100,000--111,000 decisions per configuration. The protocol was frozen with every runtime source hashed. Before freezing, a test showed that on synthetic latent-class models the controller's utility equals the research planners' exactly for D2, D3, Rollout-LIG and LIG-stop.

The gate supports the methods of Section~\ref{sec:main-methods}:
\begin{itemize}
\item \code{gate.search = "rollout\_lig"} values each answer of a root investigation by the exact LIG-stop continuation;
\item \code{"lig\_stop"} boxes only the investigation most informative about the hypotheses;
\item \code{gate.deepening = "latent"} or \code{"value"} computes the boxes horizon by horizon, under the contract of Theorem~\ref{thm:anytime}.
\end{itemize}
Tests show that LIG-stop's value is never above the expectimax value, and that on an XOR fixture Rollout-LIG asks the two decisive bits where LIG-stop waits. Complete deepening equals plain expectimax, and an interrupted deeper pass keeps the completed shallower decision.

\emph{Evidence.} Table~\ref{tab:controller-real} reports every configuration.
\begin{itemize}
\item \textbf{Conformance.} On all 135 splits the controller's D2, D3, Rollout-LIG and LIG-stop give the research planners' held-out utilities, with largest difference $8\times10^{-15}$.
\item \textbf{Uncapped.} Rollout-LIG minus D2 is $+.0224$ ($[-.0257,.0704]$) and Rollout-LIG minus D3 is $+.0124$ ($[-.0344,.0591]$), Bonferroni-3 over nine datasets.
\item \textbf{Capped at 128 units.} Bounded D3 scores zero on wine and zoo, $-.008$ on scikit-learn diabetes and $.519$ on breast cancer, where the uncapped D2 scores $.774$, $.689$, $.486$ and $.793$. With its root search interrupted it may complete no box, and then it waits. Latent deepening matches the uncapped D2 on every dataset. Its gain over bounded D3, $+.2474$ ($[-.0786,.5734]$), has an interval that includes zero because the gain is concentrated in four datasets.
\item \textbf{Capped at 512 units.} The deepening contrast is $+.0014$ ($[-.0025,.0053]$).
\end{itemize}

\emph{Expectations.} As stated before running, the controller reproduced the research planners and the uncapped contrasts. The capped loss of bounded D3 was larger than expected: often no box completed at all, rather than a mixture of completed boxes. \emph{Scope.} The tasks come from nine local datasets. The model is a conditionally independent latent-class model, so there are no LLM channels, verifier or memory, and only $\mu=0$ is tested. Work is counted in logical units.

\begin{table}[htbp]
\centering\small
\caption{The full controller on nine real datasets. Utility is held-out, averaged over datasets. Work is planning units per decision. Interrupted counts decisions whose search hit the cap.}\label{tab:controller-real}
\begin{tabular}{@{}llrrr@{}}
\toprule
Gate & Cap & Utility & Work & Interrupted \\
\midrule
D2 (default depth) & none & .5688 & 36 & 0 \\
D3 & none & .5788 & 197 & 0 \\
Rollout-LIG & none & .5912 & 74 & 0 \\
LIG-stop & none & .5604 & 10 & 0 \\
D3, bounded & 128 & .3219 & 96 & 50,689 \\
D3, latent deepening & 128 & .5693 & 85 & 50,776 \\
D3, value deepening & 128 & .5690 & 85 & 50,776 \\
Rollout-LIG & 128 & .5912 & 74 & 603 \\
D3, bounded & 512 & .5782 & 171 & 3,239 \\
D3, latent deepening & 512 & .5796 & 194 & 8,333 \\
Rollout-LIG & 512 & .5912 & 74 & 0 \\
\bottomrule
\end{tabular}
\end{table}
\section{Are probabilities read from a language model real?}\label{app:elicited}
\emph{Problem.} In a new situation with no fitted model, the controller reads its likelihoods from the token log probabilities of language models (Appendix~\ref{sec:controller}). A model does not produce calibrated probabilities, and nothing guarantees that these numbers are true. Can they be measured, and what can be built on them?

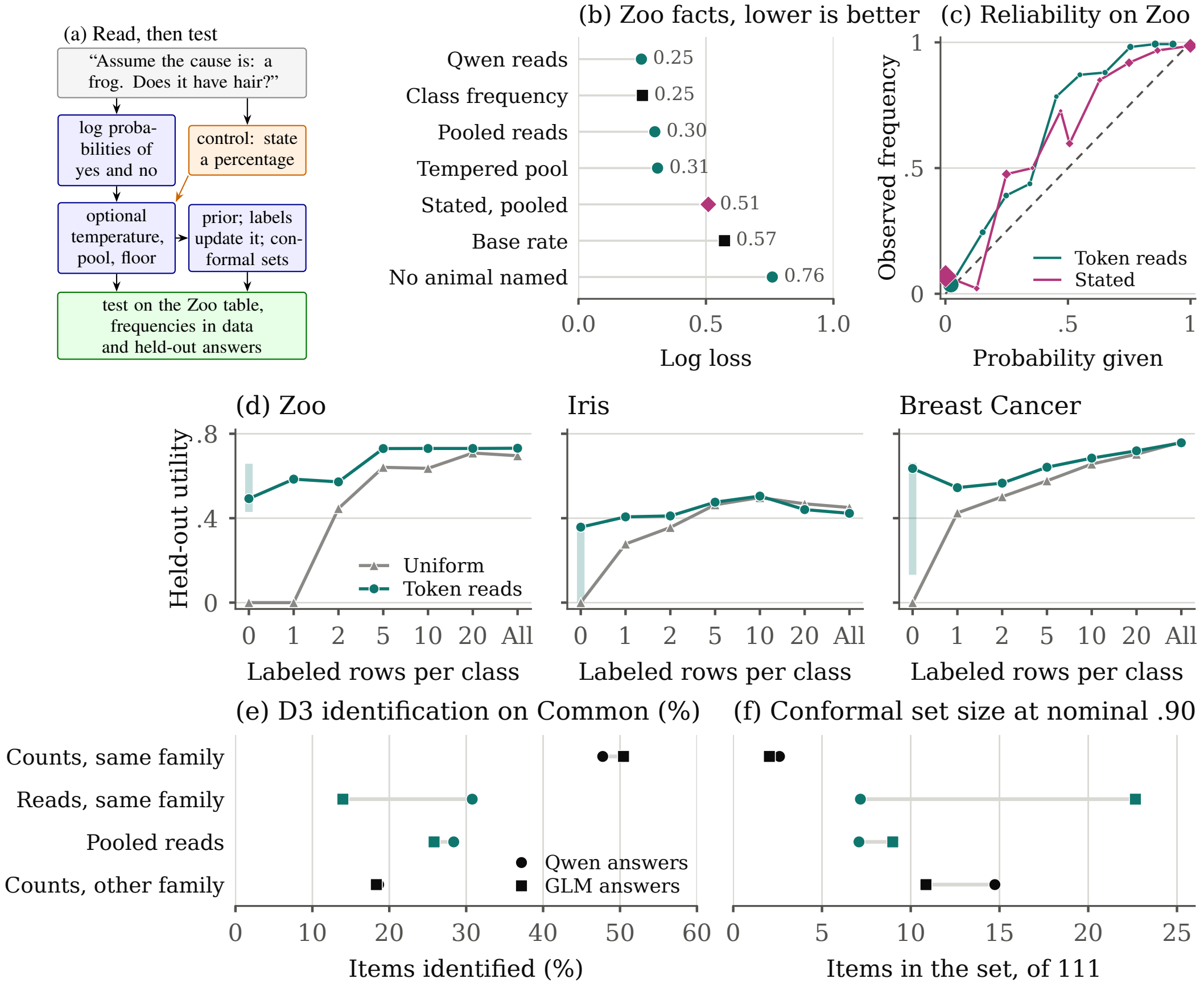
\begin{figure}[htbp]
\centering
\begin{minipage}[c]{0.3\linewidth}
\centering
\begin{tikzpicture}[font=\tiny, >={Stealth[length=1.2mm]},
  box/.style={draw=black!45, rounded corners=1pt, fill=gray!8, align=center, inner sep=2pt},
  fit/.style={draw=blue!55!black, rounded corners=1pt, fill=blue!7, align=center, inner sep=2pt},
  test/.style={draw=green!45!black, rounded corners=1pt, fill=green!10, align=center, inner sep=2pt},
  ctrl/.style={draw=orange!80!black, rounded corners=1pt, fill=orange!12, align=center, inner sep=2pt}]
\node[font=\scriptsize, anchor=west] at (-1.5,2.12) {(a) Read, then test};
\node[box, text width=2.75cm] (q) at (0,1.68) {``Assume the cause is: a frog. Does it have hair?''};
\node[fit, text width=1.22cm] (r) at (-.76,.78) {log probabilities of yes and no};
\node[ctrl, text width=1.22cm] (s) at (.76,.78) {control: state a percentage};
\node[fit, text width=1.22cm] (p) at (-.76,-.24) {optional temperature, pool, floor};
\node[fit, text width=1.22cm] (c) at (.76,-.24) {prior; labels update it; conformal sets};
\node[test, text width=2.75cm] (t) at (0,-1.26) {test on the Zoo table, frequencies in data and held-out answers};
\draw[->] (q.south-|r) -- (r); \draw[->] (q.south-|s) -- (s);
\draw[->] (r) -- (p); \draw[->] (p) -- (c);
\draw[->, draw=orange!80!black] (s.south west) -- (p.north east);
\draw[->] (p.south) -- (p.south|-t.north); \draw[->] (c.south) -- (c.south|-t.north);
\end{tikzpicture}
\end{minipage}\hfill
\begin{minipage}[c]{0.68\linewidth}
\centering
\includegraphics[width=\linewidth]{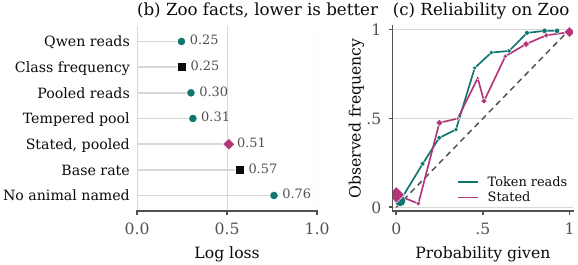}
\end{minipage}

\vspace{4pt}
\includegraphics[width=\linewidth]{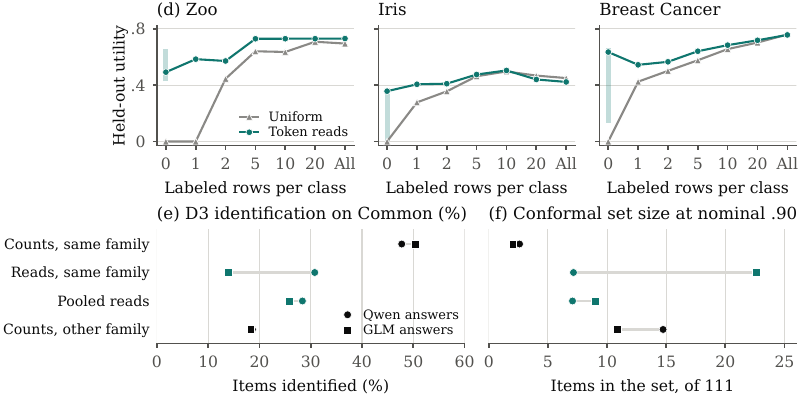}
\caption{Probabilities read from four language models, against targets outside them. (a) The route; the prompt is paraphrased. (b) Fifteen attributes of 101 animals; base rate and class frequency are fitted leaving each animal out. (c) Pooled reads and pooled stated percentages, with perfect calibration dashed and marker area proportional to the requests in a bin. Token reads lie above the diagonal at every level. The stated lowest bin gives almost 0 and comes true 7\% of the time. (d) D3 at budget three, over the real-data study's splits. The pale bar at zero rows spans the four single models. (e) Seven questions. Counts use six fitting requests per item; reads use none. The pool includes both answering families. (f) Mean coverage is .88--.91 in every cell of this panel.}\label{fig:elicited}
\end{figure}

\emph{Solution.} Use a read as a prior, not an estimate: pool four models, let labeled counts update the pool (Equation~\eqref{eq:controller-channel}), and once labels exist report conformal sets, whose coverage needs labeled exchangeable units but not calibrated numbers. Every read here goes through the runtime's own prompt, client, softmax, floor, pool, temperature and concentration code (Figure~\ref{fig:elicited}a). We read 10,946 prompts from each of Qwen, GLM, Nemotron and Gemma:
\begin{itemize}
\item 15 attributes of 101 named animals, against the UCI Zoo table;
\item every class and feature of Zoo, Iris and Breast Cancer, against frequencies in held-out rows;
\item every item and question of two Common banks, against the held-out answers of Qwen, GLM and Nemotron;
\item the Zoo and class prompts again, asking for a percentage instead, as a comparison with token reads.
\end{itemize}
A control names no animal. A frozen protocol fixed eight primary contrasts (Bonferroni-8, 10,000 paired bootstrap draws) and ten expectations before any study read; a pilot on four other prompts had checked only format and speed, and had already shown Gemma's near-hard reads. Tabular results are descriptive, because three datasets are too few for inference. With all rows, the uniform arm reproduces the earlier real-data study exactly on the ten splits whose fit had one latent class per class (Zoo and Iris), and the counted arm reproduces the earlier Common study exactly.

\emph{Lean.} No theorem covers the reads. Theorems~\ref{thm:transfer} and~\ref{thm:robust} limit regret by the model error whatever its source, but their bounds are not evaluated here because the true conditionals are unknown. Split conformal's coverage is a classical result and is not in our proofs.

\emph{Simulation} (Figure~\ref{fig:elicited}).

\paragraph{Known facts.} On Zoo the pooled read has accuracy .887, AUC .962 and log loss .299. It beats the base rate (.573; difference $-.273$, $[-.318,-.225]$), and naming the animal is what carries the information: with no animal named, the pool scores .760. It does not beat class frequencies (.251; difference $+.049$, $[-.007,.103]$), though Qwen alone matches them (.247). So we have no evidence that these models know more about a named animal than its class gives.

\paragraph{Stated percentages.} Stated percentages are about as accurate and cost far more when wrong: pooled accuracy is .879, but log loss rises by $.210$ ($[.111,.333]$). A single model's log loss rises from .247--.693 to .804--1.872. Models say exactly 0 or 100 on 89--99\% of Zoo requests (77--97\% of all stated requests), and their reads at or beyond .1 and .9 are wrong 10--21\% of the time. ECE does not show this: it is .089 for the stated pool and .095 for the token pool, and mixed across single models, so the supported claim is lower log loss and fewer confident errors, not better calibration in general. The token reads of Gemma are nearly as hard, with 94.7\% outside $[.01,.99]$, so token log probabilities are not automatically probabilities.

\paragraph{Temperatures.} A fitted temperature lowers single models' log loss on Zoo: Gemma's falls from .693 to .261 at a temperature near 6.5, Nemotron's from .577 to .429 and GLM's from .621 to .529. The accept rule refuses Qwen's fit. It does not help the pool ($+.011$, $[-.006,.026]$), which understates yes at every level: pooled reads near .25 come true 39\% of the time, near .45 78\% and near .75 98\%. A temperature scales around one half, so it cannot correct a one-sided bias; correcting after the pool was not tested. On Common, where temperatures were fitted on the other bank's answers, the accept rule refused every fit from one bank, and the accepted ones raised log loss on the model's own answers (GLM .373 to .398, Nemotron .428 to .450, averaged over both banks, one of them untempered). On Iris the pool improves slightly (.542 to .527), Qwen needs a temperature of .22 and Nemotron reaches the limit of 20, so overconfidence is not universal.

\paragraph{Frequencies in data.} On Zoo the pool's absolute error (.105) beats counts on all training rows (.127). On Iris it equals counts on one row per class (.241 against .238). On Breast Cancer it lies between them (.099, against .043 and .175).

\paragraph{Plans from few labels} (Figure~\ref{fig:elicited}d). With no labels a uniform fit abstains, while the pool earns .49, .36 and .64 on Zoo, Iris and Breast Cancer. That is 67--85\% of its utility with all rows, and worth about 2--5, 2 and 5--10 labeled rows per class. Single models vary widely (.43--.66, 0--.38, .13--.66), so the pool avoids the worst without reaching the best. With all rows the gap does not always close: the pool stays above uniform on Zoo by .036, and below it on Iris by .028, although the concentration on Iris is at most 5.1 against 35 rows per class; we have not established why. On Breast Cancer one labeled row per class lowers the pool from .635 to .545. Stated percentages plan as well as reads on Zoo and Iris, but not on Breast Cancer (.468), so a higher log loss does not always reach utility.

\paragraph{Plans on a model's answers} (Figure~\ref{fig:elicited}e). Within a family, counts win. Reads have higher log loss (Qwen $+.063$, $[.048,.079]$; GLM $+.238$, $[.220,.255]$) and lower D3 identification (Qwen $-17.0$ points, $[-27.3,-6.6]$; GLM $-36.5$, $[-45.8,-27.2]$), and a single repeated answer per item already beats them (38.0 against 30.8 for Qwen, 35.9 against 14.0 for GLM). Across families the pool leads, but it contains the answering family's own reads: with no answers counted, it reaches 28.4 and 25.8, against 18.6 and 18.3 from the other family's six requests, with lower log loss on all three examiners. Single reads from another family are mixed: 15.6, 15.6 and 29.9 on Qwen's answers, 25.2, 17.9 and 20.7 on GLM's. The pool does not reach the family mixture (44.1 and 46.6, Figure~\ref{fig:answering}b). A prior of pooled reads plus six requests gives up .8 (Qwen) and 6.0 points (GLM) within a family and gains 7--8 across, while the third model stays near 10\% on every table tested here. Depth adds nothing on reads: D3 $-$ D2 lies within 2.6 points. These comparisons beyond the primary contrasts are post hoc.

\paragraph{Coverage} (Figure~\ref{fig:elicited}f). Mean conformal coverage at nominal .90 is .898--.917 on the three class tasks and .881--.904 for Qwen's, GLM's and the pooled reads on Common, while mean set size moves with read quality. Single splits vary (pooled reads on Iris .822--.978). Of 111 Common items, sets hold 2.0--2.6 with same-family counts, 7.1--9.0 with pooled reads and up to 22.7 with GLM's reads. Coverage does not depend on the numbers being true, and poor numbers pay with larger sets, but it needs labeled units, so a new scenario has no set until labels exist.

\paragraph{Frozen expectations.} Five held: both tokens are returned (100\%) and Gemma reads near 0 or 1; Zoo accuracy, AUC and log loss clear their marks; pooling beats the median single model; stated percentages have higher log loss than token reads; coverage is at least .85. Five did not hold in full:
\begin{itemize}
\item temperatures lower log loss for single models on Zoo, but not for the pool there or on Common, and class results are mixed;
\item tempered pooled reads do not beat class frequency (.310 against .251);
\item class error lies between all rows and one row per class only on Breast Cancer;
\item within a family, D3 falls by 17 and 36 points, not at most 10 (Qwen's interval includes 10), and GLM's reads fall below the cross-family count fit (14.0 against 18.3);
\item with all rows the utility gap closes to within .02 only on Breast Cancer.
\end{itemize}

\paragraph{Limits.} These are fixed yes/no questions on tasks the models may have seen in training, not new scenarios, and truth tables contain judgment calls. On Common the target is a model's answer, not a fact, and reads and answers use different prompts, so reads $-$ counts also contrasts prompts. Temperature 0 does not fix GLM's or Gemma's log probabilities: 33\% and 43\% of repeated values changed, by at most .26 and .75, while Qwen's and Nemotron's did not change. The Gemma build is \code{gemma4:31b-mlx}, the one installed, not the build the runtime's configuration names.

\emph{Verdict.} The numbers can be measured, not guaranteed, and they should not be treated as estimates. Confirmed by the primary contrasts: token reads beat the base rate on known facts, have lower log loss than stated percentages, and lose to counts within a family, for GLM by more than ten points. Not shown: that they beat class frequencies, or that tempering before pooling helps the pool. Descriptive: a read is worth a few labeled rows, and once labels exist mean conformal coverage stays near nominal however poor the reads are.
\section{Choosing the depth from witnesses}\label{app:order-rule}
\emph{Problem.} Rollout-LIG recovers depth on the noisy grid and CUBE-NM but not on context parity, where full-horizon search is needed. The deepen-when-needed rule of Appendix~\ref{app:tier2} screens every history and saved no work. Can one planner be picked per fit instead? \emph{Approach.} Once per fit, at the root and under the fitted model, the rule computes two exact quantities:
\begin{itemize}
\item for each $k$ from 2 to the budget, a depth-$k$ witness $W_k$: the best first query, then after each answer the better of stopping and the best fixed set of $k-1$ more features;
\item Rollout-LIG's own exact root value $L$.
\end{itemize}
If $\max_k W_k>L+\epsilon$, with $\epsilon=.002$, the rule plans with exact $D_k$ for the smallest $k$ whose witness is within $\epsilon$ of the maximum; otherwise it plans with Rollout-LIG. Under the fit, with a stop option and no interruption, a rule that deepens satisfies $V(\mathrm{executed}\ D_k)\geq V_k(\mathrm{root})\geq W_k>L+\epsilon$: Theorem~\ref{thm:rollout} gives the first inequality, and the second holds because $W_k$ is one depth-$k$ policy. A kept Rollout-LIG is worth at least $L$. Since $L$ is a root value, this does not promise to beat executed Rollout-LIG. The protocol was frozen before any result. In all 375 fits the chosen policy reproduces the frozen comparator's utility to $10^{-12}$.

\emph{What a fixed subset shows.} We also report the fixed-subset witness order: the smallest $k$ such that asking some fixed set of $k$ features and then predicting beats stopping by more than $\epsilon$. Such a set shows that depth $k$ beats stopping (Theorem~\ref{thm:interaction-sufficiency}). Its absence shows nothing: Theorem~\ref{thm:interaction-order} needs the label's posterior to equal its prior after every shorter history, a stronger condition. With a fair context, two blocks of two fair bits, no noise, budget three and cost $.30$, the best fixed sets are worth $-.30$, $-.10$ and $-.40$, while the context followed by the selected block is worth $+.10$. Nor does the order fix the depth: on context parity one block of bits already predicts the label 75\% of the time, but only the context followed by the selected block identifies it.

\emph{Evidence} (Table~\ref{tab:order-rule}). On the synthetic suites the rule gets the better of the two planners: it equals full-horizon depth on every parity fit (41 points above Rollout-LIG), and on the noisy grid its mean is $.0003$ below D3. It deepened on 32 of 160 noisy fits, all 80 parity fits, and 58 of 135 real fits, including every MiniBooNE fit and 11 of 20 on digits. Held out on real data it trails Rollout-LIG by $.014$ (wide interval) because it deepens on iris, where D3 is misled by a poor fit; under the fit it is ahead, $+.0062$ ($[.0003,.0121]$, unadjusted).

\emph{Where it fails.}
\begin{itemize}
\item \textbf{Work.} Witnesses scan pairs of features after every answer to every first query: about 1.9 million transitions in the 51-feature noisy cells. The rule costs 1.24 times D3's work on the noisy grid, and more than D3 on parity and the real datasets, so on these tasks running the deeper search directly is cheaper than screening for it.
\item \textbf{Detection.} A witness is a lower bound. D3's fitted value still exceeds the rule's by more than $\epsilon$ on 9 noisy fits (at most $.035$) and 9 real fits (at most $.017$), so a failed witness does not show that depth is worthless.
\end{itemize}
The theorems say which depths can have value and that a latent score is never blind; neither decides which planner to use.

\begin{table}[htbp]
\centering\footnotesize
\setlength{\tabcolsep}{4pt}
\caption{Depth from witnesses. Mean exact utility (real data: held out) and compilation transitions. Contrasts: Bonferroni-6; noisy grid and parity over ten replicate blocks, real data over nine datasets.}\label{tab:order-rule}
\begin{tabular}{@{}lrrrrr@{}}
\toprule
Suite & Rule & Rollout-LIG & D3 / DB & Rule $-$ Rollout-LIG & Rule $-$ D3 / DB \\
\midrule
Noisy grid & .0650 & .0622 & .0653 & $+.0028$ $[-.0011,.0067]$ & $-.0003$ $[-.0025,.0020]$ \\
Context parity & .7838 & .3751 & .7838 & $+.4088$ $[.3763,.4412]$ & .0000 (all fits) \\
Nine datasets & .5773 & .5912 & .5788 & $-.0139$ $[-.0666,.0388]$ & $-.0015$ $[-.0096,.0065]$ \\
\midrule
Work, noisy grid & 915{,}264 & 61{,}110 & 738{,}271 & & \\
Work, parity & 4{,}218 & 1{,}070 & 1{,}796 & & \\
Work, nine datasets & 5{,}317 & 1{,}386 & 2{,}915 & & \\
\bottomrule
\end{tabular}
\end{table}

\section{Pessimistic choice against validation selection}\label{app:pessimistic}
\emph{Problem.} Choosing the candidate with the highest fitted value minus its on-policy bound $B$ guarantees that score as a lower bound and regret at most twice any rival's bound (Theorem~\ref{thm:pessimistic}). Does it choose better than the fitted value alone? \emph{Approach.} Four rules choose among candidate policies: greedy (highest fitted value), pessimistic (highest fitted value minus the bound of Theorem~\ref{thm:robust}), validation selection (highest value when the candidates are replayed on the validation rows the radii come from, reporting a Hoeffding lower bound), and the oracle. We apply them to the 216 cells of the controlled model-error study, with D1--D3, rollout and CMI as candidates, fits from 128--8{,}192 rows of the matched source or of a source whose label is independent of the features, and $m=n$ or $8n$ validation rows; and to the Common panels, with 20 candidates (Qwen, GLM, pooled and family-mixture fits, each with five planners), examiner repeat 0 as validation and repeats 1--2 as truth. The protocol was frozen before any rule was applied; every bound reproduces the earlier study to $10^{-12}$.

\begin{figure}[htbp]
\centering
\includegraphics[width=.92\linewidth]{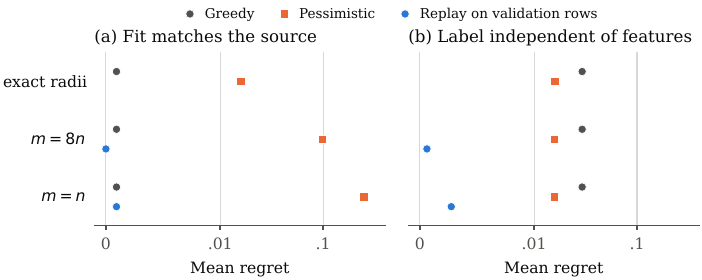}
\caption{Mean regret against the best candidate, on a scale linear below $.01$ and logarithmic above, by source and error radii, pooled over sizes and costs (108 cells per setting). Replay needs validation rows, so it has no exact-radii value.}\label{fig:pessimism}
\end{figure}

\emph{Evidence} (Figure~\ref{fig:pessimism}). Every reported lower bound held. On the matched source, greedy already picks the best policy in 107 of 108 cells; pessimism with validation radii loses $.25$ ($[.15,.34]$, Bonferroni-8) at $m=n$ and $.10$ ($[.04,.16]$) at $m=8n$, because the bound grows with the policy's tree and pushes the choice back to D1. Under transfer to the independent-label source pessimism lowers regret from $.029$ to $.016$, at every kind of radius: each unadjusted interval excludes zero (about $[-.025,-.002]$) and each Bonferroni-8 interval includes it (about $[-.030,.003]$). On the panels greedy twice picks a policy fitted to the other family (14.0\% and 16.2\% where the best is 54.1\% and 39.2\%); pessimism with exact radii avoids both (54.1\% and 30.6\%), but with radii from 111 validation rows the bound is 6.0--7.2 for every candidate on a 0--1 scale and decides nothing. Validation selection is never worse than either rule: regret at most $.003$, the best policy in 98--108 of 108 cells, lower bounds three to five times tighter, and on the panels within 2.8 points of the oracle.

\emph{Reading.} The trade-off that Theorem~\ref{thm:pessimistic} allows appears in practice: pessimism helps where the fit is wrong and costs where it is right. When the radii come from complete rows of the true distribution, replaying the candidates on those rows is the better selection rule. The bound's role is narrower: it needs only one-step statistics and covers every state of a policy, including states the validation data never reach.

\section{Scaling in features and budget}\label{app:scaling}
\emph{Problem.} The exact planners run on at most 51 feature groups with budgets up to seven, so it is unclear how the methods scale. \emph{Approach.} We generalize the authors' noiseless CUBE-NM constructor to $C\in\{2,6,12,25\}$ contexts: $8C$ equally likely class--context atoms and $d=1+10C$ features (21, 61, 121 and 251), one categorical context and $10C$ three-valued bits, with hard budgets of 3, 7 and 15. Every planner plans with the exact model, so the study measures search rather than fitting, and predicts its MAP class; accuracy is exact over the atoms. LIG, Rollout-LIG and CMI run unbounded; LIG-ordered iterative deepening to D3 and UCT with exact rollout leaves ($c=.3$, no stop before the budget) get a fixed allowance per decision (4{,}096 and 65{,}536 transitions, plus 1{,}048{,}576 at 21 and 61 features). DIME is the authors' code with sizes adapted to $C$, trained on 20 rows per atom with three seeds; its acquired features are scored by the same exact MAP class. Work is the transitions spent along an episode's decisions, each from cleared caches. The protocol was frozen before any result.

\begin{figure}[htbp]
\centering
\includegraphics[width=\linewidth]{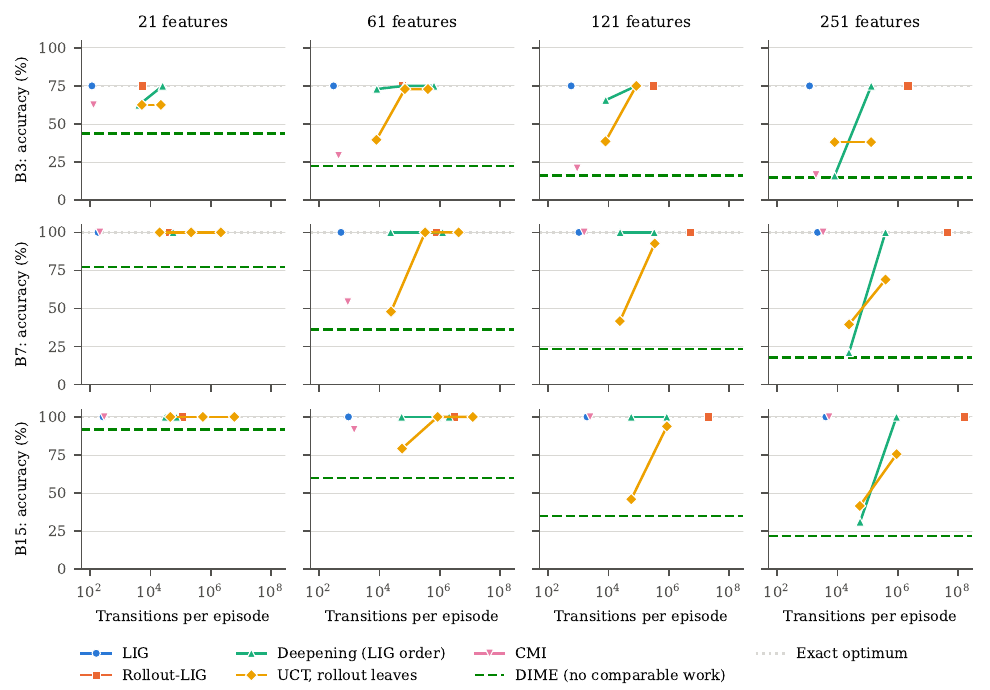}
\caption{Accuracy against transitions per episode on CUBE-NM. Rows: budgets; columns: feature counts. Lines join a method's allowances; DIME (dashed) has no comparable work; dotted grey is the exact optimum.}\label{fig:scaling}
\end{figure}

\emph{Evidence} (Figure~\ref{fig:scaling}). The exact hard-budget optimum is 75\% at B3 and 100\% at B7 and B15 at every size.
\begin{itemize}
\item \textbf{LIG} reaches the optimum in every cell with at most 4{,}100 transitions per episode; its work grows as about $d^{1.0}$ (log--log slopes .94--1.10).
\item \textbf{Rollout-LIG} also reaches it, but its forced LIG continuation runs to the end of the budget for every candidate, so its work grows steeply with $d$ and $B$: from 5{,}600 transitions per episode at 21 features and B3 to 166 million at 251 features and B15, about 40{,}000 times LIG's, growing as about $d^{2.4}$ to $d^{2.9}$. Where LIG is already optimal, rollout adds only cost.
\item \textbf{LIG-ordered deepening} stays at the optimum at 65{,}536 transitions per decision at every size; at 4{,}096 it reaches the optimum at B7 and B15 up to 121 features but collapses at 251 (16--31\%), where even depth two no longer completes.
\item \textbf{UCT} degrades with size at a fixed allowance: at 251 features it reaches 38--76\% at 65{,}536. With 1{,}048{,}576 at 21 and 61 features it reaches the optimum except at B3 (62.5\% and 72.9\%), where deepening does.
\item \textbf{CMI and DIME.} CMI falls short at B3 at every size and at 61 features with B7 and B15, but reaches the optimum at B7 and B15 with 121 and 251 features. DIME never asks for the context first and falls from 44--92\% at 21 features to 15--22\% at 251.
\end{itemize}

\emph{Expectations.} LIG and Rollout-LIG at the optimum for $B\geq7$ at every size held, as did UCT's decline. Rollout-LIG's work grew faster than the expected $d^2$, deepening at 65{,}536 did not lose accuracy as expected, and CMI was not uniformly below LIG.

\section{Utility as a share of the attainable gap}\label{app:gap-fraction}
\begin{figure}[htbp]
\centering
\includegraphics[width=\linewidth]{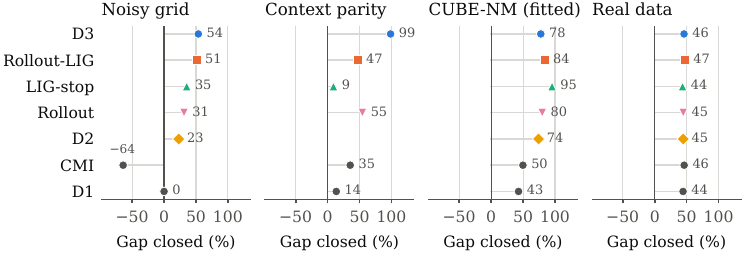}
\caption{Share of the gap from acting now to the oracle (real data: to perfect information). Floors: noisy grid and parity 0, CUBE-NM 12.5\%, real data .222 held out. Oracles: .122, .795, 100\%. D3 is full-horizon $D_B$ on parity, where D2 is not a comparator; LIG-stop is LIG on CUBE-NM, which spends a fixed budget.}\label{fig:gap-share}
\end{figure}

Raw utilities on the noisy grid are small (.028--.065), so differences of .003--.03 are hard to judge. Figure~\ref{fig:gap-share} rescales them: the floor is the true value of acting without a query; the oracle is the best policy computed with the true model at the same budget and cost (exact D3 on the noisy grid, whose budgets are at most three in 15 of 16 cells; full-horizon $D_B$ on parity; the exact hard-budget optimum on CUBE-NM). On real data, where no true model exists, the top is perfect information, which reveals the label for free. Shares are computed from equal-cell (real data: dataset-level) means. No policy was rerun.

The best policy under the true noisy-grid model is worth only $.122$, so D3 closes 54\% of the attainable gap, Rollout-LIG 51\%, rollout 31\% and D2 23\%: even exact search under the fitted model leaves 46\% to model error. On CUBE-NM at B7, LIG closes 95\%, D3 planned for its fitted terminal 78\%, and aligned D3, planned for the scored classifier, all of it. On the real datasets every planner closes 44--47\% of the perfect-information gap, so their differences stay small in these units.

\section{Representation refinement}\label{app:representation}
\emph{Problem.} The latent-visibility theorem says a one-step latent score is never blind, not that it ranks well. One observable problem has many latent descriptions. Does LIG's success depend on the problem or on the description?

\emph{Solution tested.} Two remedies. Where the decision structure is known, keep it in the latent and drop the rest. Otherwise, check LIG against exact values, which no description can change: the witness rule of Appendix~\ref{app:order-rule}, or a cheap \emph{portfolio} that plans with D2 when its root value exceeds Rollout-LIG's by more than $.002$. Under the fit, with a stop option and no interruption, the portfolio's executed value is at least Rollout-LIG's root value and within $.002$ of D2's.

\emph{Lean test.} Moving a question's answer noise into the cause leaves every exact value unchanged, for any finite latent model, and can only add visible questions (Theorem~\ref{thm:refinement}). In a two-bit family a LIG rule earns $1-2c$ under the coarse description and $0$ under the refined one, however its ties break, while D2 earns $1-2c$ under both (Theorem~\ref{thm:gap}).

\emph{Simulation test.} The protocol, contrasts and eight expectations were frozen before any result of a refined description. Python tests check the Lean family's scores and values and that refinement preserves the joint over answers and label. The coarse descriptions reproduce the earlier frozen results to $10^{-10}$ on 2,705 rows. Primary contrasts use Bonferroni-8.
\begin{itemize}
\item \textbf{Exact family.} Nuisance alphabets of 2, 3, 4 and 8 answers, budgets 2 and 3, four costs, with no fitting.
\item \textbf{Context parity.} The true joints with the saturated latent of Section~\ref{sec:main-datasets} and the structured latent (context, selected block, label), which describes the same joint, under both catalogues.
\item \textbf{Fitted suites.} The 160 noisy-grid and 135 real-data fits, with planners scored under the true model or on held-out rows. \emph{Coarse} adds a question with eight equally likely answers that the component ignores. \emph{Refined} has the component record it. \emph{Noise 1} and \emph{noise 3} add no question: the component records the answers it leaves most uncertain.
\item \textbf{Learned latents.} The noisy-grid training sets with three nuisance questions that share a hidden four-valued state and reveal it with chance $.9$, so a latent satisfying conditional independence must record it. The generic fit of Appendix~\ref{app:generic-latent} chooses up to six components per class by cross-validation.
\end{itemize}

\emph{Evidence} (Table~\ref{tab:representation}, Figure~\ref{fig:representation}).
\begin{itemize}
\item \textbf{Exact values do not move.} Between coarse and refined, root values and witnesses change by at most $2\times10^{-15}$ on all 295 fits, and the utilities of D2 and D3 change on none.
\item \textbf{LIG does.} In the exact family the score ties at two answers and the bit, having the lower index, goes first, so nothing fails. With three or more answers at budget two, LIG-stop and Rollout-LIG earn $0$ against $1-2c$. Budget three lets replanning recover: Rollout-LIG earns $1-2c$ and LIG-stop $1-3c$, both $0$ at $c=.45$. On the noisy grid refinement costs LIG-stop $.0362$ ($[.0288,.0435]$) and Rollout-LIG $.0363$ ($[.0297,.0429]$), 58\% of its utility. On real data the losses are $.0478$ and $.0146$, with intervals that include zero.
\item \textbf{No new question is needed.} Noise 1 costs Rollout-LIG $.0414$ on the noisy grid ($[.0334,.0493]$, unadjusted). Under noise 3, LIG-stop and Rollout-LIG stop at the root on all 160 fits and earn exactly 0. On real data noise 3 costs LIG-stop $.1924$ ($[.0105,.3743]$) and Rollout-LIG only $.0066$ (interval includes zero).
\item \textbf{Simplifying the latent removes the need for depth on parity.} Under the structured latent, LIG-stop and Rollout-LIG equal full-horizon search in all eight settings, $.68$--$.91$ against $.31$--$.44$ under the saturated latent. LIG-stop needs 62 transitions against 330 with two-bit blocks, and 170 against 3,262 with three.
\item \textbf{The witness rule recovers the loss, at a cost.} It is unaffected by refinement ($+.0002$, $[-.0008,.0012]$), beats refined Rollout-LIG by $.0393$ ($[.0324,.0462]$), and reaches full horizon on parity under both latents. Its work is 1.58 times D3's on the noisy grid.
\item \textbf{The portfolio recovers part of it, cheaply.} It beats refined Rollout-LIG by $.0138$ ($[.0086,.0190]$) at 14\% of D3's work, but stays $.0225$ below coarse Rollout-LIG, and on parity it does not help, because D2 is not deep enough. On real data neither check differs from refined Rollout-LIG (intervals include zero).
\item \textbf{A cross-validated fit did not fall into this.} With the nuisance block, losses are at most $.0022$, with intervals that include zero, and no planner's first question is in the block on any of 160 fits. Post hoc, on the 100 two-context fits, widening the range to ten components selects eight or ten on 18 fits, and LIG's first question is in the block on one. Fixing eight components hurts every planner (LIG-stop $-.069$, Rollout-LIG $-.081$, D2 $-.040$, witness $-.032$), so that is an oversized fit rather than a refinement effect alone.
\end{itemize}

\emph{Expectations.} Five held. One held in part: the witness rule stays within $.01$ of coarse Rollout-LIG, the portfolio does not. Two failed. We expected recorded noise to cost less than an added nuisance, and it costs as much or more. We expected the learned fit to record the block and LIG to ask it first, and neither happened.

\emph{Verdict.} Confirmed: the description, not only the score, decides LIG's outcome, and exact values do not depend on it. The witness rule's recovery is confirmed, and the portfolio's is partial. Rejected: that a cross-validated fit falls into this by itself, at least with this block. The risk lies in hand-built or oversized latents.

\emph{Scope.} The fitted suites refine by construction; only the learned suite lets a fit choose, with one kind of block on one family of tasks. Parity's structured latent uses the true joint, so this does not test learning that structure from data. The checks' guarantees assume a stop option and no interruption.

\begin{table}[htbp]
\centering\footnotesize
\setlength{\tabcolsep}{3pt}
\caption{Refinement on the fitted suites: mean exact utility (true model; real data: held out), with mean transitions under the refined description. D3 and the witness rule were run only on the coarse and refined descriptions.}\label{tab:representation}
\begin{tabular}{@{}lrrrrrrrrrr@{}}
\toprule
& \multicolumn{5}{c}{Noisy grid} & \multicolumn{5}{c}{Nine datasets} \\
\cmidrule(lr){2-6}\cmidrule(l){7-11}
Method & Coarse & Refined & Noise 1 & Noise 3 & Work & Coarse & Refined & Noise 1 & Noise 3 & Work \\
\midrule
LIG-stop & .0431 & .0069 & .0089 & .0000 & 3{,}947 & .5604 & .5125 & .5273 & .3680 & 491 \\
Rollout-LIG & .0622 & .0260 & .0209 & .0000 & 109{,}302 & .5912 & .5765 & .5871 & .5846 & 4{,}594 \\
D2 & .0280 & .0280 & .0280 & .0280 & 17{,}471 & .5688 & .5688 & .5688 & .5688 & 1{,}483 \\
D3 & .0653 & .0653 & -- & -- & 875{,}760 & .5788 & .5788 & -- & -- & 7{,}125 \\
Witness rule & .0650 & .0652 & -- & -- & 1{,}381{,}956 & .5773 & .5789 & -- & -- & 16{,}112 \\
Portfolio & .0626 & .0398 & .0328 & .0271 & 123{,}786 & .5914 & .5689 & .5868 & .5691 & 5{,}216 \\
\bottomrule
\end{tabular}
\end{table}

\FloatBarrier
\begin{table}[htbp]
\centering\small
\setlength{\tabcolsep}{4pt}
\caption{Context--parity exact signed utility over ten fits per cell. The budget is $B=k+1$. $D_B$ uses the full query horizon; $D_{B-1}$ is one step shorter. The final column gives the true-distribution optimal nonadaptive value, not a learned estimate.}\label{tab:controls}
\begin{tabular}{@{}rrrrrrr@{}}
\toprule
$k$ & $\epsilon$ & $N$ & $D_{B-1}$ & $D_B$ & Fitted subset & Oracle subset \\
\midrule
2 & 0.0 & 120 & 0.418 & 0.910 & 0.419 & 0.440 \\
2 & 0.0 & 600 & 0.421 & 0.910 & 0.428 & 0.440 \\
2 & 0.1 & 120 & 0.307 & 0.710 & 0.310 & 0.340 \\
2 & 0.1 & 600 & 0.324 & 0.710 & 0.316 & 0.340 \\
3 & 0.0 & 120 & 0.373 & 0.880 & 0.380 & 0.410 \\
3 & 0.0 & 600 & 0.386 & 0.880 & 0.380 & 0.410 \\
3 & 0.1 & 120 & 0.221 & 0.591 & 0.280 & 0.310 \\
3 & 0.1 & 600 & 0.288 & 0.680 & 0.289 & 0.310 \\
\bottomrule
\end{tabular}
\end{table}
 \FloatBarrier

\section{Complete noisy-grid results}\label{app:noisy-grid}
All cells and all prespecified primary contrasts are retained. The aggregate gives every cell equal weight. It is therefore a summary of this fixed grid, not an estimate over an application population. The policy and oracle columns use the same acquisition cost and generator within each cell.
\begin{table}[htbp]
\centering\footnotesize
\setlength{\tabcolsep}{4pt}
\caption{Exact generator signed utility, averaged over ten fresh fits in each prespecified cell. Forced D1 spends its query budget; Subset is optimal under the fitted model among nonadaptive subsets. All 16 cells are retained.}\label{tab:regimes-all}
\begin{tabular}{@{}rrrrrrrrrr@{}}
\toprule
$C$ & $N$ & $c$ & $\sigma$ & $B$ & D2 & D3 & Rollout & Forced D1 & Subset \\
\midrule
2 & 120 & 0.01 & 0.1 & 3 & 0.000 & 0.024 & 0.000 & -0.030 & 0.002 \\
2 & 120 & 0.05 & 0.1 & 3 & 0.000 & -0.006 & -0.002 & -0.143 & 0.000 \\
2 & 120 & 0.1 & 0.1 & 3 & 0.000 & 0.000 & 0.000 & -0.291 & 0.000 \\
2 & 600 & 0.01 & 0.1 & 3 & 0.109 & 0.152 & 0.128 & -0.010 & 0.053 \\
2 & 600 & 0.05 & 0.1 & 3 & 0.000 & 0.042 & 0.035 & -0.133 & 0.000 \\
2 & 600 & 0.1 & 0.1 & 3 & 0.000 & 0.000 & 0.000 & -0.285 & 0.000 \\
5 & 120 & 0.01 & 0.1 & 3 & 0.000 & 0.000 & 0.000 & -0.031 & 0.000 \\
5 & 120 & 0.05 & 0.1 & 3 & 0.000 & 0.000 & 0.000 & -0.150 & 0.000 \\
5 & 120 & 0.1 & 0.1 & 3 & 0.000 & 0.000 & 0.000 & -0.300 & 0.000 \\
5 & 600 & 0.01 & 0.1 & 3 & 0.000 & 0.000 & 0.000 & -0.030 & 0.000 \\
5 & 600 & 0.05 & 0.1 & 3 & 0.000 & 0.000 & 0.000 & -0.150 & 0.000 \\
5 & 600 & 0.1 & 0.1 & 3 & 0.000 & 0.000 & 0.000 & -0.300 & 0.000 \\
2 & 600 & 0.01 & 0.05 & 3 & 0.114 & 0.160 & 0.125 & -0.019 & 0.059 \\
2 & 600 & 0.01 & 0.2 & 3 & 0.005 & 0.118 & 0.059 & -0.013 & 0.022 \\
2 & 600 & 0.01 & 0.1 & 2 & 0.006 & 0.006 & 0.006 & -0.020 & -0.001 \\
2 & 600 & 0.01 & 0.1 & 4 & 0.213 & 0.548 & 0.255 & 0.103 & 0.080 \\
\bottomrule
\end{tabular}
\end{table}
\begin{table}[htbp]
\centering\footnotesize
\setlength{\tabcolsep}{4pt}
\caption{Additional noisy-grid controls, as exact generator signed utility. Cells follow Table~\ref{tab:regimes-all}. CMI and forced D1 spend the query budget; their negative utility can include a cost of acquisition followed by abstention. Context-first pays for context and then follows the ordinary stopping D1 policy. These are fixed controls, not policies tuned after evaluation.}\label{tab:regimes-baselines}
\begin{tabular}{@{}lrrrrrr@{}}
\toprule
Cell & D1 & CMI & Context-first & Forced D1 & Subset & D3 \\
\midrule
0 & 0.000 & 0.004 & -0.010 & -0.030 & 0.002 & 0.024 \\
1 & 0.000 & -0.122 & -0.050 & -0.143 & 0.000 & -0.006 \\
2 & 0.000 & -0.263 & -0.100 & -0.291 & 0.000 & 0.000 \\
3 & 0.000 & 0.060 & -0.010 & -0.010 & 0.053 & 0.152 \\
4 & 0.000 & -0.052 & -0.050 & -0.133 & 0.000 & 0.042 \\
5 & 0.000 & -0.198 & -0.100 & -0.285 & 0.000 & 0.000 \\
6 & 0.000 & -0.030 & -0.010 & -0.031 & 0.000 & 0.000 \\
7 & 0.000 & -0.150 & -0.050 & -0.150 & 0.000 & 0.000 \\
8 & 0.000 & -0.298 & -0.100 & -0.300 & 0.000 & 0.000 \\
9 & 0.000 & -0.029 & -0.010 & -0.030 & 0.000 & 0.000 \\
10 & 0.000 & -0.150 & -0.050 & -0.150 & 0.000 & 0.000 \\
11 & 0.000 & -0.298 & -0.100 & -0.300 & 0.000 & 0.000 \\
12 & 0.000 & 0.074 & -0.010 & -0.019 & 0.059 & 0.160 \\
13 & 0.000 & 0.043 & -0.010 & -0.013 & 0.022 & 0.118 \\
14 & 0.000 & -0.006 & -0.010 & -0.020 & -0.001 & 0.006 \\
15 & 0.000 & 0.168 & 0.011 & 0.103 & 0.080 & 0.548 \\
Mean & 0.000 & -0.078 & -0.041 & -0.113 & 0.013 & 0.065 \\
\bottomrule
\end{tabular}
\end{table}
\begin{table}[htbp]
\centering\footnotesize
\setlength{\tabcolsep}{4pt}
\caption{Known-generator values and learned-policy optimism (fitted minus exact generator utility) for all noisy cells in the order of Table~\ref{tab:regimes-all}. Zero optimism for an abstaining policy does not imply that its model captures the available acquisition value.}\label{tab:regimes-oracles}
\begin{tabular}{@{}rrrrrrr@{}}
\toprule
Cell & Oracle D2 & Oracle D3 & Oracle subset & D2 gap & D3 gap & Rollout gap \\
\midrule
0 & 0.143 & 0.208 & 0.074 & 0.000 & -0.018 & 0.000 \\
1 & 0.000 & 0.080 & 0.000 & 0.000 & 0.009 & 0.004 \\
2 & 0.000 & 0.000 & 0.000 & 0.000 & 0.000 & 0.000 \\
3 & 0.141 & 0.198 & 0.061 & 0.032 & 0.055 & 0.024 \\
4 & 0.000 & 0.081 & 0.000 & 0.000 & 0.026 & 0.017 \\
5 & 0.000 & 0.000 & 0.000 & 0.000 & 0.000 & 0.000 \\
6 & 0.000 & 0.171 & 0.000 & 0.000 & 0.000 & 0.000 \\
7 & 0.000 & 0.049 & 0.000 & 0.000 & 0.000 & 0.000 \\
8 & 0.000 & 0.000 & 0.000 & 0.000 & 0.000 & 0.000 \\
9 & 0.000 & 0.171 & 0.000 & 0.000 & 0.000 & 0.000 \\
10 & 0.000 & 0.049 & 0.000 & 0.000 & 0.000 & 0.000 \\
11 & 0.000 & 0.000 & 0.000 & 0.000 & 0.000 & 0.000 \\
12 & 0.144 & 0.201 & 0.062 & 0.032 & 0.040 & 0.023 \\
13 & 0.000 & 0.152 & 0.029 & 0.001 & -0.008 & 0.009 \\
14 & 0.011 & 0.011 & 0.005 & 0.026 & 0.026 & 0.026 \\
15 & 0.237 & 0.577 & 0.105 & 0.058 & 0.003 & 0.059 \\
\bottomrule
\end{tabular}
\end{table}
\begin{table}[htbp]
\centering\small
\setlength{\tabcolsep}{4pt}
\caption{Prespecified aggregate paired contrasts; intervals use ten replicate-block means. Final column gives the outcome scale and Bonferroni family size. Noisy grid cells have fixed equal weights.}\label{tab:synthetic-contrasts}
\begin{tabular}{@{}llrll@{}}
\toprule
Study & Contrast & Mean & Adjusted interval & Scale; family \\
\midrule
Noisy grid & D3-D2 & 0.037 & [0.032, 0.042] & utility; 2 \\
Noisy grid & D3-rollout & 0.027 & [0.018, 0.037] & utility; 2 \\
Noisy cap 512 & D3-D2 & -0.000 & [-0.001, 0.000] & utility; 6 \\
Noisy cap 512 & D3-rollout & 0.000 & [0.000, 0.000] & utility; 6 \\
Noisy cap 4,096 & D3-D2 & -0.028 & [-0.031, -0.024] & utility; 6 \\
Noisy cap 4,096 & D3-rollout & -0.039 & [-0.044, -0.033] & utility; 6 \\
Noisy cap 32,768 & D3-D2 & 0.025 & [0.022, 0.028] & utility; 6 \\
Noisy cap 32,768 & D3-rollout & 0.015 & [0.007, 0.023] & utility; 6 \\
Parity & adaptive & 0.434 & [0.425, 0.443] & utility; 2 \\
Parity & depth & 0.442 & [0.435, 0.449] & utility; 2 \\
\bottomrule
\end{tabular}
\end{table}
 \FloatBarrier

\section{Noiseless protocol and resource controls}\label{app:noiseless-controls}
A1--A3 denote classifier-aligned terminals at depths one through three. These controls use the original five confirmation fits within the same finite noiseless task. Cross-classifier scores are descriptive, and timings describe these particular executions.
\begin{table}[htbp]
\centering\small
\setlength{\tabcolsep}{4pt}
\caption{Fresh noiseless CUBE-NM: exact 40-pattern accuracy (\%) averaged over five independent policy/classifier fits. A1--A3 use classifier-aligned terminals at the corresponding depths. MLP scores use each fit's own frozen classifier; internal is MAP accuracy on the acquired history, a distinct diagnostic.}\label{tab:official-fresh}
\begin{tabular}{@{}lrrrr@{}}
\toprule
Policy & MLP $B=3$ & MLP $B=5$ & MLP $B=7$ & Internal $B=7$ \\
\midrule
D1 & 28.00 & 42.00 & 50.00 & 57.00 \\
D2 & 53.00 & 74.50 & 77.50 & 100.00 \\
D3 & 53.00 & 75.50 & 80.50 & 100.00 \\
Rollout & 53.00 & 76.50 & 82.50 & 100.00 \\
CMI & 31.50 & 37.00 & 56.00 & 67.00 \\
A1 & 31.00 & 43.50 & 55.50 & 55.50 \\
A2 & 72.00 & 98.00 & 99.50 & 99.50 \\
A3 & 72.00 & 100.00 & 100.00 & 100.00 \\
\bottomrule
\end{tabular}
\end{table}
\begin{table}[htbp]
\centering\small
\setlength{\tabcolsep}{4pt}
\caption{Fresh noiseless primary contrasts at seven queries, in accuracy percentage points. Five paired own-classifier fits; Bonferroni-3 $t_4$ intervals.}\label{tab:official-fresh-contrasts}
\begin{tabular}{@{}lrl@{}}
\toprule
Contrast & Mean & Adjusted interval \\
\midrule
D3-D2 & 3.000 & [-6.600, 12.600] \\
D3-rollout & -2.000 & [-10.518, 6.518] \\
A3-D3 & 19.500 & [13.726, 25.274] \\
\bottomrule
\end{tabular}
\end{table}
\begin{table}[htbp]
\centering\footnotesize
\setlength{\tabcolsep}{4pt}
\caption{Seven-query scoring across independently trained classifiers. Own is the five diagonal entries; Other is the twenty off-diagonal entries. Scorer columns average all five policy fits for that classifier. All values are percentages; this matrix is descriptive and uses the same finite task.}\label{tab:official-cross-scorers}
\begin{tabular}{@{}lrrrrrrr@{}}
\toprule
Policy & Own & Other & $f_1$ & $f_2$ & $f_3$ & $f_4$ & $f_5$ \\
\midrule
D1 & 50.00 & 47.50 & 46.50 & 48.50 & 50.50 & 48.00 & 46.50 \\
D2 & 77.50 & 77.38 & 75.00 & 76.50 & 81.50 & 78.00 & 76.00 \\
D3 & 80.50 & 79.38 & 78.50 & 77.00 & 83.50 & 81.50 & 77.50 \\
Rollout & 82.50 & 83.12 & 81.00 & 81.50 & 83.50 & 83.50 & 85.50 \\
CMI & 56.00 & 56.50 & 58.50 & 54.00 & 58.00 & 55.50 & 56.00 \\
A1 & 55.50 & 48.38 & 51.50 & 54.00 & 45.50 & 51.00 & 47.00 \\
A2 & 99.50 & 86.63 & 86.00 & 87.50 & 88.50 & 92.00 & 92.00 \\
A3 & 100.00 & 88.50 & 88.50 & 90.50 & 89.00 & 94.50 & 91.50 \\
\bottomrule
\end{tabular}
\end{table}
 \FloatBarrier
\begin{table}[htbp]
\centering\footnotesize
\setlength{\tabcolsep}{4pt}
\caption{Fresh noiseless accuracy and actual work under equal transition allowances. Accuracy averages five own-classifier fits. Time is the sum of recorded acquisition-decision times along an episode, in milliseconds.}\label{tab:official-matched}
\begin{tabular}{@{}rrlrrr@{}}
\toprule
$B$ & Allowance & Policy & Accuracy (\%) & Transitions & Time (ms) \\
\midrule
3 & 512 & D1 & 28.00 & 381 & 0.74 \\
3 & 512 & D2 & 28.00 & 1,138 & 5.85 \\
3 & 512 & D3 & 28.00 & 1,252 & 6.07 \\
3 & 512 & Rollout & 28.00 & 1,246 & 7.20 \\
3 & 4,096 & D1 & 28.00 & 381 & 0.81 \\
3 & 4,096 & D2 & 56.00 & 8,415 & 11.19 \\
3 & 4,096 & D3 & 37.00 & 8,254 & 11.14 \\
3 & 4,096 & Rollout & 37.00 & 8,328 & 26.56 \\
3 & 32,768 & D1 & 28.00 & 381 & 0.81 \\
3 & 32,768 & D2 & 53.00 & 32,605 & 28.43 \\
3 & 32,768 & D3 & 37.00 & 48,991 & 41.63 \\
3 & 32,768 & Rollout & 53.00 & 48,895 & 175.76 \\
7 & 512 & D1 & 50.00 & 846 & 1.61 \\
7 & 512 & D2 & 50.00 & 3,001 & 8.38 \\
7 & 512 & D3 & 50.00 & 3,063 & 9.37 \\
7 & 512 & Rollout & 50.00 & 3,098 & 11.33 \\
7 & 4,096 & D1 & 50.00 & 846 & 1.80 \\
7 & 4,096 & D2 & 74.50 & 24,525 & 27.34 \\
7 & 4,096 & D3 & 56.50 & 24,513 & 26.45 \\
7 & 4,096 & Rollout & 65.00 & 24,410 & 90.97 \\
7 & 32,768 & D1 & 50.00 & 846 & 1.68 \\
7 & 32,768 & D2 & 77.50 & 88,001 & 79.32 \\
7 & 32,768 & D3 & 56.50 & 176,809 & 156.27 \\
7 & 32,768 & Rollout & 72.50 & 176,876 & 586.78 \\
\bottomrule
\end{tabular}
\end{table}
\begin{table}[htbp]
\centering\small
\setlength{\tabcolsep}{4pt}
\caption{Noiseless seven-query contrasts under equal transition allowances, in percentage points. Five paired fits; Bonferroni-6 intervals.}\label{tab:official-matched-contrasts}
\begin{tabular}{@{}rlrl@{}}
\toprule
Allowance & Contrast & Mean & Adjusted interval \\
\midrule
512 & D3-D2 & 0.000 & [0.000, 0.000] \\
512 & D3-rollout & 0.000 & [0.000, 0.000] \\
4,096 & D3-D2 & -18.000 & [-34.451, -1.549] \\
4,096 & D3-rollout & -8.500 & [-13.351, -3.649] \\
32,768 & D3-D2 & -21.000 & [-36.625, -5.375] \\
32,768 & D3-rollout & -16.000 & [-27.882, -4.118] \\
\bottomrule
\end{tabular}
\end{table}
 \FloatBarrier
\begin{table}[htbp]
\centering\footnotesize
\setlength{\tabcolsep}{4pt}
\caption{Matched search with classifier-aligned terminals for all four methods. Accuracy (\%) averages five own-classifier fits; transitions and planning-time classifier-history evaluations are actual episode means. Both search and classifier prediction caches are cleared between decisions. These are equal transition ceilings, not identical realized work or FLOPs.}\label{tab:aligned-matched}
\begin{tabular}{@{}rrlrrr@{}}
\toprule
$B$ & Allowance & Policy & Accuracy (\%) & Transitions & MLP histories \\
\midrule
3 & 512 & D1 & 31.00 & 381 & 381 \\
3 & 512 & D2 & 31.00 & 1,272 & 1,253 \\
3 & 512 & D3 & 31.00 & 1,272 & 1,360 \\
3 & 512 & Rollout & 31.00 & 1,272 & 1,199 \\
3 & 4,096 & D1 & 31.00 & 381 & 381 \\
3 & 4,096 & D2 & 65.50 & 8,436 & 7,416 \\
3 & 4,096 & D3 & 43.50 & 8,436 & 7,388 \\
3 & 4,096 & Rollout & 47.50 & 8,436 & 7,755 \\
3 & 32,768 & D1 & 31.00 & 381 & 381 \\
3 & 32,768 & D2 & 72.00 & 32,604 & 16,302 \\
3 & 32,768 & D3 & 43.50 & 49,054 & 24,911 \\
3 & 32,768 & Rollout & 72.00 & 48,927 & 36,865 \\
7 & 512 & D1 & 55.50 & 847 & 847 \\
7 & 512 & D2 & 55.50 & 3,297 & 3,562 \\
7 & 512 & D3 & 55.50 & 3,297 & 3,611 \\
7 & 512 & Rollout & 55.50 & 3,297 & 3,629 \\
7 & 4,096 & D1 & 55.50 & 847 & 847 \\
7 & 4,096 & D2 & 98.00 & 24,800 & 21,634 \\
7 & 4,096 & D3 & 61.50 & 24,797 & 21,648 \\
7 & 4,096 & Rollout & 82.50 & 24,797 & 23,795 \\
7 & 32,768 & D1 & 55.50 & 847 & 847 \\
7 & 32,768 & D2 & 99.50 & 88,299 & 44,150 \\
7 & 32,768 & D3 & 61.50 & 177,214 & 94,437 \\
7 & 32,768 & Rollout & 98.00 & 177,257 & 159,693 \\
\bottomrule
\end{tabular}
\end{table}
\begin{table}[htbp]
\centering\small
\setlength{\tabcolsep}{4pt}
\caption{Aligned-objective seven-query contrasts, in percentage points. Five paired fits and Bonferroni-6 intervals. This additional comparison was specified after the noisy-grid analysis and before any result from the aligned matched stage.}\label{tab:aligned-matched-contrasts}
\begin{tabular}{@{}rlrl@{}}
\toprule
Allowance & Contrast & Mean & Adjusted interval \\
\midrule
512 & D3-D2 & 0.000 & [0.000, 0.000] \\
512 & D3-rollout & 0.000 & [0.000, 0.000] \\
4,096 & D3-D2 & -36.500 & [-48.986, -24.014] \\
4,096 & D3-rollout & -21.000 & [-30.075, -11.925] \\
32,768 & D3-D2 & -38.000 & [-49.758, -26.242] \\
32,768 & D3-rollout & -36.500 & [-49.562, -23.438] \\
\bottomrule
\end{tabular}
\end{table}
 \FloatBarrier

\section{Independent Common panels and sensitivity analyses}\label{app:common-panels}
The two question banks are retained separately as well as averaged equally. Every prior concentration was specified before scoring, and neither the better bank nor a better prior is selected as the primary configuration.
\begin{table}[htbp]
\centering\small
\setlength{\tabcolsep}{4pt}
\caption{Common primary contrasts at seven queries, six references and concentration one, in percentage points. Category-stratified paired target resampling; Bonferroni-8 intervals conditional on the two banks and fitted panels.}\label{tab:llm-fresh-contrasts}
\begin{tabular}{@{}llrl@{}}
\toprule
Direction & Contrast & Mean & Adjusted interval \\
\midrule
$Q\to Q$ & D3-D2 & 9.159 & [4.204, 14.414] \\
$Q\to Q$ & D3-rollout & -5.556 & [-12.312, 1.314] \\
$Q\to G$ & D3-D2 & 1.802 & [-2.402, 5.856] \\
$Q\to G$ & D3-rollout & 1.652 & [-4.167, 7.508] \\
$G\to Q$ & D3-D2 & 2.402 & [-2.703, 7.207] \\
$G\to Q$ & D3-rollout & 2.703 & [-4.505, 9.873] \\
$G\to G$ & D3-D2 & 6.006 & [-0.300, 12.462] \\
$G\to G$ & D3-rollout & -1.652 & [-9.159, 5.669] \\
\bottomrule
\end{tabular}
\end{table}
\begin{table}[htbp]
\centering\small
\setlength{\tabcolsep}{4pt}
\caption{Common seven-query accuracy (\%) by question-bank generator. Six fitting requests and concentration one; neither bank was selected using scores.}\label{tab:llm-banks}
\begin{tabular}{@{}llrrrr@{}}
\toprule
Bank generator & Policy & $Q\to Q$ & $Q\to G$ & $G\to G$ & $G\to Q$ \\
\midrule
Qwen & D1 & 35.44 & 15.02 & 35.74 & 12.61 \\
Qwen & D2 & 44.44 & 15.32 & 42.94 & 16.52 \\
Qwen & D3 & 56.46 & 16.22 & 49.55 & 17.72 \\
Qwen & Rollout & 70.27 & 13.51 & 54.35 & 15.92 \\
Qwen & CMI & 69.37 & 18.92 & 53.15 & 18.32 \\
GLM & D1 & 30.03 & 18.62 & 39.64 & 15.62 \\
GLM & D2 & 32.73 & 17.72 & 45.95 & 15.92 \\
GLM & D3 & 39.04 & 20.42 & 51.35 & 19.52 \\
GLM & Rollout & 36.34 & 19.82 & 49.85 & 15.92 \\
GLM & CMI & 36.34 & 21.92 & 49.85 & 16.52 \\
\bottomrule
\end{tabular}
\end{table}
\begin{table}[htbp]
\centering\small
\setlength{\tabcolsep}{4pt}
\caption{Common seven-query accuracy (\%) under all prespecified prior concentrations, with six fitting requests and equal bank weights. Concentrations are not selected using examiner outcomes.}\label{tab:llm-priors}
\begin{tabular}{@{}rlrrrr@{}}
\toprule
$\alpha$ & Policy & $Q\to Q$ & $Q\to G$ & $G\to G$ & $G\to Q$ \\
\midrule
0.2 & D1 & 32.73 & 16.67 & 37.99 & 14.11 \\
0.2 & D2 & 38.59 & 16.82 & 44.44 & 16.52 \\
0.2 & D3 & 47.75 & 18.32 & 50.60 & 19.22 \\
0.2 & Rollout & 53.30 & 16.67 & 53.15 & 15.32 \\
0.2 & CMI & 54.80 & 19.22 & 50.45 & 17.57 \\
1.0 & D1 & 32.73 & 16.82 & 37.69 & 14.11 \\
1.0 & D2 & 38.59 & 16.52 & 44.44 & 16.22 \\
1.0 & D3 & 47.75 & 18.32 & 50.45 & 18.62 \\
1.0 & Rollout & 53.30 & 16.67 & 52.10 & 15.92 \\
1.0 & CMI & 52.85 & 20.42 & 51.50 & 17.42 \\
4.0 & D1 & 32.58 & 17.12 & 37.24 & 14.11 \\
4.0 & D2 & 38.44 & 16.97 & 44.44 & 16.07 \\
4.0 & D3 & 47.45 & 18.77 & 50.30 & 18.77 \\
4.0 & Rollout & 53.30 & 17.72 & 51.95 & 15.62 \\
4.0 & CMI & 42.64 & 17.72 & 44.59 & 17.57 \\
\bottomrule
\end{tabular}
\end{table}
 \FloatBarrier
\begin{table}[htbp]
\centering\small
\setlength{\tabcolsep}{4pt}
\caption{Common exact-policy accuracy (\%) with six fitting requests, averaged across the two fixed banks. Examiner requests are disjoint from all fitting.}\label{tab:llm-n6}
\begin{tabular}{@{}lrlrrrr@{}}
\toprule
Fit & $B$ & Policy & $Q\to Q$ & $Q\to G$ & $G\to G$ & $G\to Q$ \\
\midrule
n6 & 3 & D1 & 7.06 & 4.95 & 6.91 & 4.50 \\
n6 & 3 & D2 & 7.06 & 4.95 & 6.91 & 4.50 \\
n6 & 3 & D3 & 7.06 & 4.95 & 6.91 & 4.50 \\
n6 & 3 & Rollout & 7.06 & 4.95 & 6.91 & 4.50 \\
n6 & 3 & CMI & 7.06 & 5.26 & 7.21 & 5.11 \\
n6 & 5 & D1 & 20.87 & 13.21 & 20.12 & 10.36 \\
n6 & 5 & D2 & 21.62 & 13.66 & 24.47 & 12.31 \\
n6 & 5 & D3 & 26.88 & 14.26 & 27.03 & 12.91 \\
n6 & 5 & Rollout & 25.23 & 10.81 & 24.32 & 6.31 \\
n6 & 5 & CMI & 24.17 & 12.16 & 24.02 & 11.86 \\
n6 & 7 & D1 & 32.73 & 16.82 & 37.69 & 14.11 \\
n6 & 7 & D2 & 38.59 & 16.52 & 44.44 & 16.22 \\
n6 & 7 & D3 & 47.75 & 18.32 & 50.45 & 18.62 \\
n6 & 7 & Rollout & 53.30 & 16.67 & 52.10 & 15.92 \\
n6 & 7 & CMI & 52.85 & 20.42 & 51.50 & 17.42 \\
\bottomrule
\end{tabular}
\end{table}
 \FloatBarrier
\begin{table}[htbp]
\centering\small
\setlength{\tabcolsep}{4pt}
\caption{Common seven-query accuracy (\%) under equal per-decision transition allowances, averaged over both banks. The policies are bounded approximations; actual work is reported separately.}\label{tab:llm-matched}
\begin{tabular}{@{}rlrrrr@{}}
\toprule
Allowance & Policy & $Q\to Q$ & $Q\to G$ & $G\to G$ & $G\to Q$ \\
\midrule
512 & D1 & 32.73 & 16.82 & 37.69 & 14.11 \\
512 & D2 & 32.58 & 16.97 & 38.89 & 14.11 \\
512 & D3 & 32.73 & 16.67 & 37.54 & 14.11 \\
512 & Rollout & 32.73 & 16.67 & 37.54 & 14.11 \\
4,096 & D1 & 32.73 & 16.82 & 37.69 & 14.11 \\
4,096 & D2 & 38.59 & 16.52 & 44.44 & 16.22 \\
4,096 & D3 & 34.38 & 18.02 & 38.14 & 15.02 \\
4,096 & Rollout & 38.59 & 17.42 & 40.54 & 14.71 \\
32,768 & D1 & 32.73 & 16.82 & 37.69 & 14.11 \\
32,768 & D2 & 38.59 & 16.52 & 44.44 & 16.22 \\
32,768 & D3 & 38.74 & 17.72 & 41.74 & 15.47 \\
32,768 & Rollout & 49.40 & 19.07 & 49.25 & 15.02 \\
\bottomrule
\end{tabular}
\end{table}
\begin{table}[htbp]
\centering\footnotesize
\setlength{\tabcolsep}{4pt}
\caption{Actual Common transitions per seven-query episode under the shared per-decision allowances. Values average both banks. Different examiner trajectories can induce different work for the same frozen policy; none of these counts is FLOPs.}\label{tab:llm-actual-work}
\begin{tabular}{@{}rlrrrr@{}}
\toprule
Allowance & Policy & $Q\to Q$ & $Q\to G$ & $G\to G$ & $G\to Q$ \\
\midrule
512 & D1 & 406 & 406 & 406 & 406 \\
512 & D2 & 3,008 & 3,008 & 3,008 & 3,008 \\
512 & D3 & 2,996 & 2,996 & 2,996 & 2,996 \\
512 & Rollout & 3,089 & 3,089 & 3,088 & 3,088 \\
4,096 & D1 & 406 & 406 & 406 & 406 \\
4,096 & D2 & 21,060 & 21,060 & 21,060 & 21,060 \\
4,096 & D3 & 23,406 & 23,406 & 23,406 & 23,406 \\
4,096 & Rollout & 23,511 & 23,511 & 23,512 & 23,511 \\
32,768 & D1 & 406 & 406 & 406 & 406 \\
32,768 & D2 & 21,060 & 21,060 & 21,060 & 21,060 \\
32,768 & D3 & 166,719 & 166,719 & 166,719 & 166,719 \\
32,768 & Rollout & 132,047 & 132,069 & 132,087 & 132,054 \\
\bottomrule
\end{tabular}
\end{table}
 \FloatBarrier

\section{Interpreting value error and panel diagnostics}\label{app:confirmation-diagnostics}
\paragraph{Value error changes sign with the examiner.}
The fitted seven-query D3 values are 32.69\% for Qwen and 36.55\% for GLM. Both underpredict their same-model examiner accuracy and overpredict cross-model accuracy. Smoothing strongly affects these fitted values without comparably changing executed accuracy: varying concentration from $.2$ to $4$ changes Qwen's fitted D3 value from 46.09\% to 13.61\%, while same-model measured accuracy changes from 47.75\% to 47.45\%. After averaging banks, replacing empirical joint examiner vectors by their item-conditional independent marginals changes D3 and rollout values by at most .28 points across the four directions. With only three examiner vectors per item, this does not identify the population effect of dependence, so we report it as a descriptive diagnostic.

Binary negative log likelihood and Brier score average over the 32 questions and all target--repeat pairs; their denominators count entries, not independent replicates. Full-vector identification supplies all 32 answers to the fitted naive-Bayes predictor. It is not an upper bound on subset-policy accuracy, because additional correlated or mismodeled evidence can change an otherwise correct prediction. Policy optimism is the fitted complete-policy value minus the evaluation value. An abstaining policy can have exactly zero optimism even when its fitted model misses a useful acquisition sequence, as the five-context noisy cells show.
\begin{table}[htbp]
\centering\small
\setlength{\tabcolsep}{4pt}
\caption{Common channel diagnostics at six references and concentration one. Binary NLL (nats) and Brier score average question entries. All-32 accuracy (\%) uses every answer and is not an upper bound on subset-policy accuracy.}\label{tab:llm-channels}
\begin{tabular}{@{}llrrr@{}}
\toprule
Bank generator & Direction & Binary NLL & Brier & All-32 accuracy \\
\midrule
Qwen & $Q\to Q$ & 0.103 & 0.016 & 82.58 \\
Qwen & $Q\to G$ & 0.470 & 0.140 & 26.43 \\
Qwen & $G\to G$ & 0.177 & 0.043 & 70.27 \\
Qwen & $G\to Q$ & 0.412 & 0.124 & 30.63 \\
GLM & $Q\to Q$ & 0.082 & 0.008 & 42.94 \\
GLM & $Q\to G$ & 0.171 & 0.038 & 23.12 \\
GLM & $G\to G$ & 0.095 & 0.013 & 60.66 \\
GLM & $G\to Q$ & 0.162 & 0.035 & 22.82 \\
\bottomrule
\end{tabular}
\end{table}
\begin{table}[htbp]
\centering\small
\setlength{\tabcolsep}{4pt}
\caption{Common complete-policy value diagnostics, averaged over both banks at seven queries, six references and concentration one. Fitted, independent empirical marginals (Ind. marg.) and examiner accuracy are percentages; optimism is their fitted-minus-examiner difference in percentage points. The empirical-marginal calculation reuses the examiner data and is not a population dependence estimate.}\label{tab:llm-value-diagnostics}
\begin{tabular}{@{}llrrrr@{}}
\toprule
Direction & Policy & Fitted & Ind. marg. & Examiner & Optimism \\
\midrule
$Q\to Q$ & D3 & 32.69 & 47.75 & 47.75 & -15.05 \\
$Q\to Q$ & Rollout & 36.15 & 53.03 & 53.30 & -17.16 \\
$Q\to G$ & D3 & 32.69 & 18.13 & 18.32 & 14.38 \\
$Q\to G$ & Rollout & 36.15 & 16.87 & 16.67 & 19.48 \\
$G\to G$ & D3 & 36.55 & 50.70 & 50.45 & -13.90 \\
$G\to G$ & Rollout & 37.70 & 52.07 & 52.10 & -14.41 \\
$G\to Q$ & D3 & 36.55 & 18.62 & 18.62 & 17.94 \\
$G\to Q$ & Rollout & 37.70 & 16.02 & 15.92 & 21.78 \\
\bottomrule
\end{tabular}
\end{table}
 \FloatBarrier
\section{Statistical comparisons and interpretation}\label{app:confirmation-analysis}
We report nominal 95\% intervals adjusted within declared contrast families. A zero-width interval records zero observed fitting-sample variance; it does not prove population-wide policy equivalence. The noisy grid's two primary contrasts are equal-cell mean D3--D2 and D3--rollout utility, using ten independent replicate blocks and Bonferroni-2 $t_9$ intervals. The dependency controls similarly test full-budget depth against one shorter horizon and against nonadaptive subset selection. Noisy cell contrasts use Bonferroni-32 adjustment.

The five fresh noiseless fits support paired $t_4$ intervals: Bonferroni-3 for seven-query alignment/depth/rollout contrasts and Bonferroni-6 for seven-query transition-budget contrasts. The same transition allowances also evaluate every fit of the 16-cell noisy grid; its six aggregate contrasts use Bonferroni-6 $t_9$ intervals. This matched extension was frozen after the unmatched grid was analyzed and retains every original cell. The additional aligned-objective matched stage uses a separate Bonferroni-6 family of seven-query contrasts over five fits, specified before that stage runs. Cross-classifier matrices, mask counterfactuals and repeated timing trials are descriptive. These adjustments apply within their declared families, not jointly to every comparison in the paper. Common's eight primary seven-query contrasts compare D3 with D2 and rollout in each fitting/examiner direction, averaging the two banks equally. Ten thousand category-stratified paired target-bootstrap draws keep all examiner repeats together and apply Bonferroni-8 adjustment. The protocol records three percentage points as a practically relevant Common accuracy difference. These intervals condition on the two banks and fitted panels; they do not establish population uncertainty over question banks or LLM families. Protocol amendments and independent replay are documented in Appendix~\ref{app:confirmation-provenance}.
\section{Protocols, environment and scope}\label{app:confirmation-provenance}
Each stage froze its comparison rules and archived its source bytes before the affected results; later stages followed earlier analyses, so this is a staged study, not a global preregistration. New experiments do not replace earlier evidence.

\paragraph{Environment and pinned benchmarks.} The host is an Apple M5 Max with 18 CPU cores and 64\,GiB of memory. Experiments use Python 3.13.5, NumPy 2.5.4 and PyTorch 2.14.1 with one numerical thread, and local Ollama serves the language models. The source snapshots use AFABench \code{8ebf5e9f0287a7dac09218e1c3c9684b546faddb} and the UoT vocabulary \code{e08f17ff4ceec0f734da66e04d7d35df04ae6460}. Named author code is loaded from checksum-verified files. The experiment calls the shared classifier directly on all 40 support patterns; this is a local replay of the scoring contract, not a run of every author pipeline or a reproduction of published scores.

\paragraph{Common panels.} Both models use the native \code{/api/generate} endpoint, with structured JSON, \code{think=false}, an 8,192-token context limit and a 192-token answer limit. Answer temperature is $.3$ and bank-generation temperature $.7$. Fitting and examiner panels never share a request or seed. There are 3,996 panel completions and 127,872 binary entries, which are not independent experimental units, because each request returns all 32 answers jointly. All completed at their first attempt, and server model digests did not change during collection. The answering model is told the hidden item during panel construction; a compiled policy sees only the selected entries of one examiner vector.

\paragraph{Checks.} Independent probability-product replay agrees with 2,090 noisy-grid and 1,200 context--parity policies to at most $2.67\times10^{-15}$, and Common replay covers 364 frozen policies and 728 policy--examiner scores to $1.84\times10^{-15}$. Protocol amendments and failed attempts are kept in the study records; none selected a result. These checks are internal validation, not third-party replication.

\paragraph{Limits of the causal interpretation.} The parity control isolates a designed interaction. The noisy grid changes several aspects of estimation difficulty through context count and reference size, so its results do not identify a universal depth rule. The noiseless classifier fits measure variation within a 40-pattern task. Independent question banks and answer requests separate sampling from transfer, but do not establish factual truth or generalization across vocabularies, prompts, serving systems or model families. Rankings under bounded search depend on the candidate order and fallback. Shared computational ceilings are not identical realized work, floating-point operations, energy or latency.
 
\FloatBarrier
\FloatBarrier
\section{Learned-acquisition comparison protocol}\label{app:learned-baselines}
The additional comparison uses the same noiseless CUBE-NM acquisition contract: 600 training and 200 validation observations, a separately trained shared masked MLP, an empty initial mask, 51 unit-cost groups, and budgets $B\in\{3,5,7\}$. The five original confirmation fits are explicitly reused; generator/split seeds 62142--62146 and classifier seeds 63000--63004 define five additional fits. The two cohorts are reported separately. No test outcome chooses an architecture, checkpoint or acquisition hyperparameter.

\paragraph{DIME.}
DIME learns a discriminative estimate of conditional information gain \citep{gadgil2024dime}. We execute AFABench's pinned \code{MaskingPretrainer} and \code{CMIEstimator} definitions with its noiseless CUBE-NM configuration: two 128-unit ReLU hidden layers, dropout $.3$, Adam learning rate $.001$, batch size 128, at most 200 pretraining epochs and 250 acquisition-training epochs per budget. The author validation checkpoint, learning-rate, epsilon-decay and early-stopping rules remain active. The first two linear modules are shared between predictor and value networks. Training uses inverse-frequency class weights normalized as in the benchmark; pretraining uses its separate author weighting rule. The training horizon is set to the evaluated hard budget, matching the workflow override, and the author grouped unmasker reveals the entire context with one selection. Scalar raw-coordinate costs are $.2$ for the five context coordinates and one otherwise, giving unit group costs. The private predictor supplies DIME's training signal; final accuracy uses the same external classifier as every planner.

The pinned implementation permits repeated selections during training and includes shared parameters twice in the optimizer list, which produces a warning in the installed PyTorch version. The primary DIME baseline retains those behaviors for implementation fidelity. Deployment masks acquired groups and cannot repeat a query. Consequently, this is a comparison with the pinned benchmark implementation, not evidence about every possible implementation or tuning of DIME.

A descriptive control, DIME-U, includes each shared tensor once in Adam; it reaches 33.00\% at seven queries, against 33.50\% for the original.

\paragraph{SEFA.}
SEFA trains stochastic feature encodings and scores acquisitions using prediction gradients through latent samples \citep{norcliffe2025}. We use the author repository at revision \code{3af8da2c9212c294fc7fedad27bc31045db7607a}, with the 50 scalar coordinates followed by one categorical context taking five values. This is a lossless reordering of the 51 acquisition groups, and its categorical embedding represents a missing context separately from every observed category. The common scorer always receives the original 55-coordinate representation.

We prespecify author configurations 1, 4 and 7. They use respectively latent dimensions $4/6/4$ per feature, predictor widths $100/150/180$, encoder widths $20/50/40$, bottleneck coefficients $.0005/.0005/.0008$, learning rates $.001/.001/.0005$, and plateau patience $5/5/8$. Each has two hidden layers in the predictor and continuous encoder, batch size 128, and 120 epochs. Every candidate uses 100 training and 200 prediction/acquisition latent samples. Within each candidate, the author training loop selects the checkpoint by its internal validation accuracy integrated over acquisitions zero through seven. For each evaluated budget, the external validation accuracy averaged over three fixed acquisition seeds selects among the three checkpoints; ties choose the smaller configuration ID. Selection records are saved before evaluation begins.

SEFA's acquisition procedure remains stochastic. We evaluate ten fixed, disjoint acquisition seeds on each of the 40 support atoms, then average within each fitted model. The support is exhaustively enumerated, but the expectation over SEFA's internal randomness is estimated rather than exact. A separate ten-seed evaluation uses the reserved 200-row test split. Neither the ten draws nor the repeated support atoms are treated as independent fitted models.

\paragraph{Compatibility.} Author source bytes and licenses are retained. SEFA's block-linear layer changes \code{view} to \code{reshape} for the current PyTorch runtime, and separate calculations verify its values and gradients. These adaptations do not reproduce the authors' software environment or their published scores.

\paragraph{Inference and resource accounting.}
The primary learned-baseline contrasts use the five additional fits at seven queries: ordinary D3 and aligned D3 versus each of DIME and SEFA. Paired $t_4$ intervals use Bonferroni adjustment across these four contrasts. A separate two-contrast family replicates D3--D2 and D3--rollout. These small-sample intervals assume approximately normal paired fitting effects; they do not estimate variation over tasks. Original-cohort, pooled-cohort, other-budget and reserved-test results are descriptive. Matching training data, feature costs and the scorer does not match optimization effort, floating-point operations or energy. In particular, three SEFA configurations and three budget-specific DIME fits have different training costs from fitting a finite observation model.

GDFS \citep{covert2023gdfs} is evaluated under the same contract (Table~\ref{tab:gdfs}). The pinned AFABench AACO path evaluates only one additional group, so Appendix~\ref{app:aaco} instead adapts the authors' non-myopic AACO code. NM-PPG and BRiG-AFA are relevant recent alternatives \citep{nmppg2026,feng2026brig}; their training pipelines are not reproduced here.

Matching the final scorer does not match how acquisition uses it. Aligned planners query the frozen shared MLP at hypothetical histories. DIME and SEFA use their private predictors for acquisition; SEFA's external validation score selects a configuration but does not retrain its acquisition objective. Consequently, comparisons with aligned search also change predictor coupling. The ordinary-D3 comparisons and the separate within-model depth contrasts help distinguish these interventions, but none removes all inductive-bias and optimization differences between model classes.

\begin{table}[htbp]
\centering\small
\caption{Reused cohort shared-classifier accuracies (percent).}\label{tab:learned-reused}
\begin{tabular}{@{}lrrr@{}}
\toprule
Method & B3 & B5 & B7 \\
\midrule
D1 & 28.00 & 42.00 & 50.00 \\
D2 & 53.00 & 74.50 & 77.50 \\
D3 & 53.00 & 75.50 & 80.50 \\
Rollout & 53.00 & 76.50 & 82.50 \\
CMI & 31.50 & 37.00 & 56.00 \\
Aligned D1 & 31.00 & 43.50 & 55.50 \\
Aligned D2 & 72.00 & 98.00 & 99.50 \\
Aligned D3 & 72.00 & 100.00 & 100.00 \\
DIME & 21.50 & 27.00 & 33.00 \\
DIME-U & 23.00 & 26.50 & 31.50 \\
SEFA & 28.80 & 37.65 & 46.35 \\
\bottomrule
\end{tabular}
\end{table}

\FloatBarrier

\begin{table}[htbp]
\centering\small
\caption{Fresh cohort shared-classifier accuracies (percent).}\label{tab:learned-fresh}
\begin{tabular}{@{}lrrr@{}}
\toprule
Method & B3 & B5 & B7 \\
\midrule
D1 & 29.00 & 42.50 & 51.00 \\
D2 & 56.00 & 77.00 & 79.00 \\
D3 & 56.00 & 76.50 & 77.50 \\
Rollout & 56.00 & 77.00 & 85.00 \\
CMI & 27.00 & 49.00 & 63.50 \\
Aligned D1 & 30.50 & 44.00 & 54.50 \\
Aligned D2 & 71.00 & 99.00 & 99.50 \\
Aligned D3 & 71.00 & 99.00 & 100.00 \\
DIME & 23.50 & 26.50 & 33.50 \\
DIME-U & 23.50 & 27.50 & 33.00 \\
SEFA & 28.95 & 36.55 & 45.55 \\
\bottomrule
\end{tabular}
\end{table}

\FloatBarrier

\begin{table}[htbp]
\centering\small
\caption{Fresh-cohort primary learned-baseline contrasts, percentage points. Paired t intervals use four degrees of freedom and Bonferroni-four adjustment.}\label{tab:learned-contrasts}
\begin{tabular}{@{}lr@{}}
\toprule
Contrast & Difference [adjusted interval] \\
\midrule
D3--DIME & +44.00 [25.13, 62.87] \\
D3--SEFA & +31.95 [24.01, 39.89] \\
Aligned D3--DIME & +66.50 [50.28, 82.72] \\
Aligned D3--SEFA & +54.45 [52.26, 56.64] \\
\bottomrule
\end{tabular}
\end{table}

\paragraph{GDFS.} A separate stage, frozen before any GDFS model was trained, adds greedy dynamic feature selection \citep{covert2023gdfs}. It uses the pinned AFABench implementation and its CUBE configuration: [128,128] ReLU networks with dropout .3, Adam at $10^{-3}$, masked pretraining for up to 100 epochs, and up to 250 GDFS epochs with patience 5. Splits, group costs and shared scorers are those of the DIME and SEFA comparison, and acquisition spends the hard budget greedily.

GDFS reaches 22.5\%, 33.0\% and 38.0\% at three, five and seven queries in the fresh cohort, and 39.0\% at seven queries in the original cohort (Table~\ref{tab:gdfs}). That is between DIME and SEFA, about 40 points below the MAP-terminal planners, as the frozen expectation stated. The comparator columns reproduce Table~\ref{tab:learned-baselines}.
\begin{table}[htbp]
\centering\footnotesize
\setlength{\tabcolsep}{4pt}
\caption{GDFS in the learned-acquisition comparison on noiseless CUBE-NM: own shared-MLP accuracy (\%) on the 40-pattern support, five fits per cohort. GDFS uses the pinned AFABench implementation and CUBE configuration under the same splits, group costs and scorers as DIME and SEFA; comparators are the frozen results of that stage. Planner columns use the fitted MAP terminal. Contrasts are paired over five fits, Bonferroni-6 $t_4$, in percentage points.}\label{tab:gdfs}
\begin{tabular}{@{}lrrrrrrrr@{}}
\toprule
Cohort & $B$ & GDFS & DIME & SEFA & CMI & D2 & D3 & Rollout \\
\midrule
Fresh & 3 & 22.5 & 23.5 & 28.9 & 27.0 & 56.0 & 56.0 & 56.0 \\
Fresh & 5 & 33.0 & 26.5 & 36.5 & 49.0 & 77.0 & 76.5 & 77.0 \\
Fresh & 7 & 38.0 & 33.5 & 45.5 & 63.5 & 79.0 & 77.5 & 85.0 \\
Original & 7 & 39.0 & 33.0 & 46.4 & 56.0 & 77.5 & 80.5 & 82.5 \\
\multicolumn{2}{l}{Fresh B7} & \multicolumn{7}{l}{GDFS $-$ DIME: +4.5 [$-$10.5, +19.5]} \\
\multicolumn{2}{l}{Fresh B7} & \multicolumn{7}{l}{GDFS $-$ SEFA: $-$7.5 [$-$13.4, $-$1.7]} \\
\multicolumn{2}{l}{Fresh B7} & \multicolumn{7}{l}{GDFS $-$ D3: $-$39.5 [$-$50.6, $-$28.4]} \\
\multicolumn{2}{l}{Original B7} & \multicolumn{7}{l}{GDFS $-$ DIME: +6.0 [$-$8.7, +20.7]} \\
\multicolumn{2}{l}{Original B7} & \multicolumn{7}{l}{GDFS $-$ SEFA: $-$7.4 [$-$19.0, +4.3]} \\
\multicolumn{2}{l}{Original B7} & \multicolumn{7}{l}{GDFS $-$ D3: $-$41.5 [$-$47.7, $-$35.3]} \\
\bottomrule
\end{tabular}
\end{table}

\begin{table}[htbp]
\centering\small
\caption{Descriptive diagnostics for the five additional fits (percent). External and private predictors use the same selected masks. Information denotes mean maximum true class posterior among compatible support atoms; it is an evaluation-only diagnostic. Test uses the reserved 200-row split.}\label{tab:learned-diagnostics}
\begin{tabular}{@{}lrrrrrr@{}}
\toprule
Method & Budget & External & Private & Information & Test & Context first \\
\midrule
DIME & 3 & 23.50 & 18.50 & 25.00 & 21.50 & 0.00 \\
DIME & 5 & 26.50 & 25.50 & 36.50 & 25.00 & 0.00 \\
DIME & 7 & 33.50 & 32.50 & 37.50 & 34.90 & 0.00 \\
SEFA & 3 & 28.95 & 29.35 & 31.15 & 27.62 & 0.00 \\
SEFA & 5 & 36.55 & 39.65 & 43.98 & 38.53 & 0.00 \\
SEFA & 7 & 45.55 & 47.65 & 51.32 & 46.10 & 0.00 \\
\bottomrule
\end{tabular}
\end{table}
 \FloatBarrier
\section{A third-model Common examination}\label{app:third-examiner}
\emph{Problem.} Both families in the main comparison also fitted the policies and wrote the banks. Does the collapse hold for a family that did neither? \emph{Approach.} With all banks, fitting panels and policies frozen, the locally installed Nemotron answered three new requests per item and bank (666 examiner vectors). The four primary contrasts are D3 $-$ D2 and D3 $-$ rollout for each fitting family at seven questions, with 10,000 category-stratified paired-target bootstrap draws and Bonferroni-4.

\emph{Evidence} (Table~\ref{tab:nemotron}). D3 $-$ D2 is $-0.90$ ($[-4.65,2.85]$) points for Qwen fits and $-1.50$ ($[-4.95,1.65]$) for GLM fits, and no contrast establishes a D3 gain over D2 or rollout. After these results were inspected, six further Nemotron fitting requests per item (1,332 calls) gave a same-model fit scored on the same examiners: D3 reaches 29.28\%, D2 27.18\% and rollout 30.93\%, and neither adjusted interval excludes zero. This follow-up is exploratory, because the examiners were no longer untouched. \emph{Reading.} Fixed likelihoods from either family transfer poorly to a third. The study cannot separate an effect of depth from the change of model.

\begin{table}[htbp]
\centering\small
\caption{Nemotron examination at seven queries, equal-bank accuracy (\%). Policies and banks are unchanged.}\label{tab:nemotron}
\begin{tabular}{@{}lrr@{}}
\toprule
Method & Qwen fit & GLM fit \\
\midrule
D1 & 9.46 & 7.81 \\
D2 & 9.61 & 8.26 \\
D3 & 8.71 & 6.76 \\
Rollout & 5.86 & 6.16 \\
CMI & 8.71 & 6.76 \\
\bottomrule
\end{tabular}
\end{table}
  \end{document}